\documentclass[twoside,11pt]{article}

\usepackage{jmlr2e}
\usepackage{multirow}
\usepackage{pdfsync}

\newcommand{\commentout}[1]{}

\jmlrheading{1}{2000}{1-48}{4/00}{10/00}{S. Reece, S. Ghosh,
  A. Rogers, N. Jennings and S. Roberts}

\ShortHeadings{Periodic Latent Force Models}{S. Reece, S. Ghosh,
  A. Rogers, N. R. Jennings and S. Roberts}
\firstpageno{1}

\usepackage{graphicx}
\usepackage{epstopdf}
\usepackage{amsmath}
\usepackage{amssymb}

\usepackage{algorithm}
\usepackage{algorithmic}

\begin{document} 

\title{Efficient State-Space Inference of Periodic Latent Force
  Models}

\author{\name Steven Reece \email reece@robots.ox.ac.uk\\
\name Stephen Roberts \email sjrob@robots.ox.ac.uk\\
\addr Department of Engineering Science\\
University of Oxford\\
Parks Road\\
Oxford OX1 3PJ, UK
\AND
\name Siddhartha Ghosh \email sg2@ecs.soton.ac.uk\\ 
\name Alex Rogers \email acr@ecs.soton.ac.uk\\ 
\addr Electronics and Computer Science\\ 
University of Southampton\\
Southampton SO17 1BJ, UK
\AND
\name Nicholas R. Jennings \email nrj@ecs.soton.ac.uk\\
\addr Electronics and Computer Science\\  
University of Southampton\\ 
Southampton SO17 1BJ, UK\\ 
and\\ 
Department of Computing and Information Technology\\
King Abdulaziz University\\
Saudi Arabia}

\editor{}

\maketitle

\begin{abstract}%
  Latent force models (LFM) are principled approaches to incorporating
  solutions to differential equations within non-parametric inference
  methods. Unfortunately, the development and application of LFMs can
  be inhibited by their computational cost, especially when
  closed-form solutions for the LFM are unavailable, as is the case in
  many real world problems where these latent forces exhibit periodic
  behaviour.  Given this, we develop a new sparse representation of
  LFMs which considerably improves their computational efficiency, as
  well as broadening their applicability, in a principled way, to
  domains with periodic or near periodic latent forces. Our approach
  uses a linear basis model to approximate one generative model for
  each periodic force.  We assume that the latent forces are generated
  from Gaussian process priors and develop a linear basis model which
  fully expresses these priors.  We apply our approach to model the
  thermal dynamics of domestic buildings and show that it is effective
  at predicting day-ahead temperatures within the homes.  We also
  apply our approach within queueing theory in which quasi-periodic
  arrival rates are modelled as latent forces. In both cases, we
  demonstrate that our approach can be implemented efficiently using
  state-space methods which encode the linear dynamic systems via
  LFMs. Further, we show that state estimates obtained using periodic
  latent force models can reduce the root mean squared error to $17\%$
  of that from non-periodic models and $27\%$ of the nearest rival
  approach which is the resonator model
  \citep{sarkka12,hartikainen12}.
\end{abstract}

\begin{keywords}
latent force models, Gaussian processes, Kalman filter, kernel
  principle component analysis, queueing theory
\end{keywords}

\section{INTRODUCTION}
\label{sec:intro} 

Latent force models (LFMs) have received considerable interest in the
machine learning community as they combine underlying physical
knowledge of a system with data driven models expressed as Bayesian
non-parametric Gaussian process (GP) priors
\citep[see, for example,][]{alvarez09,hartikainen10}.  In more detail, the physical
process that generates the data is typically represented by one or
more differential equations.  These differential equations can then be
accommodated within covariance functions along with the data driven
priors.  Doing so allows inferences to be drawn in regimes where data
may be sparse or absent, where a purely data driven model will
typically perform poorly.  To date, such models have been applied in
areas such as computational biology and understanding motion patterns
\citep{alvarez09,alvarez10}.

Despite growing interest in LFMs, their real world applicability has
been limited as inference using LFMs expressed directly through
covariance functions can be computationally prohibitive on large data
sets.  It is well known that regression with GPs imposes high
computational cost which scales as $\mathcal{O}(N^3 T^3)$ during
training, where $N$ is the dimension of the data observed at each time
point and $T$ is the number of time points.  However, it has also been
shown that training LFMs using state-space methods can be considerably
less computationally demanding \citep{rasmussen06,hartikainen10} as
state-space methods scale as $\mathcal{O}(N^3 T)$.  It is this
computational saving that motivates the state-space approach to LFM
inference in this paper.

The state-space approach to LFM inference advocated by
\citet{hartikainen10,hartikainen11} augments the state vector so that
Mat\'ern and squared-exponential priors can be accommodated (although
only approximately in the case of the squared-exponential).  All the
information encoded within the GP prior (that is, process smoothness,
stationarity etc) is fully captured within their state-space
representation.  However, their approach assumes that the LFM kernel's
inverse power spectrum can be represented by a power series in the
frequency domain.  Unfortunately, this requirement severely inhibits
the applicability of their approach and, consequently, only a small
repertoire of GP priors have been investigated within LFMs to date,
namely, squared-exponential and Mat\'ern kernels.  Specifically, the
state-space approach advocated by \citet{hartikainen10} does not
accommodate periodic kernels as we shall demonstrate in this paper.
This is a key limitation as periodicity is common in many physical
processes as we shall demonstrate in our empirical evaluation.
Expressing our prior knowledge of the periodicity, as a GP prior,
within the state-space approach is the key challenge problem addressed
in this paper.

Thus, against this background, we describe a principled method for
incorporating stationary periodic, non-stationary periodic and
quasi-periodic Gaussian process priors within state-space approaches
to LFM inference.  Within our approach all LFM parameters can be
inferred using Bayesian methods or maximum likelihood and thus we
circumvent the need to set any of these parameters by hand.  Further,
to accommodate periodic and quasi-periodic models within LFMs we
develop a novel state-space approach to inference.  In particular, we
propose to represent periodic and quasi-periodic driving forces, which
are assumed smooth, by linear basis models (LBMs) with eigenfunction
basis functions derived using kernel principal component analysis
(KPCA) in the temporal domain. These basis models, although parametric
in form, are optimised so that their generative properties accurately
approximate the driving force kernel.  We will show that efficient
inference can then be performed using a state-space approach by
augmenting the state with additional variables which sparsely
represent the periodic latent forces.

Our LBM approach to accommodating periodic kernels is inspired by the
{\em resonator model} \citep{sarkka12,hartikainen12} in which the
periodic process is modelled as a superposition of {\em resonators},
each of which can be represented within the state-vector.
Unfortunately, the resonator model, in its current form, does not
encode all the underlying GP prior information of the periodic process
as the resonator is not tailored to accommodate all the prior
information encoded via the covariance function (see
Section~\ref{sec:statebasedapproaches} for more detail).  An
alternative approach to modelling stationary kernels, including
periodic kernels, is sparse spectrum Gaussian process regression
(SSGPR) of \citet{gredilla11}.  This approach is similar in spirit to
the resonator model in that it encodes stationary GP priors via basis
functions (sinusoidal functions, in this case).  However, unlike the
resonator model, the SSGPR is able to encode the GP prior by
reinterpreting the spectral density of a stationary GP kernel as a
probability density function over frequency space.  This pdf is then
sampled using Monte Carlo to yield the frequencies of the sinusoidal
basis functions.  Unfortunately, this stochastic approach can often
provide a poor approximation to the covariance function (see
Section~\ref{sec:LBM} for more detail).

We shall develop a LBM which captures all the information encoded
within the GP prior and demonstrate its superior accuracy over the
resonator model and the SSGPR.  We shall also establish the close link
between the resonator basis and the eigenfunction basis used in our
approach and consequently, derive a novel method for tailoring the
resonator basis to accommodate all the information encoded within the
covariance function.

Our research is driven by two specific applications although the
methods that we propose are of general applicability.  Specifically,
we apply our approach to the estimation and prediction of the
behaviour of customer queues in call centres, based on flow models of
queue dynamics represented as LFMs.  The behaviour of queues is of
general importance in several applications including communication
networks \citep{wang96}, weather monitoring \citep{sims05} and truck
coordination at ports \citep{ji10}.  Accurate predictions of the
customer queue arrival rates based on an underlying LFM is a key
requirement for determining the number of call centre agents required
at various times throughout the day.  We also apply our approach to
the estimation and prediction of the internal temperature within a
home, based on thermal models of home heating systems represented as
LFMs. Accurate predictions of the internal temperature based on an
underlying LFM is a key component for predicting energy used in
heating a home and, consequently, an integral part of many home energy
saving systems \citep{madsenmodel}.  These applications demonstrate
our approach under two different modelling conditions, the queue LFM
is nonlinear whereas the thermal LFM is linear, while the queue application
is a tracking application and regular measurements are available
whereas the thermal application requires long term predictions (a day
ahead) during which no measurements are available.

In more detail, telephone call centre managers are concerned with
staffing and specifically, assigning the appropriate number of agents
to guarantee that the customers' queueing time does not prohibit sales
\citep{feigin06}.  Although there is significant literature on
attempts to accurately model the dynamics of queues, it has failed to
offer a method for inferring the highly quasi-periodic arrival rates
from sparse measurements of the queue lengths \citep{wang96}.
Determining such arrival rates is key to predicting queue lengths, and
hence customer waiting times.  These predictions help the call centre
manager to plan staffing throughout the day to ensure an acceptable
customer waiting time.  We will demonstrate that our approach to
modelling LFMs is capable of inferring these unknown arrival rates.
Furthermore, although the dynamic system in this application is
nonlinear and the arrival rate is quasi-periodic, it is still Markovian
and, consequently, a state-space approach to inference is ideally
suited to this application.

Energy saving in homes is a key issue as governments aim to reduce the
carbon footprint of their countries.  A significant amount of energy
is expended in heating homes and home owners need to be encouraged to
reduce their energy consumption and carbon emissions incurred through
home heating \citep{mackay09,decc09}.  Consequently, we apply our
approach to the estimation and prediction of internal temperatures
using thermal models of home heating systems. Our approach allows us
to make day ahead predictions of the energy usage, which can then be
fed back to the householder in real-time so that they can take
appropriate mitigating actions to reduce their energy consumption.
Home heating systems typically consist of a thermostat with a
set-point that controls the activations of a gas or electrical boiler
to ensure that the internal temperature follows the set-point.
Although there is significant literature on attempts to accurately
model the thermal dynamics of buildings, it has failed to take into
account the daily human behaviours within their homes, which can have
a significant impact on the energy signatures obtained from similar
homes \citep{madsenmodel}.  For instance, during cold periods, a
householder may deploy an additional heater or, in hot periods, open a
window. Furthermore, the thermal dynamics of real homes are more
complex in reality than existing thermal models suggest; sunlight
through windows contributes to extra heat while open windows cause
heat loss.  Residual heat can also be retained by thermal blocks such
as walls and ceilings that then re-radiate heat.  Crucially, many of
these heat sources are periodic in nature.  For instance, an
additional heater may be switched on every night during cold periods,
whilst the diurnal sun cycle will contribute additional heat during
the day.  We will demonstrate that our approach is capable of
inferring these unknown periodic heat sources.  Again, the dynamic
system in this application is linear and Markovian and, consequently,
a state-space approach to inference is again ideally suited to this problem.

In undertaking this work, we advance the state of the art in the
following ways:
\begin{itemize}
\item \label{novelassert} We offer the only principled approach to incorporating all
  Gaussian process prior models within a state-space approach to
  inference with LFMs.~\footnote{What this paper does not aim to
    establish is the value of GP models per se over other models.  The
    paper thus focusses on developing efficient, scalable
    representations and tools for performing GP inference.}
\item We are the first to demonstrate that the eigenfunction model of
  Gaussian process priors out-performs an alternative approach to
  modelling periodic Gaussian process priors; namely, the sparse
  spectrum Gaussian process regression (SSGPR) approach developed
  by \citet{gredilla11}.
\item We demonstrate, for the first time, the close link between the
  eigenfunction model and the resonator
  model \citep{sarkka12,hartikainen12,solin13}.  Consequently, we
  offer a novel mechanism for incorporating all information encoded
  within the latent force covariance function into the resonator
  model.
\item We propose the only approach that is able to incorporate
  all types of periodic Gaussian process priors within a state-space
  approach to LFM inference.  These priors include stationary
  periodic, non-stationary periodic and quasi-periodic covariance
  functions.
\item We are the first to apply LFMs to queueing theory, specifically
  to the modelling of queue arrival rates. Through empirical
  evaluation, we show that for tracking the customer queue lengths in
  the call centre application, the RMSE of our approach using a
  quasi-periodic kernel model of the arrival rate can be $17\%$ of
  that using the same approach with a non-periodic kernel model.
\item We are the first to apply LFMs to the modelling of thermal
  dynamics within real homes, specifically to unknown physical thermal
  processes. We show that for day ahead predictions of temperature in
  homes, the RMSE of our approach is $45\%$ of that obtained using the
  resonator model \citep{solin13} when the latent forces exhibit
  quasi-periodic behaviour.
\end{itemize}

The structure of our paper is as follows: in Section~\ref{sec:gpintro}
we review approaches to regression and time-series analysis using
Gaussian processes and the Kalman filter.  In Section~\ref{sec:lfm} we
review LFMs with a particular focus on periodic latent forces and then
in Section~\ref{sec:statebasedapproaches} we present a critique of the
existing state-space approaches to inference with LFMs.  In
Section~\ref{sec:LBM} we present a novel approach to representing
periodic LFMs by linear basis models.  We critique the existing
spectral models for representing periodic, stationary Gaussian process
priors and argue that kernel principal component analysis is the most
effective approach to inferring the Fourier basis for the
corresponding LBMs. In Section~\ref{sec:quasi} we extend our approach
to representing quasi-periodic latent forces by linear basis models.
Then, in Section~\ref{sec:sslfm} (with further details in
Appendix~\ref{app:discretejump}), we derive a state-space approach to
inference with LFMs which accommodates both periodic and
quasi-periodic forces via LBMs.  In Section~\ref{sec:appqueue}, we
empirically demonstrate the utility of our approach in tracking the
length of call centre customer queues in the presence of, initially,
unknown arrival rates which are modelled as latent forces.  In
Section~\ref{sec:appthermal} we also apply our approach to predicting
the internal temperature of homes in the presence of, a priori,
unknown residual heat periodic latent forces.  Furthermore, we
demonstrate our approach on both single output and multi-output
Gaussian process thermal models.  We conclude in
Section~\ref{sec:conclusions}.  Finally, in Appendix~\ref{app:eigres}
we demonstrate the theoretical link between the eigenfunction basis
used in our approach and the basis used within the resonator model.
Consequently, we offer a novel method for encoding periodic latent
force covariance functions within the resonator model.

\section{A REVIEW OF GAUSSIAN PROCESS PRIORS AND INFERENCE}
\label{sec:gpintro} 
A Gaussian process (GP) is often thought of as a Gaussian distribution 
over functions \citep{rasmussen06}.  A GP is fully
described by its {\em mean function}, $\mu$, and {\em covariance
  function}, $K$.  A draw, $f$, from a GP is traditionally written,
\begin{eqnarray*}
f\sim\mathcal{GP}(\mu,K)\ .
\end{eqnarray*}
The value of the function, $f$, at inputs $X$
is denoted $f(X)$.  Similarly, the value of the mean function and
covariance function at these inputs are denoted $\mu(X)$ and $K(X,X)$,
respectively.  The meaning of a GP becomes clear when we consider
that, for any finite set of inputs, $X$, $f(X)$ is a draw from a
multi-variate Gaussian, $f(X)\sim\mathcal{N}(\mu(X),K(X,X))$.

Suppose we have a set of training data,
\begin{eqnarray}
D=\{(x_1,y_1),\ldots,(x_n,y_n)\}\ ,
\label{evidence}
\end{eqnarray}
drawn from a function, $f$,
\begin{align*}  
y_i=f(x_i)+\epsilon_i\ ,
\end{align*}  
where $\epsilon_i$ is a zero-mean Gaussian random variable with  
variance $\sigma^2$.  For convenience both inputs and outputs are  
aggregated into sets $X=\{x_1,\ldots,x_n\}$ and
$Y=\{y_1,\ldots,y_n\}$, respectively.  The GP estimates the value of 
the function $f$ at test inputs $X_*=\{x_{*1},\ldots,x_{*m}\}$.  The 
basic GP regression equations are given by,
\begin{align}
\bar{f}_* &= \mu(X_*)+K(X_*,X)[K(X,X)+\sigma^2I]^{-1}(Y-\mu(X))\ ,\label{GP1}\\
\text{Var}(f_*) &=
K(X_*,X_*)-K(X_*,X)[K(X,X)+\sigma^2I]^{-1}K(X_*,X)^T\ , \label{GP2}
\end{align} 
where $I$ is the identity matrix, $\bar{f}_*$ is the posterior mean 
function at $X_*$ and $\text{Var}(f_*)$ is the posterior covariance
 \citep{rasmussen06}.  The inversion operation present in 
Equations~\eqref{GP1} and~\eqref{GP2} is the source of the cubic 
computational complexity reported in the previous section.
 
The matrix $K(X,X)$ is the covariance of the Gaussian prior
distribution over $f(X)$.  The covariance matrix has elements,
\begin{align*} 
K(x_i,x_j)=\text{Cov}(f(x_i),f(x_j))\ ,
\end{align*} 
where the term $K(X_*,X)$ is the covariance between the function, $f$,
evaluated at the test inputs $X_*$ and the training inputs $X$.  The
function $K$ is alternatively called the {\em kernel} or the {\em
  covariance function}.  There are many off-the-shelf kernels
available (see, for example, \citet{rasmussen06}) and appropriate
kernels are chosen to model functions with requisite qualitative
properties such as smoothness and stationarity.  Further, basic
kernels can be combined together to form more sophisticated kernels
tailored to particular modelling needs.  The mean function encodes our
prior knowledge of the function mean.  For ease of exposition we will
assume that the mean function is zero a priori although the approaches
to GP inference presented in later sections are not limited to this
case.
 
The GP parameters $\theta$ (which includes $\sigma$ and hyperparameters
associated with the covariance function) can be inferred from the data
through Bayes' rule,
\begin{eqnarray*} 
p(\theta\mid Y,X)=\frac{p(Y\mid X,\theta)}{p(Y\mid X)}p(\theta)\ .
\end{eqnarray*} 
The parameters are usually given a vague prior distribution
$p(\theta)$.  In this paper, since our applications in
Sections~\ref{sec:appqueue} and~\ref{sec:appthermal} exploit large
data sets, we use maximum likelihood to infer the parameters and
identify the assumed unique value for $\theta$ which maximises
$p(Y\mid X,\theta)$.  This approach is preferred over full Bayesian
marginalisation \citep{bishop99} as the preponderance of data in the
applications we consider produces very tight posterior distributions
over the parameters.
 
When the Gaussian process models a time series then the input
variables, $X$, are values of time.  We shall assume that increasing
input indices correspond to sequential time stamps, $x_1 \le x_2 \le
\ldots \le x_{n-1} \le x_n$.  We are at liberty to deploy GP inference
using Equations~\eqref{GP1} and~\eqref{GP2} to either {\em
  interpolate} the function $f(x_*)$ at $x_*$ when $x_1 < x_* < x_n$
or {\em extrapolate} $f(x_*)$ when either $x_* < x_1$ or $x_* > x_n$.
When measurements are obtained sequentially, extrapolation forward in
time is termed {\em prediction} and the inference of $f(x_*)$ is
termed {\em filtering}.  Interpolation with sequential measurements is
termed {\em smoothing}.  Although both smoothing and filtering
approaches have been developed for Gaussian process regression
\citep{hartikainen10}, we shall be concerned with filtering only.
However, the eigenfunction models for periodic Gaussian processes
developed in this paper can also be used for smoothing.

In the next section we review the latent force model (LFM) which is
a principled approach to incorporating solutions to differential
equations within Gaussian process inference methods.

\section{LATENT FORCE MODELS}
\label{sec:lfm} 
In this section we present a brief introduction to latent force models
and describe their practical limitations.  Specifically, we
consider dynamic processes which can be described by a set of $E$ coupled,
stochastic, linear, first order differential equations,
\begin{eqnarray*}
\frac{d z_q(t)}{dt} = \sum_{e=1}^E F_{e,q} z_e(t)+\sum_{r=1}^R L_{r,q}
u_r(t)\ ,
\end{eqnarray*}
where $q$ and $e$ index each variable $z$, $R$ is the number of
latent forces and $r$ indexes each latent force $u$, and $L$ and $F$
are coefficients of the system.  For example, in our home heating
application (described in detail in Section~\ref{sec:appthermal}),
$z_1(t)$ models the internal temperature of a home, $z_2(t)$ the
ambient temperature immediately outside the home, $u_1(t)$ is the
heater output from a known proportional controller and $u_2(t)$ is an
unknown residual force. In this application, we assume $u_2(t)$ is
periodic as it is used to model solar warming, some habitual human
behaviour and the thermal lags in the heating system.  The resulting
differential equations can be written as,
\begin{eqnarray}
\frac{d{\bf z}(t)}{d t} = {\bf F}\ {\bf z}(t)+ {\bf L} {\bf
  u}(t)\ ,
\label{initiallfm}
\end{eqnarray} 
where ${\bf u}(t)$ is a vector of $R$ independent {\em driving forces}
(also called the {\em latent forces}).  We distinguish non-periodic
latent forces, {\tt np}, and periodic latent forces, {\tt p}, as they
will be modelled differently in our approach. Non-periodic forces will
be modelled using the existing approach advocated
in \citet{hartikainen10}, which is reviewed in
Section~\ref{sec:statebasedapproaches}, and periodic forces will be
modelled using our novel linear basis approach presented in
Section~\ref{sec:LBM}.  In Equation~\eqref{initiallfm} the $E\times E$
matrix ${\bf F}$ and the $E\times R$ matrix ${\bf L}$ are non-random
coefficients that link the latent forces to the dynamic processes.
Although we deal with first order differential equations only, all
higher order differential equations can be converted to a set of
coupled first order equations \citep{hartikainen10}.

Following \citet{alvarez09} and \citet{hartikainen10,hartikainen11} we
assume that the latent forces, ${\bf u}$, are independent draws from
Gaussian processes, $u_i\sim \mathcal{GP}(0,K_i)$ where $K_i$ is the
GP covariance function \citep{rasmussen06} for force $u_i$.
Consequently, the covariance for ${\bf z}$ at any times $t$ and
$t^\prime$ can be evaluated as follows,
\begin{eqnarray}
  E[({\bf z}(t)-\bar{\bf z}(t))({\bf z}(t^\prime)-\bar{\bf z}(t^\prime))^T]= {\bf \Phi}(t_0,t) {\bf P}^0_z {\bf
    \Phi}(t_0,t^\prime)^T+ \Gamma(t_0,t,t^\prime)\ ,
\label{cov_naive}
\end{eqnarray}
where ${\bf \Phi}(t_0,t)$ denotes the matrix exponential, ${\bf
  \Phi}(t_0,t)=\exp({\bf F}(t-t_0))$ expressed in \citet{alvarez09}, 
$\bar{\bf z}(t)=E[{\bf z}(t)]$,~\footnote{All integrals in this paper
  should be interpreted as It\^{o} integrals.}
\begin{eqnarray*}
  \Gamma(t_0,t,t^\prime)=\int_{t_0}^t \int_{t_0}^{t^\prime} {\bf \Phi}(s,t)
  {\bf L} {\bf K}(s,s^\prime) {\bf L}^T {\bf
    \Phi}(s^\prime,t^\prime)^T\ ds\ ds^\prime\ ,
\end{eqnarray*}
${\bf P}^0_z$ is the state covariance at time $t_0$, ${\bf
  P}^0_z=E[({\bf z}(t_0)-\bar{\bf z}(t_0)) ({\bf z}(t_0)-\bar{\bf
  z}(t_0))^T]$ and ${\bf K}(s,s^\prime)$ is the diagonal matrix ${\bf
  K}(s,s^\prime)=\text{diag}(K_1(s,s^\prime),\ldots,K_R(s,s^\prime))$.
Since ${\bf z}(t)$ is a vector and defined for any times $t$ and
$t^\prime$ then $E[({\bf z}(t)-\bar{\bf z}(t))({\bf
  z}(t^\prime)-\bar{\bf z}(t^\prime))^T]$ is a multi-output Gaussian
process covariance function.  A kernel for covariances between the
target, ${\bf z}$, and the latent forces, ${\bf u}$, can also be
derived.  Inference with these kernels is then undertaken directly
using Equations~\eqref{GP1} and~\eqref{GP2}.

Unfortunately, a na\"ive implementation of LFM inference using
Equations~\eqref{GP1} and~\eqref{GP2} and covariance functions derived
using Equation~\eqref{cov_naive} can be computationally prohibitive.
As we have already pointed out, this approach can be computationally
expensive due to the need to invert prohibitively large covariance
matrices.  To mitigate computational intensive matrix inversion in the
GP equations, various sparse solutions have been proposed \citep[see, for
example,][]{williams01,snelson06,gredilla11} and an early review
of some of these methods is presented in \citet{candela05}.
Unfortunately, the spectral decomposition approach of
\citet{gredilla11} is sub-optimal in that it randomly assigns the
components of a sparse spectral representation and this limitation is
explored in detail in Section~\ref{sec:LBM}.  The Nystr\"om method for
approximating eigenfunctions is used in \citet{williams01} to derive a
sparse approximation for the kernel which can then be used to improve
the computational efficiency of the GP inference Equations~\eqref{GP1}
and~\eqref{GP2}.  Unfortunately, this approximate kernel is not used
consistently throughout the GP equations and this can lead incorrectly
to negative predicted variances.

The pseudo-input (also called {\em inducing inputs})
approach \citep{snelson06,candela05} is a successful method for
reducing the number of input samples used within GP inference without
significantly losing information encoded within the full data set.  In
essence, densely packed samples are summarised around sparsely
distributed inducing points.  Pseudo-inputs have been successfully
deployed within sparse approximations of dependent output Gaussian
processes \citep{alvarez08,alvarez11b}.  Pseudo-inputs have recently
been introduced to GP time-series inference and applied to problems
which exploit differential equations of the physical process via the
latent force model \citep{alvarez11}.  In \citet{alvarez11} the latent
force is expressed at pseudo-inputs and then convolved with a smooth
function to interpolate between the pseudo-inputs.  However, although
inducing inputs can reduce the sampling rate and summarise local
information, they still have to be liberally distributed over the
entire time sequence. We may assume, for simplicity, the pseudo-inputs
are evenly spread over time and, therefore, the number of
pseudo-inputs, $P$, would have to increase linearly with the number of
observations (although with a rate considerably lower than the
observation sampling rate).  Unfortunately, the computational
complexity of GP inference with pseudo-inputs is $\mathcal{O}(T P^2)$
where $T$ is the number of observations \citep{alvarez08}.  Thus,
although pseudo-inputs are able to improve the efficiency of GP
inference to some extent, for time series analysis their computational
cost is still cubic in the number of measurements and this can be
computationally prohibitive.

In the next section we describe a state-space reformulation of the
LFM.  The state-space approach has the advantage that it has a
computational complexity for inferring the target process, ${\bf z}$,
which is $\mathcal{O}(T)$ but at the expense of representing the
target process with extra {\em state} variables.

\section{STATE-SPACE APPROACHES TO LATENT FORCE INFERENCE}
\label{sec:statebasedapproaches}

In this section we review the current state-space approach to
inference with LFMs \citep{hartikainen10} and show how some covariance
functions can be represented exactly in state-space.  Unfortunately,
we shall also demonstrate that periodic kernels cannot be incorporated
into LFMs using the approach advocated by \citet{hartikainen10}. To
address this key issue, we will then propose to approximate a periodic
covariance function with a sparse linear basis model.  This will
allow us to represent periodic behaviour within a LFM efficiently and
also incorporate information encoded within the periodic kernel prior.
Our work is inspired by, and can be seen as, an extension of the
resonator model \citep{sarkka12,hartikainen12}, which is an
alternative linear basis model that allows periodic processes to be
modelled within the state-space approach.  Our LBM approach, described
in Section~\ref{sec:LBM}, builds on the resonator model and extends it
by incorporating the prior information encoded within the latent force
covariance function.

When the target processes, ${\bf z}$ as per
Equation~\eqref{initiallfm}, can be expressed in Markov form, we can
avoid the need to invert large covariance matrices and also avoid the
need to evaluate Equation~\eqref{cov_naive} over long time intervals,
$[t_0,\ t]$, by using the more efficient state-space inference
approach advocated by \citet{hartikainen10} and in this paper.  The
temporal computational complexity of the state-space approach is
$\mathcal{O}(T)$ as we integrate over short time intervals, $[t_0,\
t]$, and then reconstruct long term integrations by conflating the
local integrations via the Kalman filter.  This is an alternative
approach to that advocated by \citet{alvarez09} in which we integrate
the differential equations, as per Equation~\eqref{cov_naive}, over
long intervals, $[t_0,\ t]$, and then regress using
Equations~\eqref{GP1} and~\eqref{GP2}.  Both approaches are
mathematically equivalent in that they produce identical inferences
when they are applied to the same differential model, latent force
covariance functions and data.

The Kalman filter is a state-space tool for time series estimation
with Gaussian processes \citep{kalman60}.  The {\em Kalman smoother}
is also available for interpolation with sequential data. The Kalman
filter is a state-space inference tool which summarises all
information about the process, $f$, at time $x$ via a {\em state}
description.  The advantage of the Kalman filter is that any process
$f_*$ at any future time $x_*$ can be inferred from the current state
without any need to refer to the process history.  The state at any
time $x$ is captured by a finite set of Gaussian distributed {\em
  state variables}, ${\bf U}$, and we assume that $f$ is a linear
function of the state variables.  In \citet{hartikainen10} the state
variables corresponding to each latent force $f$ are the function $f$
and its derivatives.  In our approach the state variables
corresponding to each periodic latent force will be the eigenfunctions
of the periodic covariance function.  The key advantage of the Kalman
filter is that its computational complexity is linear in the amount of
data from a single output time-series.  Contrast this with the
standard Gaussian process approach, as per Equations~\eqref{GP1}
and~\eqref{GP2}, which require the inversion of a covariance matrix
and thus, have a computational complexity which is cubic in the amount
of data.~\footnote{The Kalman filter has a cubic computational
  complexity in the number of measured processes for multi-output
  Gaussian processes. We shall clarify the computational complexity of
  Kalman filter models for multi-output GPs in
  Section~\ref{sec:sslfm} and investigate an application of
  multi-output GPs in Section~\ref{sec:appthermal}.}

To illustrate the state-space approach consider a single non-periodic
latent force, $u_r(t)$, indexed by $r$, in
Equation~\eqref{initiallfm}.  We assume that this force is drawn from
a Gaussian process,
\begin{eqnarray*}
u_r\sim \mathcal{GP}(0,K_r)\ ,
\end{eqnarray*}
where $K_r$ is a stationary kernel.  In \citet{hartikainen10} the
authors demonstrate that a large range of stationary Gaussian process
kernels, $K_r$, representing the latent force prior can be transformed
into multivariate linear time-invariant (LTI) stochastic
differential equations of the form,
\begin{equation}\label{gp_ltisde}
\frac{d {\bf U}_r(t)}{dt} = {\bf F}_r\ {\bf U}_r(t) + {\bf W}_r\
\omega_r(t)\ ,
\end{equation} 
where ${\bf U}_r(t) = (u_r(t),\ \frac{du_r(t)}{dt},\ \cdots , 
\frac{d^{p_r-1}u_r(t)}{dt^{p_r-1}})^T$ and, 
\begin{equation}
{\bf F}_r = \begin{pmatrix}
    0 & 1 & & \\
     & \ddots  & \ddots & \\
     &  & 0 & 1\\
    -c_{r}^0 & \cdots & -c_{r}^{p_r-2} & -c_r^{p_r-1}\\
\end{pmatrix}\ ,\hspace*{3cm} 
{\bf W}_r = \begin{pmatrix}
    0 \\
    \vdots \\
    0\\
    1\\
  \end{pmatrix}\ ,\label{hartstoch}
\end{equation}
where $c$ are coefficients which can be set using spectral
analysis of the kernel as per \citet{hartikainen10}.  The force,
$u_r(t)$, can be recovered from ${\bf U}_r(t)$ using the indicator
vector ${\bf \Delta}_r=(1,0,\ldots,0)$ where,
\begin{eqnarray*}
u_r(t)={\bf \Delta}_r{\bf U}_r(t)\ .
\end{eqnarray*}
By choosing the coefficients $c_{r}^0,\ldots,c_r^{p_r-1}$ in
Equation~\eqref{hartstoch}, the spectral density of the white noise
process $\omega_{r}(t)$ in Equation~\eqref{gp_ltisde} and the
dimensionality $p_r$ of ${\bf U}_r(t)$ appropriately, the covariance
of $u_r(t)$, corresponding to the dynamic model, can be chosen to
correspond to the GP prior $K_r$.  The differential equations
expressed in Equation~\eqref{gp_ltisde} can then be integrated into
the LFM to form the augmented dynamic model expressed later in
Equation~\eqref{diffmodel}.  The coefficients
$c_{r}^0,\ldots,c_r^{p_r-1}$ are found by initially taking the Fourier
transform of both sides of Equation~\eqref{gp_ltisde}.  The
coefficients can then be expressed in terms of the spectral density of
the latent force kernel, $K_r$, provided that its spectral density,
$S_r(\varpi)$, can be written as a rational function of
$\varpi^2$,
\begin{align} 
S_r(\varpi) = \frac{\text{(constant)}}{\text{(polynomial in $\varpi^2$)}}\
.
\label{powerspec}
\end{align} 
The {\em inverse} power spectrum is then approximated by a polynomial
series from which the transfer function of an equivalent stable Markov
process for the kernel can be inferred along with the corresponding
spectral density of the white noise process.  The stochastic
differential equation coefficients are then calculated from the
transfer function.  For example, for the first-order Mat\'ern kernel,
\begin{eqnarray}
K_r(t,t^\prime)=\sigma_r^2\exp\left(-\frac{|t-t^\prime|}{l_r}\right)\ ,
\label{firstordermat}
\end{eqnarray}
with output scale $\sigma_r$ and input scale $l_r$,
$u_r\sim\mathcal{GP}(0,K_r)$ can be represented by
Equation~\eqref{gp_ltisde} with ${\bf U}_r(t)=u_r$, ${\bf W}_r=1$ and,
\begin{eqnarray}
{\bf F}_r=-1/l_r\ .
\label{firstordermat1}
\end{eqnarray}
The spectral density, $\lambda_r$, of the white noise process, $\omega_r$, is,
\begin{eqnarray}
\lambda_r=\frac{2\sigma_r^2\sqrt{\pi}}{l_r\ \Gamma(0.5)}\ ,
\label{firstordermat2}
\end{eqnarray}
and $\Gamma$ is the Gamma function \citep{hartikainen10}.

Now, by augmenting the state vector, ${\bf z}$ in
Equation~\eqref{initiallfm} appropriately with the non-periodic forces
${\bf U}_r(t)$ and their derivatives \citet{hartikainen11}
demonstrate that the dynamic equation can be rewritten as a joint
stochastic differential model,
\begin{eqnarray}
\frac{d{\bf z}_a(t)}{dt}= {\bf F}_a\ {\bf z}_a(t) + {\bf
  L}_a\omega_a(t)\ ,
\label{diffmodel}
\end{eqnarray} 
where 
\begin{eqnarray*}
{\bf z}_a(t)&=&({\bf z}(t),\  {\bf U}_1(t),\ \ldots\ ,\ {\bf U}_R(t))^T\ , \\
{\bf F}_a&=& \begin{pmatrix} {\bf F} & {\bf L} {\bf S}_1 {\bf \Delta}_1 & \ldots &
  {\bf L}{\bf S}_R {\bf \Delta}_R\\
                               {\bf 0} & {\bf F}_1 & \ldots
                               & {\bf 0}\\
                              & & \vdots & \\
                             {\bf 0} & {\bf 0} & \ldots & {\bf
                               F}_R
\end{pmatrix}\ ,
\end{eqnarray*}
$R$ is the number of latent forces, ${\bf S}_r=(0,\ldots,1,\ldots,0)$
is the indicator vector which extracts the $r$th column of ${\bf L}$
corresponding to the $r$th force, $u_r$, and $\omega_a(t)$ is the
appropriate scalar process noise,
\begin{eqnarray}
{\bf \omega}_a(t) &=& (0,\ \omega_1(t),\ \ldots\ ,\ \omega_R)^T\ ,\label{omega}\\
{\bf L}_a &=& \text{blockdiag}({\bf 0},\ {\bf
  W}_1,\ \ldots\ ,{\bf W}_R)\ . \label{L}
\end{eqnarray}
 These differential equations have the solution,
\begin{eqnarray*} 
{\bf z}_a(t)= {\bf \Phi}(t_0,t){\bf z}_a(t_0)+{\bf q}_a(t_0,t)\ ,
\end{eqnarray*} 
where, again, ${\bf \Phi}(t_0,t)$ denotes the matrix exponential, ${\bf
  \Phi}(t_0,t)=\exp({\bf F}_a(t-t_0))$ expressed in \citet{alvarez09}.
The process noise vector, ${\bf q}_a(t_0,t)$, is required to
accommodate the Mat\'ern or SE latent forces within the discrete time
dynamic model, ${\bf q}_a(t_0,t)\sim \mathcal{N}({\bf 0},{\bf Q}_a(t_0,t))$
where,
\begin{eqnarray*}
{\bf Q}_a(t_0,t)=\int_{t_0}^t {\bf \Phi}(s,t) {\bf L}_a \Lambda_a
{\bf L}_a^T{\bf \Phi}(s,t)^T ds\ ,
\end{eqnarray*}
and $\Lambda_a$ is a diagonal matrix,
\begin{eqnarray}
\Lambda_a=\text{diag}(0,\lambda_1,\ldots,\lambda_R)\ ,
\label{Lambda}
\end{eqnarray}
where $\lambda_r$ is the spectral density of the white noise process
corresponding to the Mat\'ern or SE process, $K_r$ \citep{hartikainen10}.

\begin{figure}[ht]
\begin{center}
\includegraphics[width=\textwidth]{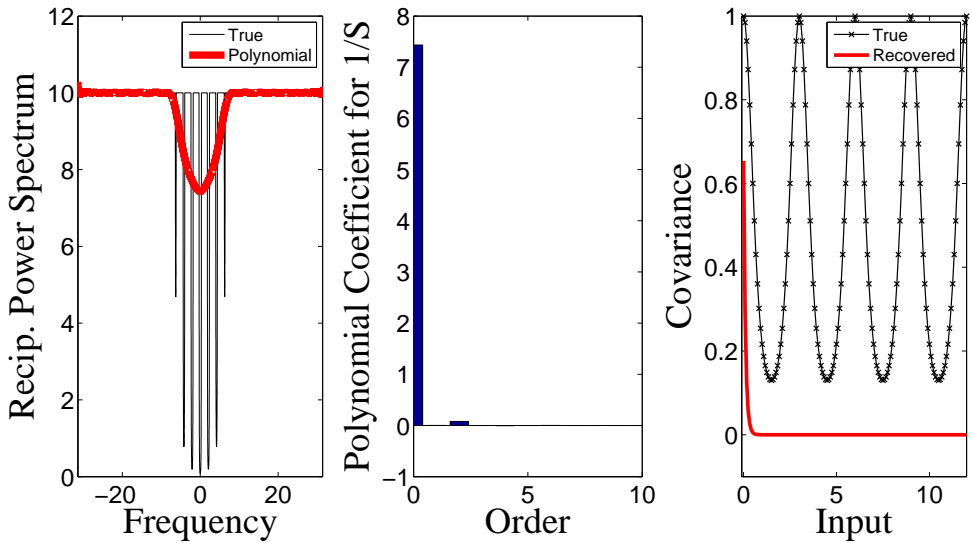}
\end{center}
\caption{\label{powerspectrum} Spectral analysis of a periodic
  covariance function.  The left panel shows inverse power spectrum
  for a periodic squared-exponential kernel (thin line) and its
  polynomial approximation (thick line).  The central panel shows the
  coefficients of the polynomial approximation.  The right panel shows
  the true covariance function (crossed line) and its approximation
  (solid line) recovered from the polynomial representation of the
  inverse power spectrum.}
\end{figure}

We now briefly describe the reasons why this spectral analysis
approach advocated by \citet{hartikainen10,hartikainen11} cannot be
immediately applied to periodic kernels.  For illustrative purposes we
shall investigate the commonly used squared-exponential periodic
kernel,
\begin{eqnarray}
K_\text{SE}(t,t^\prime)=\exp\left(-\frac{\sin\left(\frac{\pi(t-t^\prime)}{D}\right)^2}{l^2}\right)\ ,
\label{periodicsquaredexp}
\end{eqnarray}
with input scale $l=3$, an implicit output scale of unity and period
$D=0.7$, although our analysis and conclusions apply to all periodic
kernels, in general.  Unfortunately, as is shown in the left panel of
Figure~\ref{powerspectrum}, the power spectrum for this periodic
kernel is a weighted sum of Dirac delta functions, each delta function
identifying a sinusoidal mode. The inverse of the power spectrum is
highly nonlinear and not amenable to the polynomial series
approximations expressed in Equation~\eqref{powerspec}.  The left
panel also shows the best (in a least squares sense) polynomial fit to
the inverse spectrum.  The polynomial coefficients are shown in the
central panel and very little weight is assigned to higher order
frequencies.  Now, using the approach advocated in
\citet{hartikainen10} we can infer the covariance function
corresponding to this polynomial approximation of the inverse
spectrum.  This covariance function and the true periodic covariance
function are shown in the right panel of Figure~\ref{powerspectrum}.
It is clear that the covariance function obtained using
\citet{hartikainen10} is a poor representation of the true periodic
kernel.

So, it is not possible to formulate all periodic latent forces via
Equation~\eqref{gp_ltisde}.  However, by approximating the latent
force as a linear sum of basis functions, such that each basis
function, $\phi$, can be formulated via Equation~\eqref{gp_ltisde},
\begin{eqnarray}
u_r(t)=\sum_j a_{rj} \phi_j(t)\ ,
\label{basicLBM}
\end{eqnarray}
then it is possible to represent the periodic latent force within the
KF. In essence, the latent force, $u_r$, is decomposed into a weighted
sum of basis latent forces, $\{\phi_j\}$, such that each $\phi$
satisfies Equation~\eqref{gp_ltisde}.  This is the approach of
\citet{sarkka12} for representing both stationary and quasi-periodic
latent forces via their {\em resonator model}.  In
\citet{hartikainen12}, the {\em resonator}, $\phi_r$, is chosen to be
a Fourier basis, $\phi_r(t)=\cos(f_r t)$ or $\phi_r(t)=\sin(f_r t)$.
The resonator can be represented by Equation~\eqref{gp_ltisde} as a
state comprising the instantaneous resonator value, $\phi_r(t)$, and
its derivative, $\dot{\phi}_r(t)$ thus, ${\bf U}_r(t)=(\phi_r(t),\
\dot{\phi}_r(t))^T$.  The corresponding SDE has ${\bf F}_r=[0 \ \ 1\
;\ -f_r^2 \ \ 0]$ and ${\bf W}_r=0$.  The Fourier basis is
particularly useful for modelling stationary covariance
functions.  

In \citet{hartikainen12} quasi-periodic latent forces were implemented
as a superposition,
\begin{eqnarray}  
u(t)=\sum_{j} \psi_j(t)\ , 
\label{lbmres}
\end{eqnarray}    
of resonators, $\psi$, of the form,
\begin{eqnarray}
\frac{d^2 \psi_j(t)}{d t^2} = -(2\pi f_j(t))^2 \psi_j(t)+\omega_j(t)\ ,
\label{resonator1}
\end{eqnarray}  
where $\omega$ is a white noise component. Crucially, the resonator
{\em frequencies}, $f$, are time variant and this supports
non-stationary and quasi-periodic forces.  This model is very flexible
and both periodic and quasi-periodic processes can be expressed using
the resonator model (as detailed in Appendix~\ref{app:eigres}).
However, currently no mechanism has been proposed to incorporate prior
information encoded in periodic GP kernels within the resonator model.
Further, inferring the parameters and the frequency profiles, $f(t)$,
for each resonator can be prohibitively computationally expensive (as
we demonstrate in Appendix~\ref{app:eigres}).  Despite these
shortcomings there is a very close connection between the resonator
model for periodic latent forces and the eigenfunction approach
proposed in this paper.  This connection is explored in detail in
Appendix~\ref{app:eigres} in which we assert that the eigenfunction
basis is an instance of the resonator basis for perfectly periodic
covariance functions.  We subsequently demonstrate how the
eigenfunction approach can both inform the resonator model of the GP
prior and also simplify the inference of the resonator model
parameters including the frequency profile. Further, we show that the
optimal minimum mean-square resonator model is an alternative way of
representing the corresponding eigenfunction basis within the Kalman
filter.

In the original implementation of the resonator model \citep{sarkka12}
the model parameters were set by hand.  Recently, a new variation of
the resonator model has been proposed in which the most likely model
parameters are learned from the data \citep{solin13}.  In this version
the resonator is the solution to the time invariant second order
differential equation,
\begin{eqnarray}
\frac{d^2 \psi_j(t)}{d t^2} =  A_j \psi_j(t)+ B_j\frac{d\psi_j(t)}{dt}+\omega_j(t)\ ,
\label{resonator2}
\end{eqnarray} 
where $A$ and $B$ are constant coefficients.  We note that this
variation of the resonator model is a special case of the original
resonator model with a frequency profile
$f_j(t)=\frac{i}{2\pi}\sqrt{A_j+B_j\frac{1}{\psi_j(t)}\frac{d\psi_j(t)}{dt}}$
in Equation~\eqref{resonator1}. To model quasi-periodic processes
Equation~\eqref{resonator2} comprises a {\em decay} term via the first
derivative of the resonator function. This new model is
computationally efficient as it imposes constant coefficients unlike
the original resonator model in \citet{sarkka12}.  However, the
computational efficiency of the model in Equation~\eqref{resonator2},
gained by losing the requirement to infer a frequency profile for each
resonator, is at the expense of the model's flexibility.  We compare
the resonator model in Equation~\eqref{resonator2} with our
eigenfunction approach on a real world application in
Section~\ref{sec:appthermal}.

In preparation for the approach advocated in this paper, in which we
also represent the periodic kernel via a linear basis model, the
following section compares the two key alternative approaches to
directly inferring linear basis models from Gaussian Process kernels,
namely the sparse spectrum Gaussian process regression \citep[SSGPR,]
[]{gredilla11} and kernel principal component analysis
\citep{scholkopf98}.

\section{REPRESENTING PERIODIC LATENT FORCES WITH LINEAR BASIS MODELS}
\label{sec:LBM}   
In this section, we exploit linear basis models and propose a novel
approach to representing periodic latent force GP kernels. Our aim is
to derive a sparse representation for periodic kernels so that they
can be accommodated within a state-space formulation of the LFM.
Linear basis models (LBMs) have a long history in machine learning. In
particular, special cases of them include kernel density estimators
\citep{parzen62} and the Relevance Vector Machine \citep{tipping01}.
There are two key advantages to representing periodic kernels using a
sparse basis model: firstly, they can approximate the kernel using a
weighted sum over a finite set of functions.  As we will see, for
relatively smooth kernels the number of basis functions can be small.
The second advantage, as we will show in Section~\ref{sec:sslfm}, is
that the LBM representation is amenable to inference using
computationally efficient state-space methods.  We exploit the
Nystr\"om approximation as opposed to other sparse approximations
(such as the sparse spectrum Gaussian process regression (SSGPR)
method of \citet{gredilla11}) as, we will see, the eigenfunctions of
the kernel form the most efficient basis for the corresponding driving
forces.  This approximation will accommodate both the prior
information about the driving forces (encoded in the kernel) within a
state-space approach and also provide a means to learn these driving
forces from data using iterative state-space methods.  Approximating
Gaussian process priors via the Nystr\"om method is not new \citep[see, for
example,][]{williams01}.  However, using this to accommodate
periodic and quasi-periodic latent forces within LFMs is novel.

In order to develop our LBM for latent forces we shall first
investigate current approaches to sparse representations of stationary
covariance functions and then demonstrate that one of these
approaches, namely the eigenfunction approach, generalises to
non-stationary covariance functions.  Bochner's theorem asserts that
all stationary covariance functions can be expressed as the Fourier
transform of their corresponding spectral densities \citep[where the
spectral density exists.  See, for example,][]{rasmussen06}.
Furthermore, in the stationary case, the Fourier basis is the
eigenfunctions of the covariance function. There has been a long
history of research into the spectral analysis of stationary Gaussian
process kernels \citep[see, for example,][]{bengio04}.  However, only
recently has the Fourier basis been investigated in the context of
latent force models.
To date, two approaches have been proposed to incorporate knowledge of
all stationary kernels, including periodic kernels, within the linear
basis representation via spectral analysis: the SSGPR
\citep{gredilla11} and the KPCA \citep{drineas} method.  The key
advantage of these approaches is that the basis frequencies can be
calculated from the prior latent force kernel.  These approaches are
described and compared next.

\begin{figure}[ht]
\begin{center}
\includegraphics[width=0.7\textwidth]{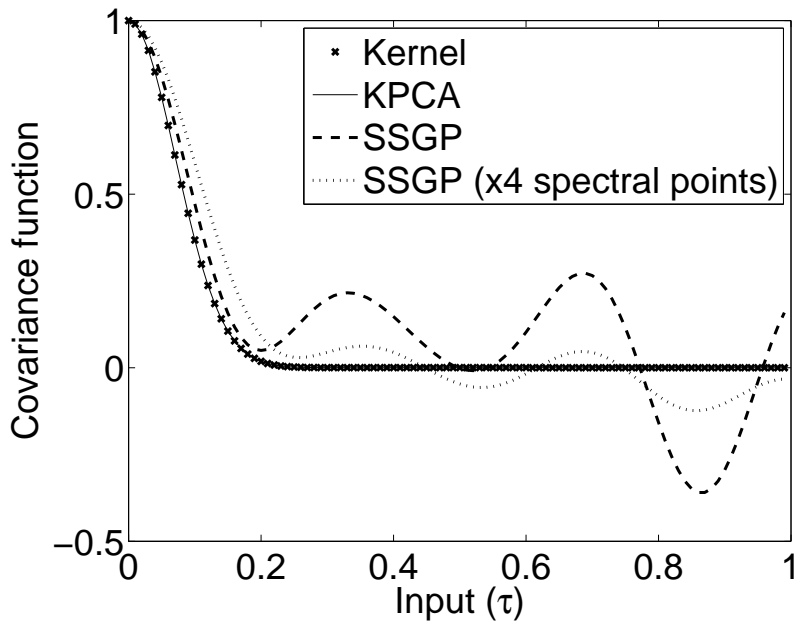}
\end{center}
\caption{\label{ssgp} A comparison of SSGPR and eigenfunction approaches to
  modelling GP kernels via basis functions. The plots show the covariance
  functions corresponding to each of the eigenfunction and SSGPR models.}
\end{figure}

The SSGPR \citep{gredilla11} approach reinterprets the spectral
density of a stationary GP kernel as the probability density function
over frequency space.  This pdf is then sampled using Monte Carlo to
yield the frequencies of the sinusoidal basis functions of the
LBM.~\footnote{In their code, available at {\tt
    http://www.tsc.uc3m.es/$\sim$miguel/downloads.php}, the authors try
  several frequency initialisations and use the best one.}  The
advantage of this approach is that a sparse set of sinusoidal basis
functions is identified such that the most significant frequencies of
these sinusoidal basis functions have the greatest probability of being chosen.
The phase of each basis function is then inferred from the data.  The
disadvantage of this approach is it can often provide a poor
approximation to the covariance function as we will demonstrate
shortly in Figure~\ref{ssgp}.

An alternative approach to the SSGPR is KPCA which effectively
intelligently samples the most informative frequencies within the
spectral density.  Mercer's theorem \citep{mercer09} allows us to
represent each periodic latent force, $u(t)$, at arbitrary inputs,
$t$, via an infinite set of basis functions, $\phi_j$,
\begin{eqnarray}    
u(t)=\sum_{j=1}^\infty a_j\phi_j(t)\ ,
\label{lbm}
\end{eqnarray}      
where $\{a_j\}$ are the model {\em weights} which are independently 
drawn from a Gaussian, 
\begin{eqnarray} 
a_j\sim \mathcal{N}(0,\mu^\phi_j)\ ,
\label{lbma} 
\end{eqnarray} 
where $\mu^\phi_j$ is the variance of $a_j$.  For any choice of probability
density function, $p$, there exists an orthonormal basis, $\{\phi\}$,
such that,
\begin{eqnarray*}
\int \phi_i(t) \phi_j(t) p(t) dt=
\begin{cases}
1 & \text{\ if \ } i=j\ ,\\
0 & \text{\ otherwise}\ .
\end{cases}
\end{eqnarray*}
Furthermore, the latent force prior, $K(t,t') =E[u(t)u(t^\prime)]$,
can be expressed as,
\begin{eqnarray}
K(t,t')=\sum_{j=1}^\infty \mu^\phi_j \phi_j(t) \phi_j(t^\prime)\ ,
\label{eigenexact}
\end{eqnarray}
where, $\phi_j$ are the {\em eigenfunctions} of the kernel, $K$, under
$p$ such that,
\begin{eqnarray}
\int K(t,t^\prime)\phi_j(t^\prime) p(t^\prime) dt^\prime=\mu^\phi_j
\phi_j(t)\ ,
\label{eigeneqn}
\end{eqnarray}
and the variance, $\mu^\phi_j$, is also an {\em eigenvalue} of the kernel.

Of course, it is not feasible to actually use an infinite basis.
Thus, we approximate the infinite sum in Equation~\eqref{lbm} by
a finite sum over a subset of significant eigenfunctions which have
the $J$ most significant eigenvalues, $\mu^\phi$,
\begin{eqnarray}
u(t)\approx \sum_{j=1}^J a_j \phi_j(t)\ .
\label{sparseexact}
\end{eqnarray}
Fortunately, kernel principal component analysis (KPCA) allows us to
identify the most significant $J$ eigenfunctions a priori as well as
compute their form approximately \citep{scholkopf98}. 

The role of $p$, in Equation~\eqref{eigeneqn}, is to weight the values
of time $t$.  We are free to choose the probability density function,
$p(t)$, as we wish.  For stationary covariance functions, a uniform
pdf is appropriate as it weights each time instance, $t$, equally.  To
evaluate the integral in Equation~\eqref{eigeneqn} we use a
quadrature-based method and $N$ equally spaced quadrature points, $S$, of
$t$, where $S=\{s_1,\ldots,s_N\}$ \citep[see, for example,][]{shawe05},
\begin{eqnarray}
\int K(t,t^\prime)\phi_j(t^\prime) p(t^\prime)
dt^\prime\approx\frac{1}{N}\sum_{i=1}^N K(t,s_i)\phi_j(s_i)\ .
\label{quadrature}
\end{eqnarray}
The points, $S$, are also used to construct an $N\times N$ covariance
matrix, $G$, called the {\em Gram matrix}, where,
\begin{eqnarray}
G_{ij}=K(s_i,s_j)\ .  
\label{gramG}
\end{eqnarray}
The Nystr\"om approach is then used to derive approximate eigenfunctions
of $K$ using the eigenvectors, ${\bf v}$, and eigenvalues, $\mu$, of
the Gram matrix \citep{drineas}. We denote the Nystr\"om approximation
for $\phi_j$ with uniform pdf $p$ as $\tilde{\phi}_j$.  For each
eigenvector, ${\bf v}_j$,
 \begin{eqnarray}
\tilde{\phi}_j(t)=\frac{\sqrt{N}}{\mu_j} K(t,S) {\bf v}_j\ .
\label{phi}
\end{eqnarray}  
Since $\{{\bf v}_j\}$ are orthonormal then $\{\tilde{\phi}_j\}$ are
orthogonal.  Now, substituting the approximation for $\phi$ into
Equation~\eqref{sparseexact},
\begin{eqnarray}
u(t)\approx \sum_{j=1}^J a_j \tilde{\phi}_j(t)\ .
\label{sparse}
\end{eqnarray}
By forming the covariance between $u(t)$ and $u(t^\prime)$ we can
derive the following relationship between the latent force prior, the
approximate eigenfunctions and the variances $\mu^\phi_j$ of the model
weights, $a_j$,
\begin{eqnarray}
K(t,t^\prime)\approx \sum_{j=1}^J \mu^\phi_j \tilde{\phi}_j(t)
\tilde{\phi}_j(t^\prime)\ ,
\label{approxkern}
\end{eqnarray}
where $\mu^\phi_j\approx\mu_j/N$, is the scaled Gram matrix eigenvalue
\citep{williams01}.  

As we can compare the covariance function, $K$, with the corresponding
Nystr\"om covariance function approximation, as per
Equation~\eqref{approxkern}, then the sample set, $S$, can be chosen a
priori to provide a comprehensive representation of the kernel $K$.
Furthermore, as $N\rightarrow \infty$ then
$\tilde{\phi}_j\rightarrow\phi_j$.  Finally, although the
eigenfunction LBM is a parametric model, the eigenfunctions accurately
reproduce the periodic GP prior across an entire period and
undesirable extrapolation errors often associated with spatially
degenerate LBMs are alleviated here \citep{rasmussen06}.

Throughout this paper the LBMs will comprise the most significant
eigenfunctions according to the following definition, 
\begin{definition}\label{sigdef}
  An eigenfunction is {\em significant} if its eigenvalue is more than
  a pre-defined fraction $\gamma$ of the maximum eigenvalue.
\end{definition}
We have found that $\gamma=1/100$ is a robust choice for the
applications in Sections~\ref{sec:appqueue} and~\ref{sec:appthermal}
in which fewer than $30$ basis functions are required to model the
latent forces.
\\
\begin{figure}[ht]
\begin{center}
\hspace*{-0.5cm}\begin{tabular}{cc}
\includegraphics[width=0.5\textwidth]{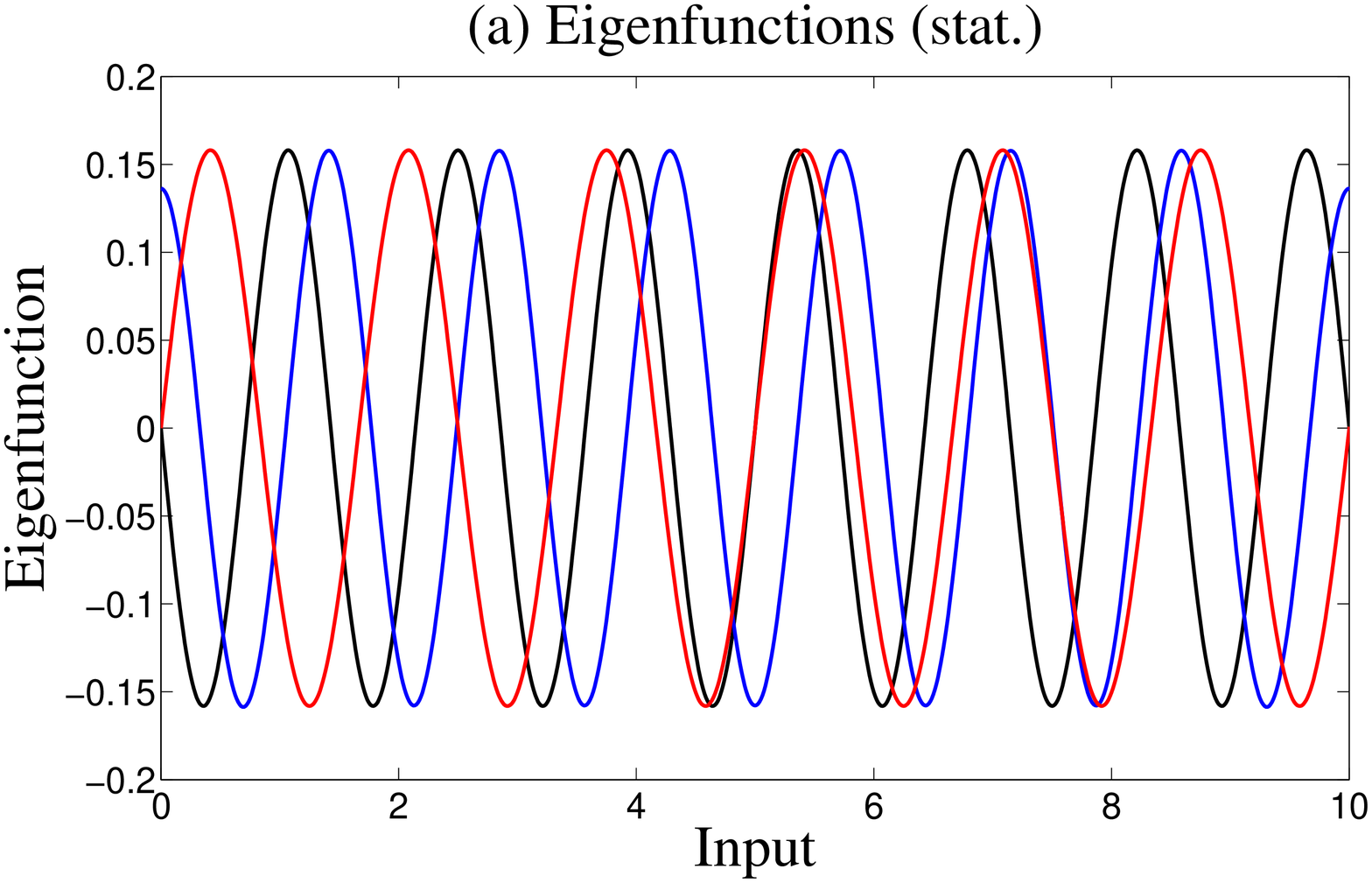}&
\hspace{-0.5cm}
\includegraphics[width=0.5\textwidth]{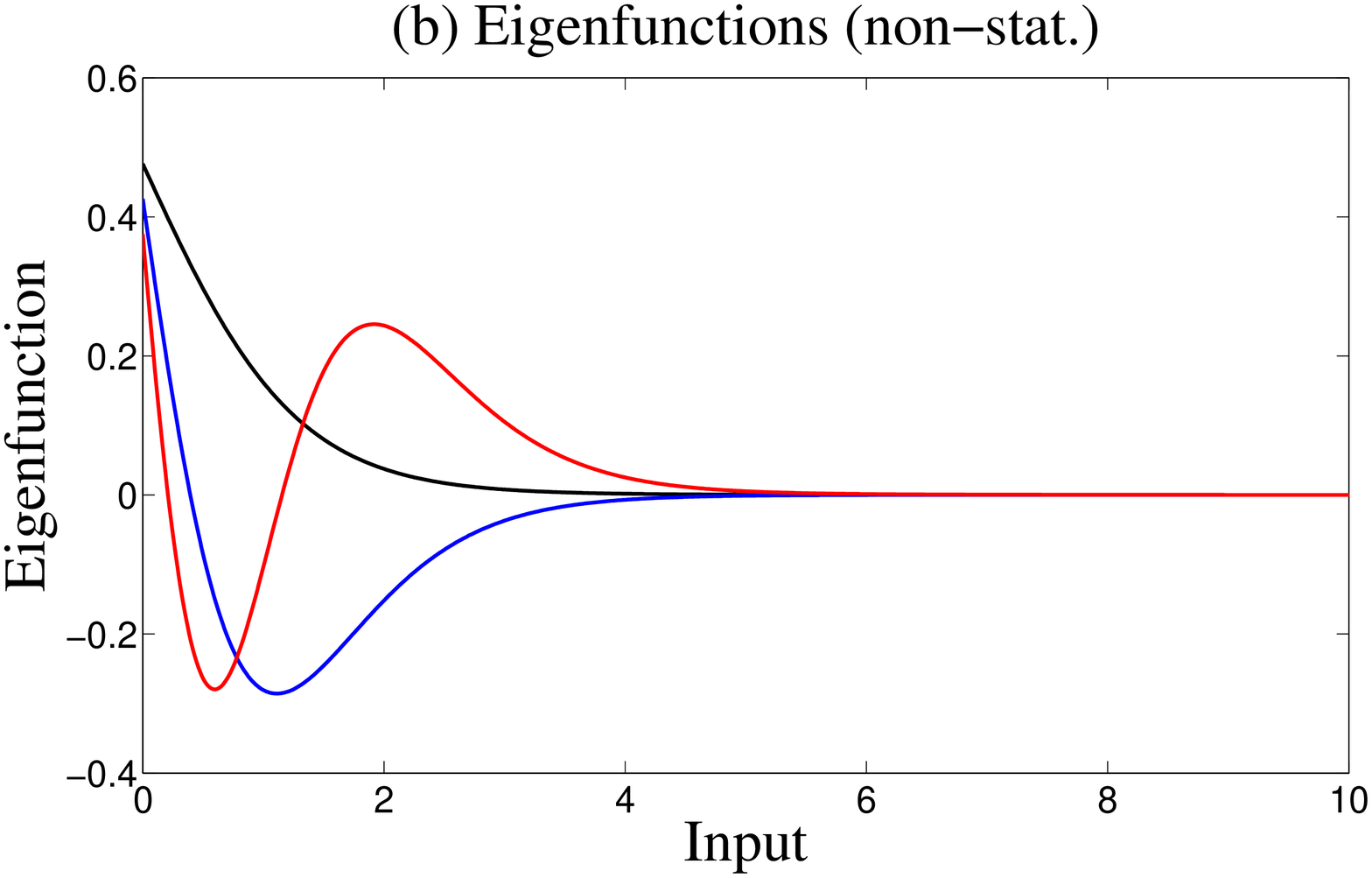}\\
\includegraphics[width=0.5\textwidth]{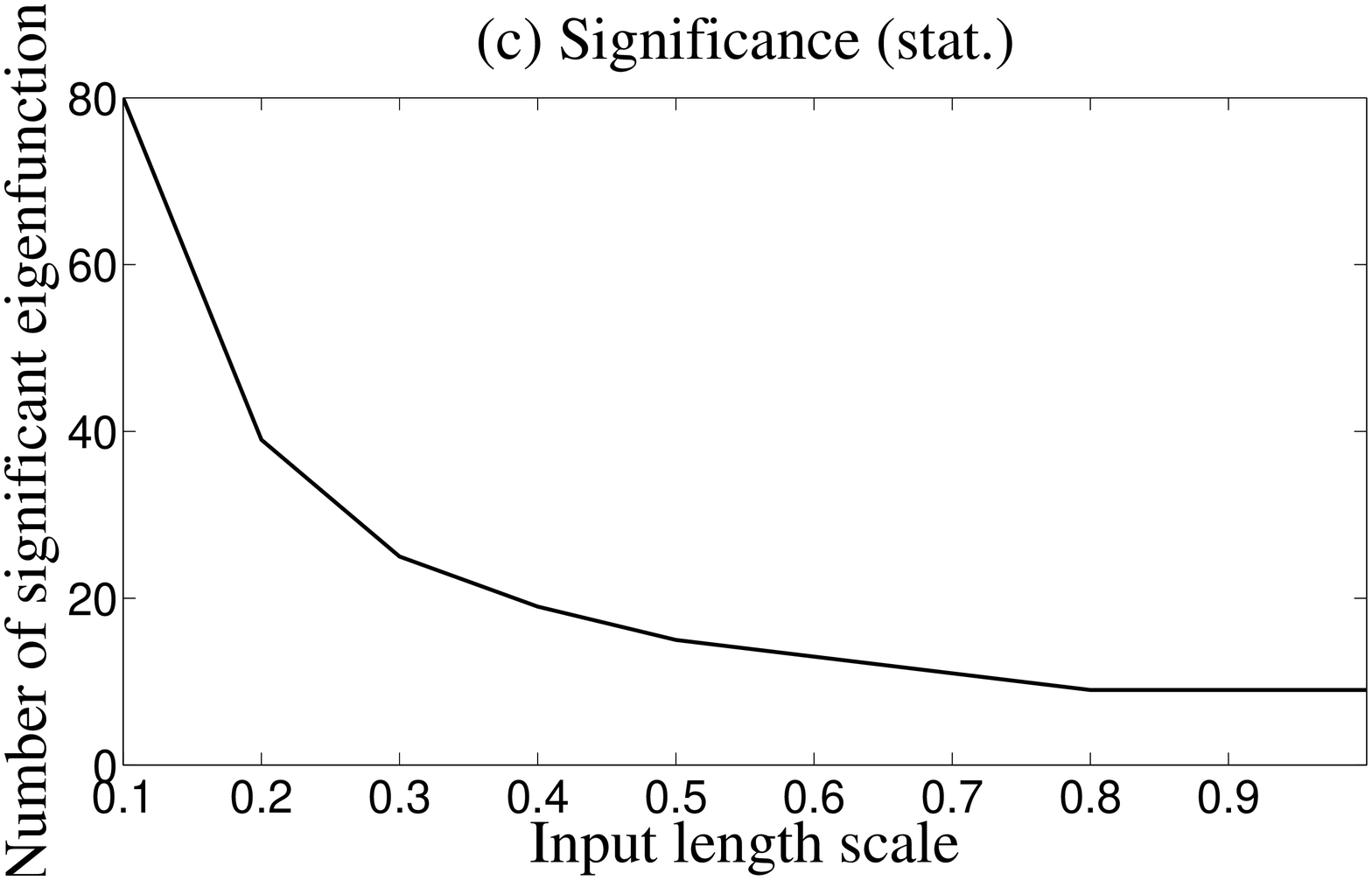}&
\hspace{-0.5cm}
\includegraphics[width=0.5\textwidth]{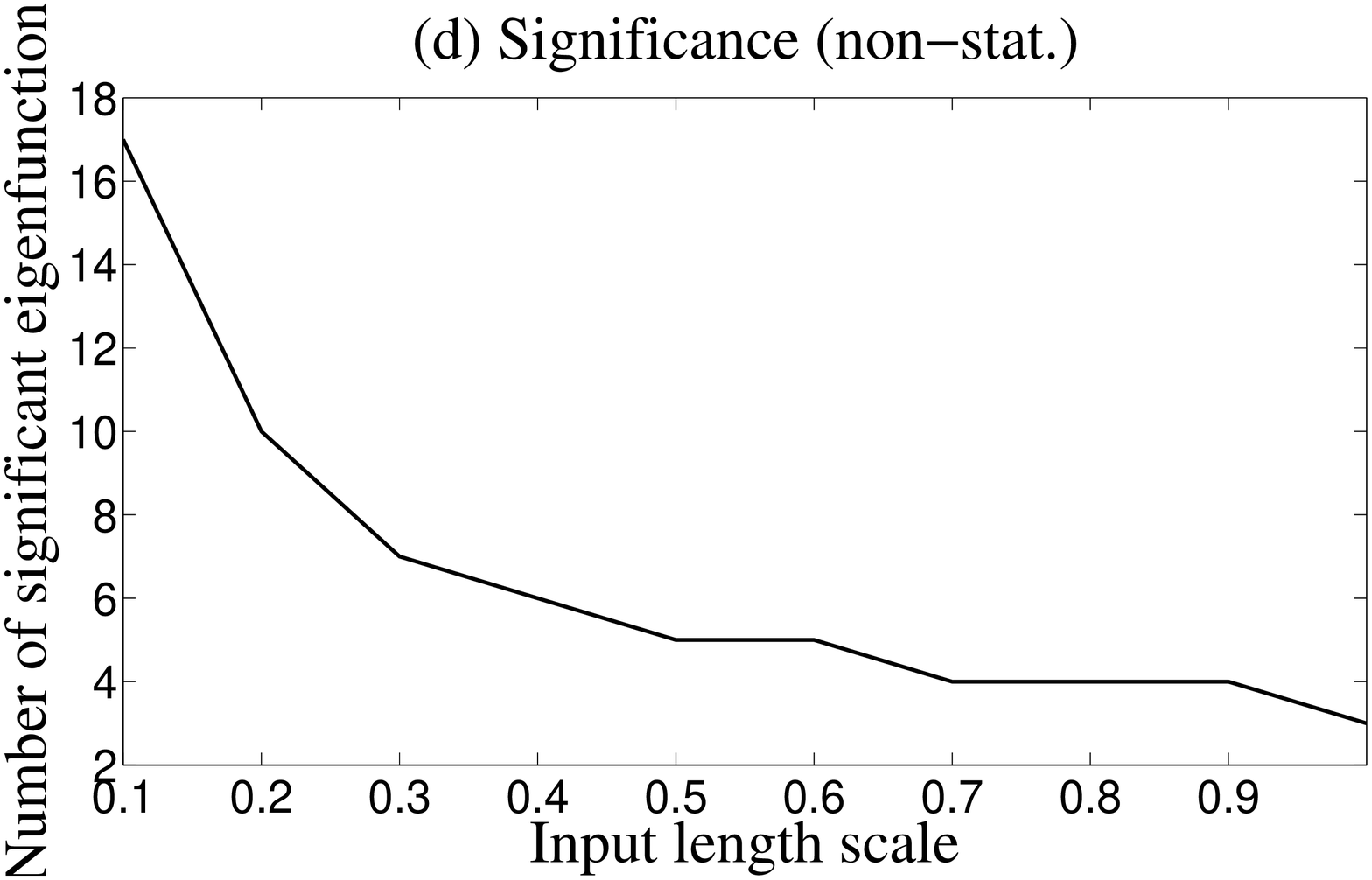}
\end{tabular}
\end{center}
\caption{\label{sparseK} Example eigenfunctions for (a) stationary
  periodic and (b) non-stationary covariance functions, both with
  period 10 units.  Also, the number of significant eigenfunctions for
  input length scales, $l$, for the (c) stationary periodic and (d)
  non-stationary covariance function.}
\end{figure}

To demonstrate the eigenfunction approach to representing Gaussian
process priors via a finite basis, Figure~\ref{sparseK}(a) shows
example eigenfunctions for a stationary periodic Mat\'ern process.
The Mat\'ern kernel is defined \citep{rasmussen06},
\begin{eqnarray}  \label{matern}
\text{Mat\'ern}(\tau,\nu,\sigma,l)=\sigma^2
\frac{2^{1-\nu}}{\Gamma(\nu)}\left(\frac{\sqrt{2\nu}}{l}\tau\right)^\nu
\check{K}_\nu\left(\frac{\sqrt{2\nu}}{l}\tau\right)\ ,
\end{eqnarray} 
where $\tau\ge 0$, $\Gamma$ and $\check{K}_\nu$ are the gamma and
modified Bessel functions, respectively, $\nu$ indicates the order,
$\sigma$ is the output scale which governs the amplitude of the kernel
and $l$ is the input length scale which governs the smoothness of the
kernel.  When the target function is periodic it is a direct function
of the period {\em phase}, $\kappa(\tau)=|\sin(\pi \tau /D)|$ where
$D$ is the function period.  Consequently, the periodic Mat\'ern is
given by $\tt{\text{Mat\'ern}(\kappa(\tau),\nu,\sigma,l)}$.  The
periodic Mat\'ern is of particular interest to us as it is used in
Section~\ref{sec:appqueue} to model customer call centre arrival rates
and in Section~\ref{sec:appthermal} to model the residual dynamics
within home heating.

We observe that the eigenfunctions of the periodic kernel are the
sinusoidal basis functions as shown in Figure~\ref{sparseK}. This
basis corresponds to the Fourier basis functions for the power
spectrum that can be obtained by Fourier analysis of the kernel.
However, although through Fourier analysis we would be able to
determine the power spectrum of the covariance function, and
consequently the magnitude of the basis function, we would be unable
to determine the phase of the basis function. KPCA, in contrast, is
able to determine both the magnitude, and consequently phase, of the
Fourier basis functions.  

A key property of the KPCA approach is that the eigenfunctions are not
limited to the Fourier basis and, consequently, KPCA is also able to
model non-stationary periodic covariance functions efficiently, in
which case the eigenfunctions, which are inferred using KPCA from the
non-stationary covariance function, are anharmonic as we will now
demonstrate.  Figure~\ref{sparseK}(b) shows the first three most
significant eigenfunctions for an exponentially moderated periodic
kernel,
\begin{eqnarray}
K(t,t^\prime)=\text{Mat\'ern}(\kappa(t-t^\prime),\nu,\sigma,l)\exp\left(-|t|-|t^\prime|\right)\ 
.
\label{nonstatex}
\end{eqnarray} 
Figure~\ref{sparseK}, panels (c) and (d) show how the number of
significant eigenfunctions decreases with increasing kernel smoothness
for both the harmonic and anharmonic kernels above.  The smoothness of
the kernel is parameterised by the phase length scale, $l$.  As above,
we choose to declare an eigenfunction as significant if its eigenvalue
is more than one hundredth of the maximum eigenvalue.  Although this
is a conservative definition of significance we can see that only a
small number of basis functions are required to model these kernels.

We now compare the SSGPR, described above, and the eigenfunction
approaches to modelling stationary kernels.  For stationary kernels
both the SSGPR and eigenfunction methods use a linear basis model with
sinusoidal basis functions.  The only difference between the
approaches is that SSGPR assigns basis function frequencies (called
{\em spectral points}) by sampling the kernel power spectrum.  Both
sine and cosine functions are used for each frequency.  The KPCA
infers its frequencies deterministically from the kernel and uses the
basis functions with the most significant eigenvalues. Each spectral
point corresponds to a Fourier basis function with known frequency
with indeterminate phase.  So, $S$ spectral points produce $S$ Fourier
basis functions which has the same complexity as $S$ Fourier basis
functions in the eigenfunction approach. We compare the efficacy of
both linear basis approaches when representing the squared-exponential
kernel.  The SSGPR was specifically developed with this kernel in mind
and thus we present the fairest comparison.  In order to investigate
this difference and isolate the inference procedure by which the GP
hyperparameters are learned from the data, the SSGPR algorithm is
changed only slightly so that the actual kernel hyperparameters used
correspond to the actual hyperparameters of the model which generated
the training data.  We also use the known generative GP
hyperparameters within the eigenfunction model.

\begin{figure}[ht]
\begin{center}
\includegraphics[width=0.75\textwidth]{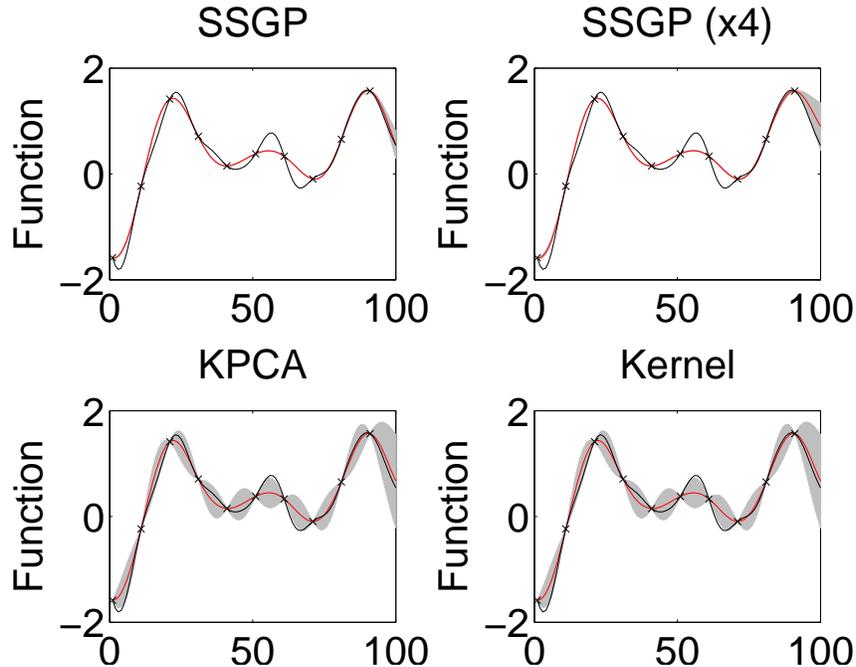}
\end{center}
\caption{\label{ssgp_example} A comparison of SSGPR and eigenfunction
  approaches to modelling GP kernels via basis functions.  The plots
  show typical example function estimates drawn using both approaches.
  The KPCA uses 22 basis functions and the SSGPR uses 22 spectral
  points and 88 spectral points respectively.  The grey regions are
  the first standard deviation confidence regions.}
\end{figure}

To demonstrate the superiority of the eigenfunction approach over the
SSGPR approach, Figures~\ref{ssgp} and~\ref{ssgp_example} compare the
SSGPR and eigenfunction representations of a squared-exponential
kernel with an input scale of $10$ units and output scale of $1$ unit.
The significant twenty two eigenfunctions were used and, equivalently,
twenty two SSGPR spectral points were randomly chosen from the SE
spectral density as proposed by \citet{gredilla11}.  Further, the
eigenfunction approach used $20$ evenly spaced points to construct the
Gram matrix.  In the case of the KPCA the corresponding covariance
functions differed by no more than $9.6\times10^{-5}$ from the actual
covariance function.  The SSGPR, using the same number of Fourier
basis functions, deviated by as much as $0.36$ (that is $36\%$ of the
prior function variance) when $22$ spectral points were used.
Figures~\ref{ssgp} and~\ref{ssgp_example} also show the covariance
function for the SSGPR when $88$ spectral points were used.  In this
case, the SSGPR covariance function approximation differed by as much
as $0.23$ (that is, $23\%$ of the prior function variance). Clearly, the
eigenfunction model is a much more accurate representation of the
actual generative kernel even when using only a quarter of the number
of basis functions as the SSGPR.

The error in the SSGPR representation of the covariance function can
have a significant impact on the accuracy of GP inference as the SSGPR
can significantly underestimate the posterior variance of the target
function.  To illustrate the extent of this problem,
Figure~\ref{ssgp_example} shows the posterior distributions of a
sparsely measured function inferred using Equations~\eqref{GP1}
and~\eqref{GP2} and the SSGPR and eigenfunction approximations of the
covariance functions.  Clearly, the SSGPR variances in the top two
panes are less than those calculated using the squared-exponential
model (bottom right pane) and the approximate eigenfunction model
(lower left pane).  Furthermore, Table~\ref{SSGP_tab} compares the
RMSE and expected log likelihood for the SSGPR and KPCA approaches
over $100$ functions drawn from the GP.  Each function is measured
every $10$ units, as above, with no measurement noise.  The SSGPR
propensity to underestimate the posterior variance is demonstrated by
a very low expected loglikelihood of $-6.9\times 10^4$ compared to
$75$ for the KPCA eigenfunction method.  Even when the number of
spectral points is increased four fold the KPCA approach is still more
accurate.

In summary, the eigenfunction model is a more efficient representation
than the SSGPR in that it identifies an orthogonal basis and
consequently requires fewer basis functions to capture the significant
features of the generative kernel.  Further, as we saw earlier, the
eigenfunction approach generalises to non-stationary kernels which can
be represented efficiently by non-sinusoidal basis functions.
Consequently, we advocate the eigenfunction approach over the SSGPR
approach for generating the basis for use with LFMs.

\begin{table}[ht]
  \caption{\label{SSGP_tab} RMSE and expected log likelihood for
    KPCA and
    SSGPR with the same number of basis functions and also SSGPR with
    four fold increase in the number basis functions.}
\begin{center}
\begin{tabular}{c|c|c}
\hline Method & RMSE & ELL \\ \hline\hline  
SSGPR & $3.03 \pm 0.04$ & $-6.9\times 10^4 \pm 0.6\times 10^4$ \\  
SSGPR (x4) & $2.74 \pm 0.03$ & $-1.2\times 10^4 \pm 0.1\times 10^4$ \\   
KPCA & ${\bf 2.49 \pm 0.03}$ & ${\bf 75.0 \pm 0.9}$ \\\hline
\end{tabular}
\end{center}
\end{table}
  
In the next section we extend our eigenfunction approach to quasi-periodic
latent force models.  This is a key contribution of our paper.

\section{REPRESENTING QUASI-PERIODIC LATENT FORCES WITH LINEAR BASIS MODELS}
\label{sec:quasi}
The eigenfunction basis model presented in the previous section assumes
that the latent force is perfectly periodic.  However, the force may
change gradually from cycle to cycle despite the latent force kernel
parameters remaining fixed.  For example, the force's phase may change
between cycles.  In the home heating application (described in detail
in Section~\ref{sec:appthermal}), where the residual heat within a
home is modelled as a latent force, a phase shift in the residual heat
profile may arise from cooking dinner at slightly different times from
day to day.

When the latent force process, $u(t)$, is not perfectly periodic but exhibits some
regularity from cycle to cycle it is called {\em
  quasi-periodic} and is often modelled as the product of two kernels
\citep{rasmussen06},
\begin{eqnarray} 
K_{\text{quasi-periodic}}(t,t^\prime)=K_{\text{quasi}}(t,t^\prime)
K_{\text{periodic}}(t,t^\prime)\ ,
\label{qpk}
\end{eqnarray} 
where $K_{\text{periodic}}(t,t^\prime)$ is a periodic kernel
(stationary or non-stationary) and $K_{\text{quasi}}(t,t^\prime)$ is a
non-periodic kernel which reduces the inter-cycle correlations. For
example, in \citet{roberts12}, their quasi-periodic kernel is the
product of a squared-exponential kernel and a periodic
squared-exponential kernel,
\begin{eqnarray}
K_{\text{quasi-periodic}}(t,t^\prime)=\sigma^2 \exp\left(-\frac{\left(t-t^\prime\right)^2}{l_\text{quasi}^2}\right)
\exp\left(-\frac{\sin\left(\frac{\pi(t-t^\prime)}{D_\text{periodic}}\right)^2}{l_\text{periodic}^2}\right)\
.
\label{quasiperiodicexample}
\end{eqnarray}
We note that Equation~\eqref{nonstatex} is also a quasi-periodic
covariance function.

We will now demonstrate that $K_{\text{quasi-periodic}}(t,t^\prime)$ can be
modelled within the state-space approach by LBMs by letting the
eigenfunction weights, $a$ as per Equation~\eqref{lbm}, change
dynamically.  Equation~\eqref{lbm} can be extended to include
time varying process weights, $a(t)$  \citep{ohagan78},
\begin{eqnarray}  
u(t)=\sum_{j} a_j(t)\phi_j(t)\ . 
\label{stochcoeff}
\end{eqnarray}    
Thus, when $u(t)$ is generated by a quasi-periodic
kernel then,
\begin{eqnarray*}
K_\text{quasi-periodic}(t,t^\prime)=E[u(t)\ u(t^\prime)]=\sum_{ij} \phi_i(t) E[a_i(t)
a_j(t^\prime)] \phi_j(t^\prime)\ .
\end{eqnarray*}
We assume that $a_i(t)$ is drawn from a Gaussian process, 
\begin{eqnarray}
a_i\sim \mathcal{GP}(0,\mu^\phi_i K_{\text{quasi}})\ ,
\label{stochcoeffproc}
\end{eqnarray}
where $\mu^\phi_i$ is the eigenvalue for the eigenfunction, $\phi_i$,
of $K_{\text{periodic}}$ as per Equation~\eqref{eigenexact}.  We also
assume that each weight process is independent.  Thus,
\begin{eqnarray*}
E[a_i(t) a_j(t^\prime)]=
\begin{cases}
\mu^\phi_i K_{\text{quasi}}(t,t^\prime) & \text{\ if } i=j\ ,\\
0 & \text{\ if } i\not= j\ .
\end{cases}
\end{eqnarray*}
Consequently,
\begin{eqnarray*}
K_{\text{quasi-periodic}}(t,t^\prime) &=&\sum_{i} \phi_i(t) \mu^\phi_i
K_{\text{quasi}}(t,t^\prime) \phi_i(t^\prime)\\
&=&K_{\text{quasi}}(t,t^\prime)\  \sum_{i} \phi_i(t) \mu^\phi_i
\phi_i(t^\prime)\\
&=&K_{\text{quasi}}(t,t^\prime) K_{\text{periodic}}(t,t^\prime)\ .
\end{eqnarray*}
We see that the periodic component of the model, $K_\text{periodic}$,
is represented by the basis function, $\phi$, in the LBM whereas the
non-periodic component, $K_\text{quasi}$, is represented via the time
varying LBM coefficients, $a$.  Note that, whereas for the resonator
model, as per Equations~\eqref{lbmres} and~\eqref{resonator1}, the
Fourier basis functions, $\phi$, are stochastic functions of time, in
the eigenfunction approach, the coefficients, $a$, are stochastic
functions of time and they reassign weight to fixed basis functions,
$\phi(t)$.

In order to accommodate variant LBM coefficients in the Kalman filter
we assume that each LBM coefficient is drawn from a stationary
Gaussian process with covariance function, $K_\text{quasi}$, as per
Equation~\eqref{stochcoeffproc}.  In which case, we can express the
eigenfunction weight Gaussian process, $a_r(t)$, as a stochastic
differential equation as per Equation~\eqref{gp_ltisde},
\begin{equation}
\frac{d {\bf A}_r(t)}{dt} = {\bf F}_r\, {\bf A}_r(t) + {\bf W}_r\ \omega_r(t)\ ,
\label{coefftime2}
\end{equation}
where the state vector, ${\bf A}_r(t)$, comprises the coefficient time
series and its derivatives, ${\bf A}_r(t) = (a_r(t)\; \frac{d
  a_r(t)}{dt}\ ,\cdots \ ,\frac{d^{p_r-1}a_r(t)}{dt^{p_r-1}})^T$.
Thus, as ${\bf A}_r$ can be expressed as a stochastic differential
equation then it can be inferred using the Kalman filter as
demonstrated in \cite{hartikainen10}.  We can weaken the stationarity
assumption and thus permit a greater choice for $K_\text{quasi}$ by
allowing changes in $K_\text{quasi}$'s output scale at discrete time
instances called {\em changepoints}. 

We propose three forms for $K_{\text{quasi}}$ which are the {\em
  Continuous Quasi model} (CQM), the {\em Step Quasi model} (SQM) and
the {\em Wiener-step Quasi model} (WQM).  Although many other
quasi-periodic forms are possible these models are chosen as they can
each be represented efficiently within the Kalman filter state vector,
as we will see in Section~\ref{sec:sslfm}, whilst capturing the key
qualitative properties of the data we wish to model.  Specifically,
the CQM models smooth, continuous deviations from cyclic behaviour
over time, and, consequently, closely resembles the quasi-periodic
model in \citet{roberts12}.  Alternatively, the SQM and WQM impose
stationarity within a cycle but allow for function variation between
cycles. We demonstrate that each can be represented in the Kalman
filter via a single variable in the
state-vector.\\

\noindent {\bf Continuous Quasi Model (CQM)}: This stationary model imposes
changes in the cycle continuously over time $t$.  It is equivalent to
the Mat\'ern kernel with order $\nu=1/2$,
\begin{eqnarray}
K^{\text{CQM}}_{\text{quasi}}(t,t^\prime)=\sigma_r^2\exp\left(-\frac{|t-t^\prime|}{l_r}\right)\
.
\label{CQMkernel}
\end{eqnarray}
The input hyperparameter, $l_r$, is positive.  As the CQM covariance
function, $K_{\text{CQM}}$, is a first order Mat\'ern, as per
Equation~\eqref{firstordermat}, it can be represented as a Markov
process, as per Equation~\eqref{coefftime2}.  The process model, ${\bf
  F}_r$, and white noise spectral density, $q_r$, for the first order Mat\'ern are
presented in Equations~\eqref{firstordermat1}
and~\eqref{firstordermat2}.  Reproducing this model here for
completeness, if $a$ is drawn from a GP with the quasi-periodic kernel
in Equation~\eqref{CQMkernel}, $a_r\sim
\mathcal{GP}(0,K^{\text{CQM}}_{\text{quasi}})$, then,
\begin{eqnarray*}
\frac{d a_r(t)}{dt}=F_r a_r(t)+\omega_r(t)\ ,
\end{eqnarray*}
where, $\omega_r(t)$ is a white noise process with spectral density
$q_r$ and,
\begin{eqnarray*}
F_r=-\frac{1}{l_r}\ ,\hspace*{2cm} q_r=\frac{2\sigma_r^2\sqrt{\pi}}{l_r\
  \Gamma(0.5)}\ ,
\end{eqnarray*}
and $l_r$ and $\sigma_r$ are the input and output scales,
respectively, as per Equation~\eqref{CQMkernel}.  We note that, by
using the CQM kernel as part of the quasi-periodic latent force
covariance function, each LBM coefficient, $a_r(t)$, can be
represented by a single variable in the Kalman filter state vector.
In Section~\ref{sec:sslfm} we will demonstrate how this continuous
time LTI model can be
incorporated into a discrete time LFM model.\\

\noindent {\bf Step Quasi Model (SQM)}: This model can be used to
decorrelate cycles at changepoints between cycles.  This
non-stationary model preserves the variance of the periodic function
each side of the changepoint.  However, the function's correlation
across the changepoint is diminished.  For times, $t$ and $t^\prime$, with
$t$ and $t^\prime$ in the same cycle $K_{\text{quasi-periodic}}(t,t^\prime)
=K_{\text{periodic}}(t,t^\prime)$. When times $t$ and $t^\prime$ correspond to
different cycles then
$K_{\text{quasi-periodic}}(t,t^\prime)< K_{\text{periodic}}(t,t^\prime)$.  If $N$
consecutive cycles are labelled $C=1,2,\ldots,N$ and $C(t)$ denotes the
cycle index for time $t$ then,
\begin{align} 
K^{\text{SQM}}_{\text{quasi}}(t,t^\prime)=\sigma_r^2\exp\left(-\frac{|C(t)-C(t^\prime)|}{l_r}\right)\
.
\label{sqmdef}
\end{align} 
Again, the kernel input hyperparameter, $l_r$, is positive.\\
 
\noindent {\bf Wiener-step Quasi Model (WQM)}: Again, we assume the
presence of changepoints between cycles.  This non-stationary model
increases the variance of the function at the changepoint.  If $N$
consecutive cycles are labelled $C=1,\ldots,N$ then,
\begin{align} 
K^{\text{WQM}}_{\text{quasi}}(t,t^\prime)=\xi_0+\min(C({t^\prime}),C(t))
\xi_r\ ,
\label{wqmdef}
\end{align} 
where $\xi_0$ and $\xi_r$ are positive.\\

\begin{figure}[ht]
\begin{center}
\begin{tabular}{cc} 
\includegraphics[width=0.4\textwidth]{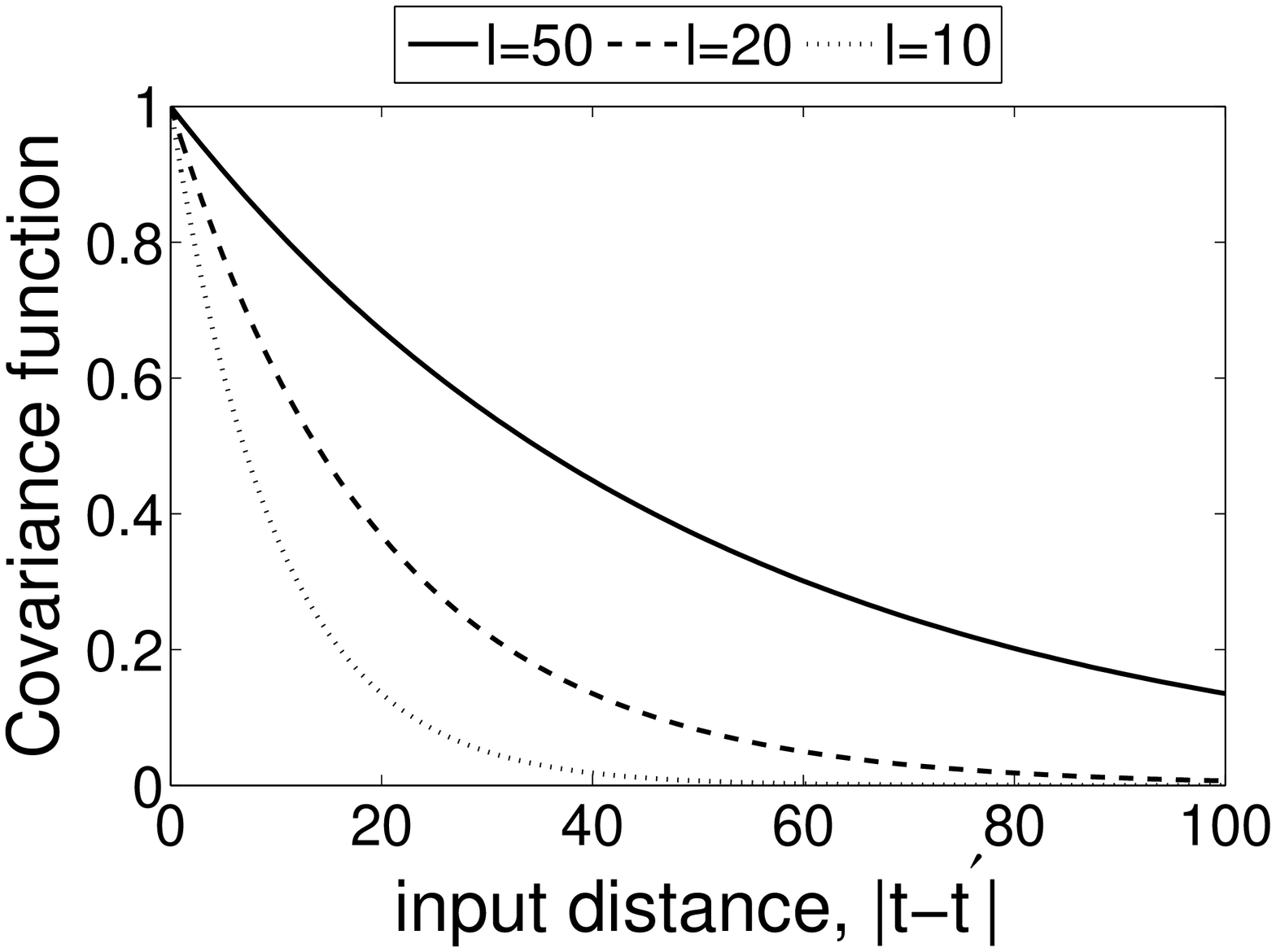}&
\includegraphics[width=0.4\textwidth]{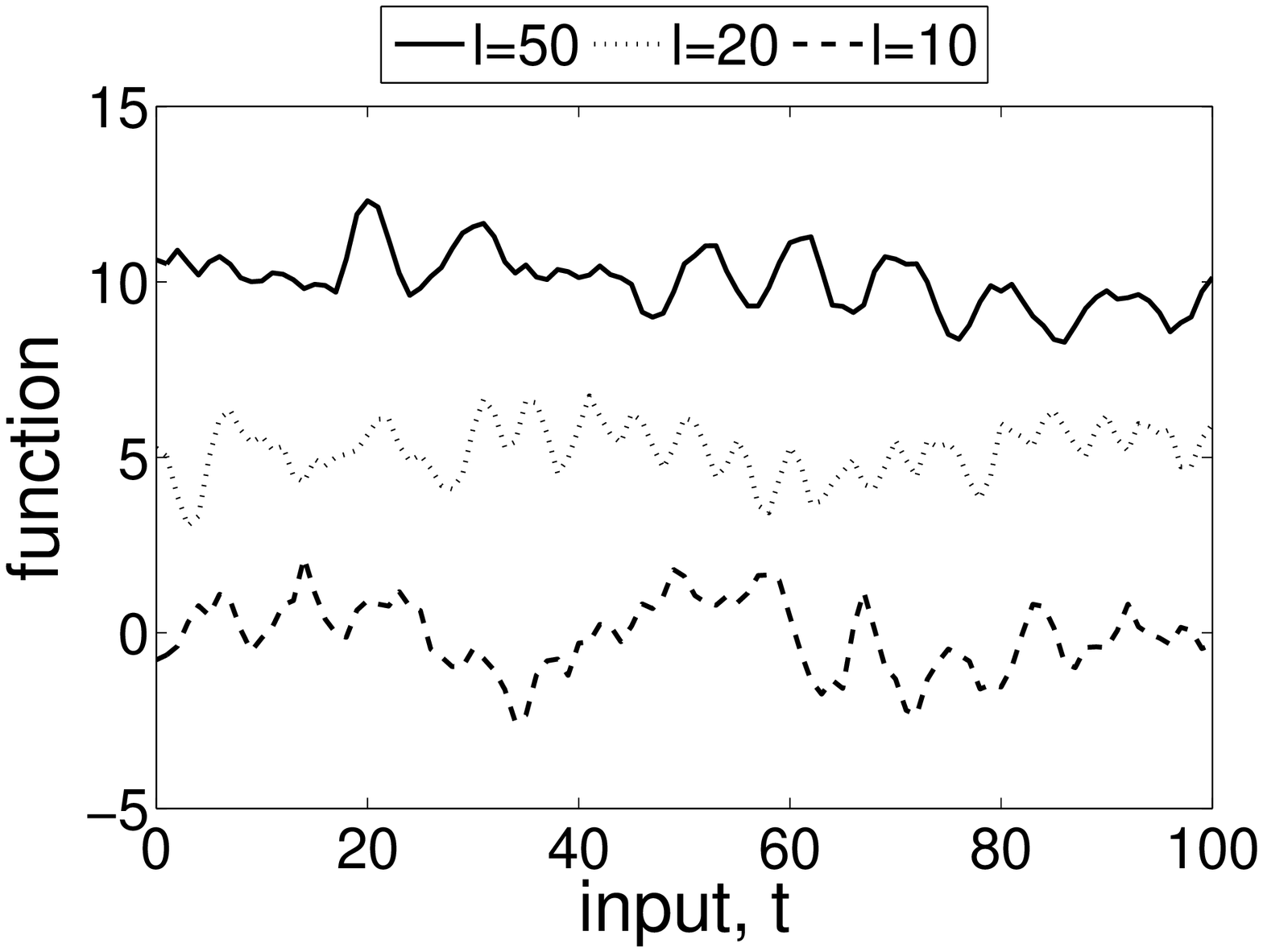}\\ 
(a) CQM Covariance functions & (b) CQM Sample functions \\ \ \\
\includegraphics[width=0.4\textwidth]{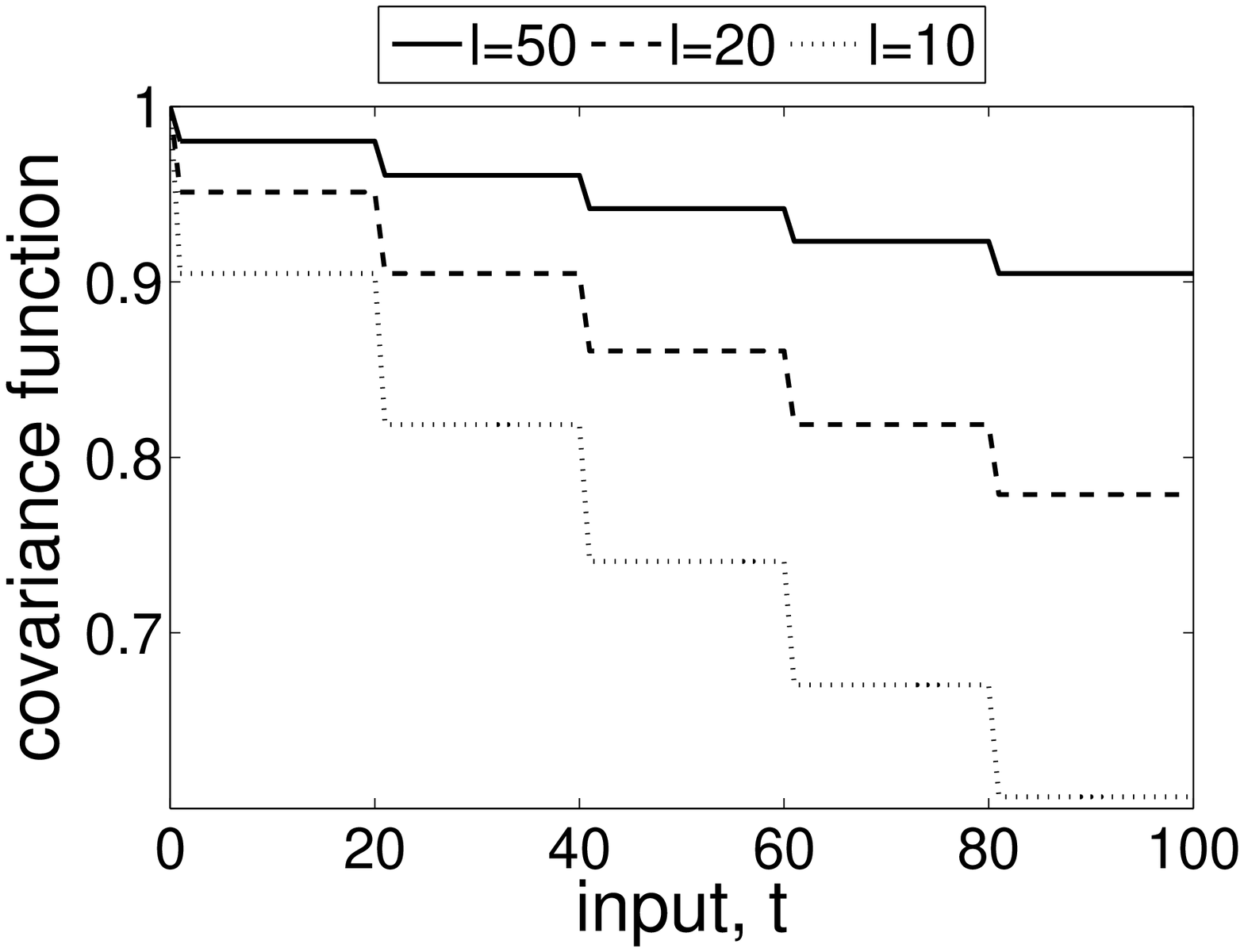} & 
\includegraphics[width=0.4\textwidth]{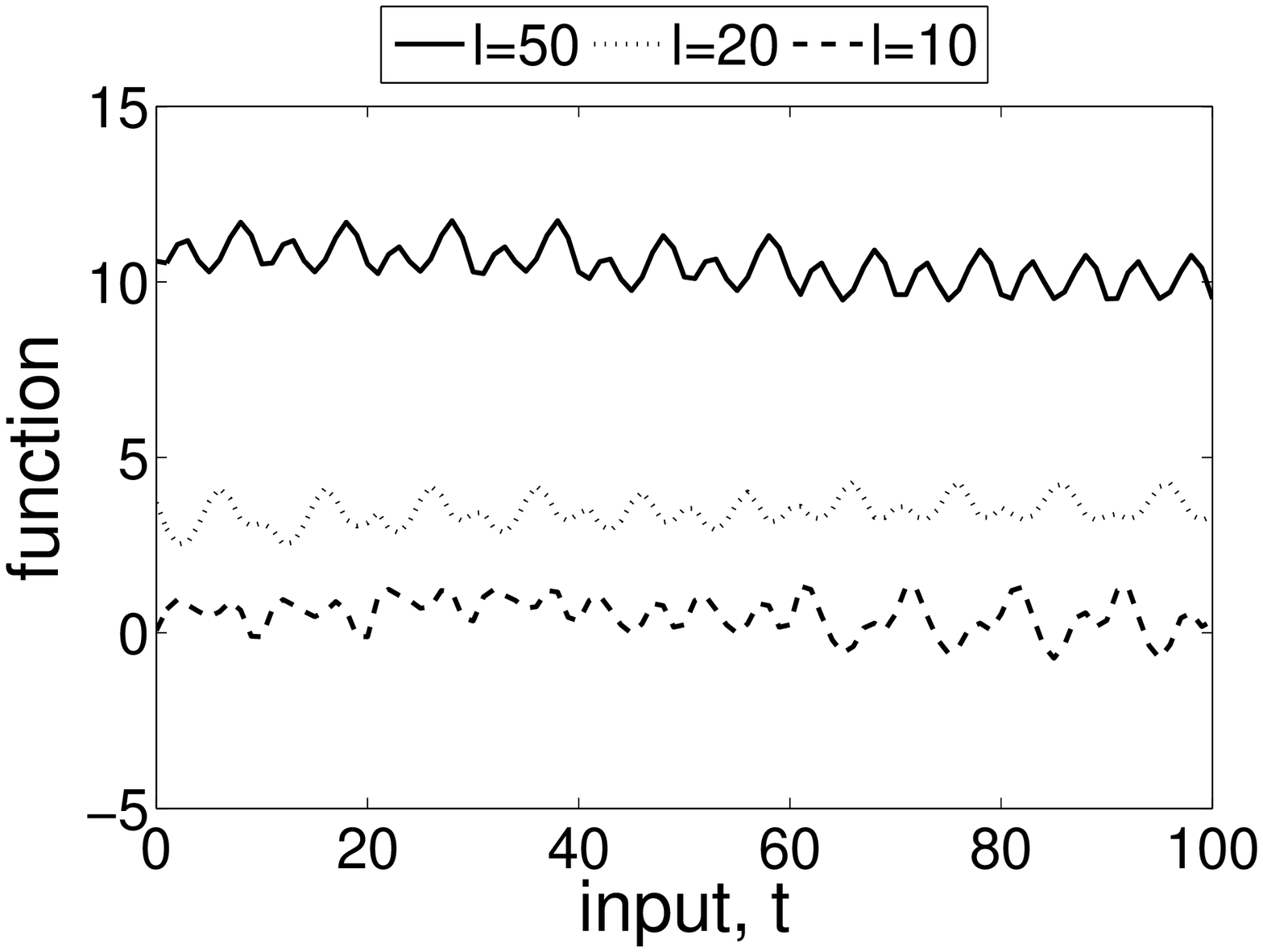}\\
(c) SQM Covariance functions & (d) SQM Sample functions\\ \ \\
\includegraphics[width=0.4\textwidth]{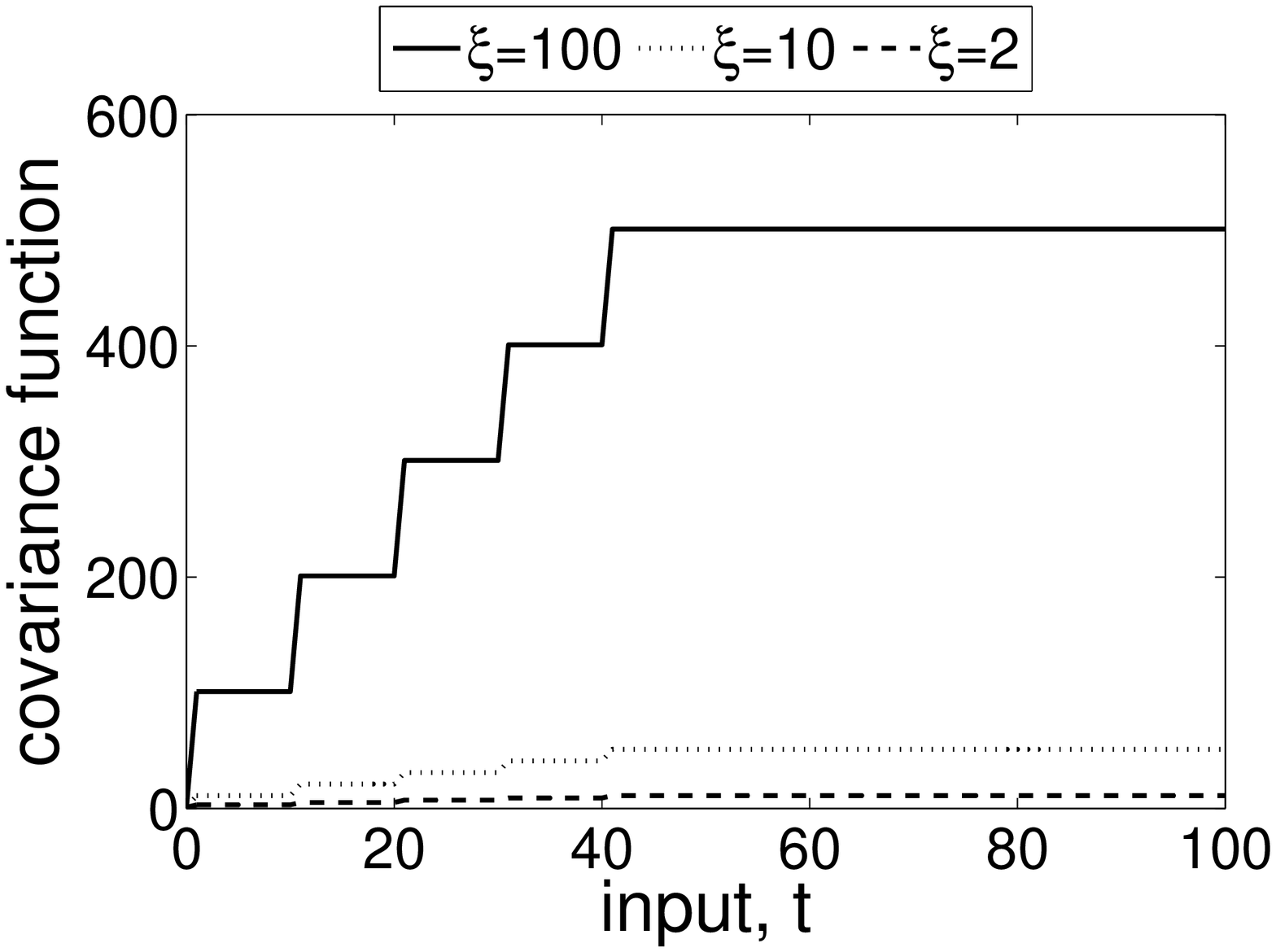} & 
\includegraphics[width=0.4\textwidth]{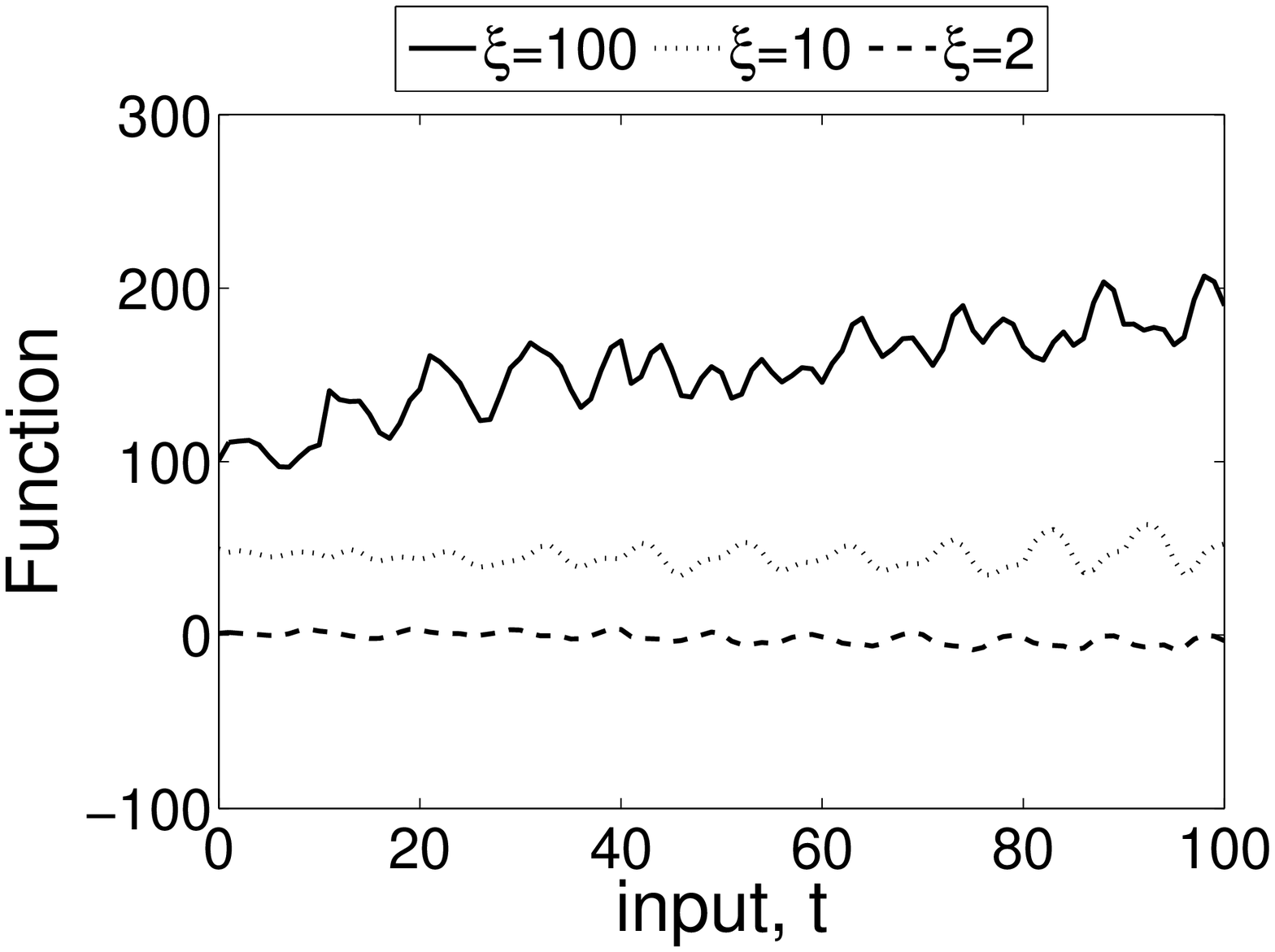}\\
(e) WQM Covariance functions & (f) WQM Sample functions
\end{tabular}
\end{center}
\caption{\label{quasicovfn} Covariance functions (left column) for
  CQM, SQM and WQM.  Also, sample quasi-periodic functions (right
  column) for CQM, SQM and WQM quasi kernels and a squared-exponential
  periodic kernel.}
\end{figure}

Example covariance functions for the three forms for $K_{\text{quasi}}$ 
are shown in Figure~\ref{quasicovfn}.  Also, sample quasi-periodic 
function draws are shown for each kernel.  The functions are drawn 
from a quasi-periodic squared-exponential kernel
$K_{\text{quasi-periodic}}(t,t^\prime)$ with $K_{\text{periodic}}(t,t^\prime)$ 
the periodic squared-exponential $K_\text{SE}$, as per 
Equation~\eqref{periodicsquaredexp}, with period $D=10$ units, various 
input scales $l$ (specified within each subfigure) and
$K_{\text{quasi}}(t,t^\prime)$ set to either
$K^{\text{CQM}}_{\text{quasi}}(t,t^\prime)$,
$K^{\text{SQM}}_{\text{quasi}}(t,t^\prime)$ or
$K^{\text{WQM}}_{\text{quasi}}(t,t^\prime)$.  In the case of SQM and WQM a 
new cycle begins every $10$ time units.

The SQM and WQM kernels can be incorporated into the discrete time
Kalman filter by firstly expressing them as continuous time
differential equations as per Equation~\eqref{gp_ltisde}.  Suppose
that either $a_r\sim\mathcal{GP}(0,K^{\text{SQM}})$ or
$a_r\sim\mathcal{GP}(0,K^{\text{WQM}})$ then,
\begin{eqnarray*}
\frac{d a_r(t)}{dt}=0\ ,
\end{eqnarray*}
everywhere, except at changepoints.  Thus, in the case of SQM and WQM
the corresponding process $a_r(t)$ can be represented via a first
order differential equation as per Equation~\eqref{coefftime2} with
${\bf A}_r(t)=a_r(t)$, ${\bf \Delta}_r=1$, ${\bf F}_r=0$ and ${\bf
  W}_r=0$.  However, at a changepoint, $\tau$, the
SQM and WQM covariance functions jump in value as can
be seen in Figure~\ref{quasicovfn} at input distances $20$ and $40$,
for example.  The value of the process, $a_r(\tau)$, immediately after
the changepoint is related to the process, $a_r(\tau_-)$, immediately
before the jump thus,
\begin{eqnarray}
a_r(\tau)=G_r^* a_r(\tau_-)+\chi_r^*(\tau)\ ,
\label{perturbation}
\end{eqnarray}
where $G_r^*$ is the process model and $\chi_r^*(\tau)$ is a Gaussian random variable,
$\chi_r^*(\tau)\sim \mathcal{N}(0,Q_r^*)$.  In Appendix~\ref{app:discretejump}
we demonstrate that the process model, $G_r^*$, and process noise
variance, $Q_r^*$, for the SQM at the changepoint are,
\begin{eqnarray}
G^*_\text{r,SQM}=\exp\left(-\frac{1}{l_r}\right)\ ,\label{GSQM}
\end{eqnarray}
and,
\begin{eqnarray}
Q^*_\text{r,SQM}=\sigma_r^2\left(1-\exp\left(-\frac{2}{l_r}\right)\right)\ ,\label{QSQM}
\end{eqnarray}
respectively.  Similarly, Appendix~\ref{app:discretejump} also shows
that the process model, $G_r^*$, and process noise variance, $Q_r^*$,
for the WQM at a changepoint are,
\begin{eqnarray} 
G^*_\text{r,WQM}=1\ ,\label{GWQM}
\end{eqnarray} 
and,
\begin{eqnarray} 
Q^*_\text{r,WQM}=\xi_r\ ,\label{QWQM}
\end{eqnarray} 
respectively.  The latent force variance increases at the changepoint
under the WQM whereas the variance remains unchanged for the SQM.  In
Section~\ref{sec:sslfm}, we demonstrate how these expressions for
$G^*$ and $Q^*$ are incorporated within the discrete form of the
Kalman filter.

The three forms for $K_{\text{quasi}}$ will be applied to both the
call centre customer queue tracking and home temperature prediction
problem domains in Sections~\ref{sec:appqueue} and~\ref{sec:appthermal}.  In
the next section we describe how we perform inference with a LFM using
a state-space approach, where the state vector is augmented with
periodic or quasi-periodic latent forces that are approximated using
the latent force eigenfunctions.

\section{RECURSIVE ESTIMATION WITH PERIODIC AND QUASI-PERIODIC LATENT FORCE MODELS}
\label{sec:sslfm} 
This section describes a state-space approach to inference with LFMs
in some detail.  We shall treat the periodic and non-periodic latent
forces differently when performing inference with them.  Following
\citet{hartikainen10,hartikainen11}, non-periodic forces will be
modelled using the power spectrum of their corresponding covariance
functions.  Alternatively, the periodic latent forces will be modelled
using the eigenfunctions of the corresponding periodic covariance
function. The key idea in this section is to infer the LFM unknowns
via the Kalman filter.  The unknowns include the non-periodic forces
and their derivatives, as per Equation~\eqref{diffmodel}, along with
the coefficients of the periodic forces, as per
Equation~\eqref{coefftime2}.  The remainder of this section describes
in detail how the KF state is predicted forward in time and how
measurements of the system are folded into the state
estimate.

We examine periodic and quasi-periodic cases separately as state-space
inference with periodic latent forces uses a more compact model.  For
the periodic case, we assume that the latent forces, ${\bf u}$, as per
Equation~\eqref{initiallfm}, can be separated into two distinct sets,
periodic forces, ${\bf u}_\text{p}$, and non-periodic forces, ${\bf
  u}_\text{np}$, so that ${\bf L}{\bf u}(t)= {\bf L}_\text{np} {\bf
  u}_\text{np}(t) + {\bf L}_\text{p} {\bf u}_\text{p}(t)$ as described
in Section~\ref{sec:lfm}.  Then, Equation~\eqref{initiallfm} becomes,
\begin{eqnarray*}
\frac{d{\bf z}(t)}{dt}= {\bf F}\ {\bf z}(t) + {\bf L}_\text{np} {\bf
  u}_\text{np}(t) + {\bf L}_\text{p} {\bf
  u}_\text{p}(t)\ .
\end{eqnarray*}
We model non-periodic latent forces and their derivatives, as per
Equation~\eqref{gp_ltisde}, and periodic forces using eigenfunctions
as per Equation~\eqref{sparse}.  We define the augmented state vector,
${\bf z}_a$, as per Equation~\eqref{diffmodel}, and also the
corresponding periodic force coefficients, ${\bf L}^a_\text{p}=[{\bf
  L}_\text{p}^T,\ {\bf 0}^T]^T$ so that the forces ${\bf u}_\text{p}$
still act on ${\bf z}$ within ${\bf z}_a$,
\begin{eqnarray}
\frac{d{\bf z}_a(t)}{dt}= {\bf F}_a\ {\bf z}_a(t) + {\bf
  L}_a\omega_a(t) + {\bf L}^a_\text{p}  {\bf u}_\text{p}(t)\ ,
\label{difflbmp}
\end{eqnarray} 
where $\omega_a$ and ${\bf L}_a$ are as per Equations~\eqref{omega}
and~\eqref{L}. 
 
We now introduce our eigenfunction model for the periodic latent
forces into Equation~\eqref{difflbmp}.  First, we consider periodic
latent forces, introduced in Section~\ref{sec:LBM}, for which the
corresponding LBM coefficients, $\{a\}$ in Equation~\eqref{sparse},
are constant over time.  Substituting our Nystr\"om approximation
basis model for the periodic forces, as per Equation~\eqref{phi}, into
the dynamic differential model, as per Equation~\eqref{difflbmp},
\begin{eqnarray}
\frac{d {\bf z}_a(t)}{dt}= {\bf F}_a\ {\bf z}_a(t) + {\bf
  L}_a\omega_a(t) +\sum_{r=1}^R\sum_{j=1}^{J_r} {\bf L}^a_p(\cdot,r)\
\tilde\phi_{rj}(t)\ a_{rj}\ ,\label{lbmp}
\end{eqnarray}  
where $R$ is the number of latent forces, $J_r$ is the number of
eigenfunctions for latent force $r$, $a_{rj}$ are the eigenfunction weights
and the vector ${\bf L}^a_\text{p}(\cdot,r)$ is the
$\text{r}^\text{th}$ column of the matrix ${\bf L}^a_\text{p}$ in
Equation~\eqref{difflbmp}.  The Nystr\"om basis function, $\tilde{\phi}_{rj}$, 
is,
\begin{eqnarray}
\tilde{\phi}_{rj}(t)=\frac{\sqrt{N_r}}{\mu_{rj}} 
K_r(t,S_r) {\bf v}_{rj}\ , \label{M}
\end{eqnarray}  
where $K_r$, $S_r$ and $N_r$ are the covariance function, the
quadrature points at which the kernel is sampled for force $r$, as per 
Equation~\eqref{quadrature}, and the cardinality of $S_r$.  The
$\mu_{rj}$ and ${\bf v}_{rj}$ are the Gram matrix eigenvalues and 
eigenvectors, respectively.

The differential equations~\eqref{lbmp} have the solution,
\begin{eqnarray} 
{\bf z}_a(t)= {\bf \Phi}(t_0,t){\bf
    z}_a(t_0)+{\bf q}_a(t_0,t) +\sum_{r=1}^R \sum_{j=1}^{J_r} a_{rj}
  {\bf M}_{rj}(t_0,t)\ ,
\label{solution2}
\end{eqnarray} 
where, again, ${\bf \Phi}(t_0,t)$ denotes the matrix exponential, ${\bf
  \Phi}(t_0,t)=\exp({\bf F}_a(t-t_0))$, and ${\bf q}_a(t_0,t)\sim \mathcal{N}({\bf
  0},{\bf Q}_a(t_0,t))$ where,
\begin{eqnarray*}
{\bf Q}_a(t_0,t)=\int_{t_0}^t {\bf \Phi}(s,t) {\bf L}_a \Lambda_a
{\bf L}_a^T{\bf \Phi}(s,t)^T ds\ ,
\end{eqnarray*}
and $\Lambda_a$, as per Equation~\eqref{Lambda}, is the spectral
density of the white noise processes corresponding to the
non-periodic latent forces.  The matrix
${\bf M}_{rj}(t_0,t)$ is the convolution of the state transition model,
${\bf \Phi}$, with each of the periodic latent force eigenfunctions,
\begin{eqnarray*}
{\bf M}_{rj}(t_0,t)=\frac{\sqrt{N_r}}{\mu_{rj}} \left[\int_{t_0}^t ds\ {\bf \Phi}(s,t)
 {\bf L}^a_\text{p}(\cdot,r)  
K_r(s,S_r) \right] {\bf v}_{rj}\ .
\end{eqnarray*}
For small time intervals $[t_0,\ t]$, which is the case for our
applications in Sections~\ref{sec:appqueue} and~\ref{sec:appthermal},
${\bf M}_{rj}$ can be calculated using numerical matrix exponential
integration methods.  Further, we note ${\bf \Phi(t_0,t)}$ is
stationary and this can mitigate the need to recalculate this matrix
exponential at each instance of the time series.
 
To accommodate the latent forces within the Kalman filter we must
ensure that our discrete time dynamic model, as per
Equation~\eqref{solution2}, has the appropriate form. Specifically,
\begin{eqnarray*}
{\bf X}(t)={\bf G}(t_0,t) {\bf X}(t_0)+\omega(t_0,t)\ ,
\end{eqnarray*}  
where the noise process, $\omega$, is i.i.d Gaussian and
zero-mean.  In order to rewrite Equation~\eqref{solution2} into the
appropriate form for Kalman filter inference we define a vector, ${\bf
  a}$, as per Equation~\eqref{lbm}, which collects together the
eigenfunction weights,
\begin{eqnarray*}
{\bf a}=(a_{11},\ldots,a_{1J_1},
a_{21},\dots,a_{2J_2}\ldots)^T\ ,
\end{eqnarray*}
and, similarly, a matrix, ${\bf M}$, which collects together the
convolutions, ${\bf M}_{rj}$,
\begin{eqnarray*}
{\bf M}(t_0,t)=({\bf M}_{11}(t_0,t),\ldots,{\bf M}_{1J_1}(t_0,t),{\bf
  M}_{21}(t_0,t),\dots,{\bf M}_{2J_2}(t_0,t) \ldots)\ .
\end{eqnarray*}
We further augment the state vector to accommodate the model weights,
${\bf a}$, corresponding to the periodic latent forces.  Let,
\begin{eqnarray}
{\bf X}(t)=( {\bf  z}^T_a(t), {\bf a}^T)^T\ ,
\label{augstatestat}
\end{eqnarray}
be our augmented state vector which now
accommodates the derivative auxiliary variables in ${\bf z}_a$
required by the non-periodic forces as per \citet{hartikainen11} and
the eigenfunction weights, ${\bf a}$, required by the periodic forces
as per our approach. When the eigenfunction weights are constant we
can rewrite Equation~\eqref{solution2},
\begin{eqnarray}
{\bf X}(t)={\bf G}(t_0,t){\bf X}(t_0)+{\bf \omega}(t_0,t)\ ,
\label{prediction}
\end{eqnarray} where,
\begin{eqnarray}
{\bf G}(t_0,t)=\begin{pmatrix} {\bf \Phi}(t_0,t) & {\bf M}(t_0,t)\\
{\bf 0} & {\bf I}\ ,
\end{pmatrix}
\label{transitionmodel}
\end{eqnarray}  
and,
\begin{eqnarray*}
{\bf \omega}(t_0,t)=\begin{pmatrix} {\bf q}_a(t_0,t) \\ {\bf
    0} \end{pmatrix}\ .
\end{eqnarray*}
Thus, predictions of the Gaussian process, ${\bf X}$, can be inferred
using the Kalman filter.  Of course, the model in
Equation~\eqref{prediction} can also be incorporated within the Kalman
Smoother to perform full (that is, forward and backward) regression over
${\bf X}(t)$ for all time $t$ if required \citep{hartikainen10}.  The
prediction equations for the state mean, $\bar{\bf X}(t\mid t_0)$, and
covariance, ${\bf P}(t\mid t_0)$, at time $t$ conditioned on
measurements obtained up to time $t_0$, are,
\begin{eqnarray}
\bar{\bf X}(t\mid t_0)&=&{\bf G}(t_0,t)\bar{\bf X}(t_0\mid t_0)\ ,\label{KF1}\\
{\bf P}(t\mid t_0)&=&{\bf G}(t_0,t){\bf P}(t_0\mid t_0){\bf G}(t_0,t)^T+{\bf Q}(t_0,t)\ ,\label{KF2}
\end{eqnarray} 
where ${\bf Q}(t_0,t)\triangleq \begin{pmatrix} {\bf Q}_a(t_0,t) &
  {\bf 0}\\ {\bf 0} & {\bf 0}\end{pmatrix}$.

We assume that measurements, ${\bf y}$, are Gaussian distributed,
\begin{eqnarray}
{\bf y}(t)={\bf H}\ {\bf X}(t)+{\bf \eta}(t)\ ,
\label{update}
\end{eqnarray}  
where ${\bf \eta}$ is zero-mean multivariate Gaussian, $\eta\sim
\mathcal{N}({\bf 0},{\bf Z})$, where ${\bf Z}$ is the observation noise
covariance matrix and the {\em measurement model}, ${\bf H}$, extracts
the appropriate elements of the state vector.  These measurements can
be folded into the Kalman filter in the usual way.  The update
equations given measurement, ${\bf y}(t)$, as per
equation~\eqref{update}, are,
\begin{eqnarray}
\bar{\bf X}(t\mid t)&=&\bar{\bf X}(t\mid t_0)+{\bf K}({\bf y}(t)-{\bf
  H}\bar{\bf X}(t\mid t_0))\ ,\\ \label{KF3}
{\bf P}(t\mid t)&=&({\bf I}-{\bf K}{\bf H}){\bf P}(t\mid t_0)\ , \label{KF4}
\end{eqnarray}
where ${\bf K}$ is the Kalman gain,
\begin{eqnarray}
{\bf K}={\bf P}(t\mid t_0) {\bf H}^T({\bf H}{\bf P}(t\mid t_0){\bf
  H}^T+{\bf Z})^{-1}\ .
\label{kalmangain}
\end{eqnarray}
 
The computational complexity of the Kalman gain is cubic in the
cardinality of the measurement vector, ${\bf y}$ (that is, not necessarily
a function of the cardinality of the state).  The cubic cost arises
from the need to invert a covariance matrix in
Equation~\eqref{kalmangain}.  For a single output Gaussian process
this covariance will be a scalar.  However, for multi-output Gaussian
processes, when each physical process is measured, ${\bf y}(t)$ will
be a vector of (noisy) measurements of each process at time $t$.  In
which case, the computational complexity of the Kalman gain will be
cubic in the number of measured physical processes.  So, although the
state vector may be augmented in order to model both physical
processes and latent forces, as described above, these additions will
not impact on the cost of the matrix inversion in
Equation~\eqref{kalmangain}.

We next extend our state-space approach to accommodate quasi-periodic
latent forces.  For the quasi-periodic latent forces the corresponding
kernel LBM coefficients, ${\bf a}$, are functions of time, as per
Equation~\eqref{stochcoeff}.  We assume that each LBM coefficient is drawn
from a Gaussian process with covariance function, $K_\text{quasi}$,
as per Equation~\eqref{qpk}, and we now demonstrate how these dynamic
weight processes, ${\bf a}(t)$, are incorporated into the Kalman
filter, Equations~\eqref{KF1} to~\eqref{KF4}. 

As above, $a_{rj}$, corresponds to the $j$th eigenfunction for latent force
$r$.  However, for quasi-periodic latent forces each eigenfunction
weight is variant and we assume $a_{rj}(t)$ can be written as a
stochastic differential equation as proposed in
Section~\ref{sec:quasi},
\begin{equation}
\frac{d {\bf A}_{rj}(t)}{dt} = {\bf F}_{rj} {\bf A}_{rj}(t) + {\bf
  W}_{rj}\ \omega_{rj}(t)\ , 
\label{processdyn}
\end{equation}
where the state vector, ${\bf A}_{rj}(t)$, comprises derivatives of
the coefficient time series, ${\bf A}_{rj}(t) = (a_{rj}(t)\ , \frac{d
  a_{rj}(t)}{dt} \ ,\cdots \
,\frac{d^{p_{rj}-1}a_{rj}(t)}{dt^{p_{rj}-1}})^T$.  We can recover the
eigenfunction coefficient from ${\bf A}_{rj}$,
\begin{eqnarray*} 
a_{rj}(t)={\bf \Delta}_{rj}{\bf A}_{rj}(t)\ ,
\end{eqnarray*} 
where the vector ${\bf \Delta}_{rj}=(1,0,\ldots,0)$ is an indicator
vector which extracts the LBM coefficient $a_{rj}$ from ${\bf
  A}_{rj}$.  Thus, the latent force, $u_r(t)$, as per
Equation~\eqref{stochcoeff}, is,
\begin{eqnarray}
  u_r(t)=\sum_j a_{rj}(t) \phi_{rj}(t)=\sum_j \phi_{rj}(t) {\bf \Delta}_{rj} {\bf A}_{rj}(t)\ ,
\label{quasiur}
\end{eqnarray}
where $\phi_{rj}$ is the $j$th eigenfunction for the latent force $r$.
Substituting our Nystr\"om approximation for the eigenfunction, $\phi(t)$ as
per Equation~\eqref{phi}, into Equation~\eqref{quasiur} we get,
\begin{eqnarray*}
  u_r(t)=\sum_j \frac{\sqrt{N_r}}{\mu_{rj}} \left[
    K_r(t,S_r) \right] {\bf v}_{rj}{\bf \Delta}_{rj}{\bf A}_{rj}(t)\ .
\end{eqnarray*}
Then, substituting $u_r$ into the differential latent force model, 
Equation~\eqref{lbmp},
\begin{eqnarray}
\frac{d{\bf z}_a(t)}{dt}= {\bf F}_a\ {\bf z}_a(t)
 + {\bf L}_a\omega_a(t)
+\sum_{r=1}^R\sum_{j=1}^{J_r} {\bf m}_{rj}(t){\bf A}_{rj}(t)\ ,
\label{diffmodel2}
\end{eqnarray}
where $R$ is the number of latent forces, $J_r$ is the number of
eigenfunctions for latent force $r$, ${\bf L}_a$ and $\omega_a(t)$ are as per
Equations~\eqref{omega} and~\eqref{L} and the vector ${\bf m}_{rj}$,
\begin{eqnarray}
{\bf m}_{rj}(t)=\frac{\sqrt{N_r}}{\mu_{rj}} \left[
 {\bf L}^a_\text{p}(\cdot,r) 
K_r(t,S_r) \right] {\bf v}_{rj}{\bf \Delta}_{rj}\ ,
\label{m}
\end{eqnarray} 
where $K_r$, $S_r$ and $N_r$ are the covariance function, the
quadrature points at which the kernel is sampled for force $r$, as per
Equation~\eqref{quadrature}, and the cardinality of $S_r$.  The
$\mu_{rj}$ and ${\bf v}_{rj}$ are the Gram matrix eigenvalues and
eigenvectors, respectively.  The vector ${\bf L}^a_\text{p}(\cdot,r)$
is the $\text{r}^\text{th}$ column of the matrix ${\bf L}^a_\text{p}$
in Equation~\eqref{difflbmp}.

Now, as for the constant eigenfunction coefficient case, to exploit
the Kalman filter for LFM inference with quasi-periodic latent forces
we gather together all the LFM Gaussian variables, ${\bf z}_a$ and
$\{{\bf A}_{rj}\}$, into a single state-vector.  In so doing, we
define a vector ${\bf A}(t)$ which collects together the 
eigenfunction coefficients and their derivatives,
\begin{eqnarray}
{\bf A}(t)\triangleq ({\bf
  A}_{11}(t)^T,{\bf A}_{12}(t)^T,\ldots,{\bf A}_{21}(t)^T,{\bf
  A}_{22}(t)^T,\ldots,)^T\ , 
\label{augstateA}
\end{eqnarray} 
a matrix ${\bf m}(t)$ which collects together the vectors
$\{{\bf m}_{rj}\}$,
\begin{eqnarray*}
{\bf m}(t)\triangleq ({\bf
  m}_{11}(t),{\bf m}_{12}(t),\ldots,{\bf m}_{21}(t),{\bf
  m}_{22}(t),\ldots,)\ , 
\end{eqnarray*}
a matrix ${\bf F}_{\bf A}$ which collects together the process models
for all eigenfunction coefficients for all latent forces, as per
Equation~\eqref{processdyn},
\begin{eqnarray*}
{\bf F}_{\bf A}\triangleq
\text{blockdiag}({\bf F}_{11},{\bf F}_{12}\ldots,{\bf F}_{21},{\bf
  F}_{22},\ldots)\ ,
\end{eqnarray*}
a vector $\omega_{\bf A}$ which collects together the noise processes
for the non-periodic forces, $\omega_a$, as per
Equation~\eqref{diffmodel2}, and noise processes for all the
eigenfunction coefficients as per Equation~\eqref{processdyn},
\begin{eqnarray}
\omega_{\bf A}\triangleq (\omega_a^T,\omega_{11},\omega_{12},\ldots,
\omega_{21},\omega_{22},\ldots)^T\ ,
\label{diffbases}
\end{eqnarray}
and the block diagonal matrix ${\bf L}_{\bf A}$ which collects together the
corresponding process noise coefficients, ${\bf L}_a$, as per
Equation~\eqref{diffmodel2} and ${\bf W}_{rj}$ as per
Equation~\eqref{processdyn},
\begin{eqnarray*}
{\bf L}_{\bf A}\triangleq
\text{blockdiag}({\bf L}_a,{\bf W}_{11},{\bf W}_{12}\ldots,{\bf W}_{21},{\bf
  W}_{22},\ldots)\ .
\end{eqnarray*}
As per Equation~\eqref{augstatestat} let,
\begin{eqnarray}
{\bf X}(t)\triangleq ( {\bf  z}^T_a(t), {\bf A}^T(t))^T\ ,
\label{augstatedyn}
\end{eqnarray}
be our augmented state vector which now accommodates the derivative
auxiliary variables required by the non-periodic forces as per
\citet{hartikainen11} and the eigenfunction weights required by the
quasi-periodic forces as per our approach. Combining
Equations~\eqref{processdyn} and~\eqref{diffmodel2},
\begin{eqnarray}
\frac{d{\bf X}(t)}{dt}= \begin{pmatrix}{\bf F}_a & {\bf m}(t)\\ {\bf 0} & {\bf
    F}_{\bf A}\end{pmatrix}{\bf
  X}(t)+{\bf L}_{\bf A}{\bf \omega}_{\bf A}(t)\ ,
\label{inhomo}
\end{eqnarray}  
where $\omega_A(t)$ is a vector of independent white noise processes.
The spectral density of the ith white noise process in this vector is
$[\Lambda_A]_i$ where,
\begin{eqnarray*}
 \Lambda_A=(\Lambda_a,q_{11},q_{12},\ldots,q_{21,},q_{22},\ldots)\ .
\end{eqnarray*}
The $\Lambda_a$, as per Equation~\eqref{Lambda}, is the spectral
density of the white noise processes corresponding to the non-periodic
latent forces and $q_{rj}$, as per Equation~\eqref{processdyn}, is the
spectral density of the white noise process for the eigenfunction weight,
$a_{rj}$.

Unfortunately, Equation~\eqref{inhomo} is an inhomogeneous SDE as ${\bf
  m}$ is a function of time.  Consequently, in this form, ${\bf X}(t)$
cannot be folded into the Kalman filter.  However, by assuming ${\bf m}(t)$
is approximately constant over the short time interval, $[t_0\ , t]$,
and asserting ${\bf m}(t)\approx {\bf m}(t_0)$ then $\frac{d{\bf X}(t)}{dt}$ can be
integrated into the appropriate form,
\begin{eqnarray}
{\bf X}(t)= {\bf \Phi}(t_0,t){\bf X}(t_0)+{\bf q}(t_0,t)\ ,
\label{Xavar}
\end{eqnarray}   
where,
\begin{eqnarray}
{\bf \Phi}(t_0,t)=\exp\left[\begin{pmatrix}{\bf F}_a & {\bf m}(t_0)\\ {\bf 0} & {\bf
    F}_{\bf A}\end{pmatrix}(t-t_0)\right]\ .
\label{variant}
\end{eqnarray} 
The process noise, ${\bf q}(t_0,t)\sim\mathcal{N}({\bf
  0},{\bf Q}(t_0,t))$, where, 
\begin{eqnarray}
{\bf Q}(t_0,t)=\int_{t_0}^t {\bf \Phi}(s,t)\  {\bf L}_A \Lambda_A{\bf
  L}_A^T {\bf \Phi}(s,t)^T ds\ .
\label{QQQQ}
\end{eqnarray} 
Thus, the state ${\bf X}(t)$ can be predicted using the Kalman filter,
as per Equations~\eqref{KF1} and~\eqref{KF2}, by defining the process
model as per Equation~\eqref{variant},
\begin{eqnarray*}
{\bf G}(t_0,t)&=&{\bf \Phi}(t_0,t)\ ,
\end{eqnarray*}
and the process noise covariance, ${\bf Q}(t_0,t)$, as per
Equation~\eqref{QQQQ}.  We note the quasi-periodic covariance
functions {\em Step Quasi} (SQM) and {\em Wiener-step Quasi} (WQM),
eigenfunction coefficients are perturbed, as per
Equation~\eqref{perturbation}, at changepoints.  The discrete form of
the Kalman filter can readily predict the value of each coefficient
across a changepoint using the process model, $G^*_j$, and process
noise variance, $Q^*_j$, for each coefficient, $a_{rj}$, as per
Equation~\eqref{perturbation}.

In general, our Kalman filter approach to LFM inference requires a
process model, ${\bf A}_{rj}$, for each eigenfunction coefficient.
Thus, the computational complexity of the prediction step of the
Kalman filter which employs quasi-periodic models increases
quadratically with the number of latent forces $R$, the number of
derivatives used to represent each non-periodic latent force ($N$
in Equation~\eqref{diffbases}) and the number of derivatives used to
represent each time variant eigenfunction coefficient.
Although, our approach supports any quasi-periodic covariance
function, for practical applications, we recommend using the
quasi-periodic covariance functions developed in
Section~\ref{sec:quasi} which are readily convertible to the Markovian
form as per Equation~\eqref{processdyn} and for which only one
variable is required to represent each time varying eigenfunction coefficient.
These quasi-periodic covariance functions are the {\em Continuous
  Quasi model} (CQM), the {\em Step Quasi model} (SQM) and the {\em
  Wiener-step Quasi model} (WQM).

In Sections~\ref{sec:appqueue} and~\ref{sec:appthermal} we determine
the efficacy of our state-space approach to LFM inference on two real
world applications: i) the inference of call centre customer arrival
rates and the tracking of customer queue lengths and ii) the inference
of periodic residual heat dynamics within real homes and the
prediction of internal temperature.  We compare our approaches for the
different periodic and quasi-periodic kernels developed in
Section~\ref{sec:quasi} on the call centre application and
demonstrate the utility of incorporating periodic latent force models
over non-periodic models.  Then we compare our approaches to periodic
and quasi-periodic LFMs to the resonator model in the thermal
application.

\section{MODELLING QUEUES WITH QUASI-PERIODIC ARRIVAL RATES}
\label{sec:appqueue} 
\begin{figure}[ht]
\begin{center}
\includegraphics[width=0.75\textwidth]{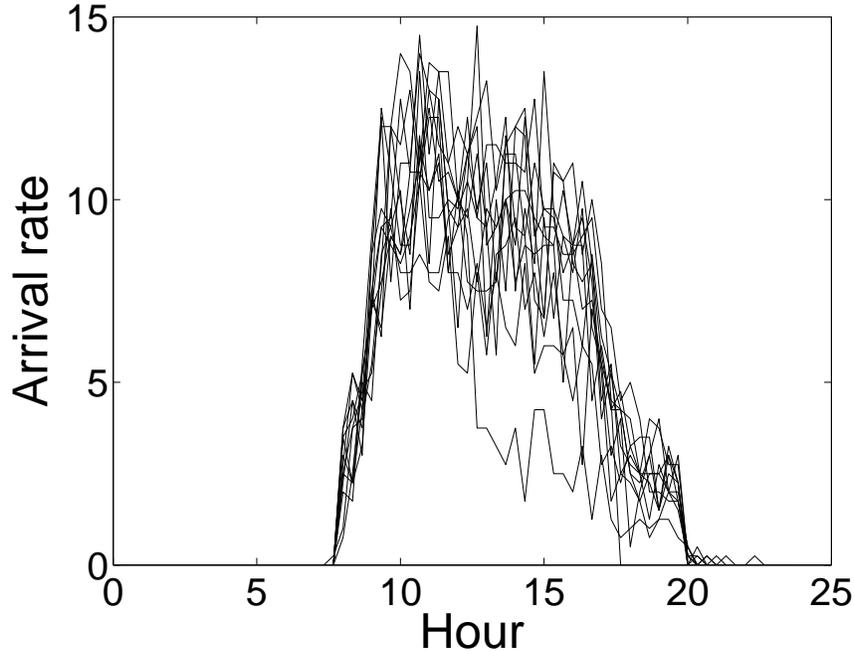}
\end{center}
\caption{\label{arrival_rate} Customer arrival rates (per minute) for
  the same week day (Thursday) over a 11 week period showing the
  quasi-periodic nature of the data.}
\end{figure}

In this section we apply our approach to LFM inference to the dynamics
of telephone queues in call centres as outlined in
Section~\ref{sec:intro} with the aim of tracking the diurnal customer
queue length when different agent deployment strategies are used.  We
use real customer arrival rate data, provided by \citet{feigin06}, in
which the customer telephone arrival rates for a loan company sales
line have been collected every 5 minutes over a three month period
starting from October 2001. The arrival rates during eleven
consecutive Thursdays over this period are shown in
Figure~\ref{arrival_rate}. 

To model the queue dynamics as a latent force model we use
the {\em Pointwise Stationary Fluid Flow Approximation} (PSFFA) for
queues \citep{wang96}.  The PSFFA models the {\em mean} queue length, $L$,
in terms of arrival processes, $\zeta$, using a first order
differential equation,
\begin{eqnarray}
\frac{d L(t)}{dt}=g(L) + \zeta(t)
\label{psffa}
\end{eqnarray}
where, $g$, is a non-negative, non-linear function of the queue
length, $L$. This model is often used to model queues in call centres
where $L(t)$ is the \emph{average} length of the queue at time $t$ and
$\zeta(t)$ is the mean {\em arrival rate}, that is the average rate at which
customers join the queue \citep{wang96}.  The PSFFA is a first order,
non-linear differential equation.  Ignoring the non-linearity of $g$
for now, we see that Equation~\eqref{psffa} is of the form of
Equation~\eqref{initiallfm}.  Thus, Equation~\eqref{psffa} is an
example of a LFM in which the queue length, $L$ in
Equation~\eqref{psffa} is the target process, ${\bf z}$ in
Equation~\eqref{initiallfm} and the customer arrival rate, $\zeta(t)$
in Equation~\eqref{psffa}, is the sole latent force, ${\bf u}$, in
Equation~\eqref{initiallfm}.  Consequently, we apply our approach to
LFM inference to the tracking of queue lengths using the PSFFA.

We consider the M/M/1 queue as it corresponds exactly to the customer
arrival process, $\zeta$, which is Poisson and the service time is
arbitrarily distributed with successive service times being
independent and identically distributed.~\footnote{The term `M/M/1' is
  Kendall's queue classification notation \citep{kendall53}
  corresponding to a stochastic process whose state-space is the set
  $\{0,1,2,3,...\}$ where, in our case, the value corresponds to the
  number of customers in the system.}  Service times have an
exponential distribution with parameter $\Omega$ in the M/M/1 queue.
A single server serves customers one at a time from the front of the
queue, according to a first-come, first-served basis. When the service
is complete the customer leaves the queue and the number of customers
in the system reduces by one.  The queue buffer is of infinite size,
so there is no limit on the number of customers it can contain.

The PSFFA allows us to represent this M/M/1
system via the following differential equation for the mean queue
length, $L$ \citep{wang96},
\begin{eqnarray}
\frac{d L(t)}{dt}=-\Omega(t)\left(\frac{L(t)}{1+L(t)}\right) +
\zeta(t)\ ,
\label{queuedyn}
\end{eqnarray}  
where $\Omega(t)$ is the mean queue {\em service rate}.  We use this
model to simulate the true queue length, $L$, using the real customer
arrival rate, $\zeta$, from data provided by \citet{feigin06} and
realistic service rate profiles, $\Omega$.  Measurements of the
instantaneous customer queue length, $y(t^*)$, are generated for
times, $t^*$, during the day,
\begin{eqnarray*}
y(t^*)=L(t^*)+\epsilon(t^*)\ ,
\end{eqnarray*}
where, $t^*$, are sufficiently spaced so that $\epsilon(t^*)$ is
zero-mean, i.i.d. Gaussian.

To recover the customer arrival rate from the measured queue length we
assume that the mean arrival rate, $\zeta$, is drawn from a Gaussian
process, $\zeta\sim \mathcal{GP}(0,K_{\text{arrival rate}})$, where
$K_{\text{arrival rate}}$ is the arrival rate process covariance
function.  Note that the arrival rate can be positive or negative.
Negative arrival rates correspond to customers who leave the system
without being served.  We choose $K_{\text{arrival rate}}$ to be
either the first order Mat\'ern kernel, a periodic first order
Mat\'ern kernel as per Equation~\eqref{matern} or a quasi-periodic
first order Mat\'ern kernel utilising a CQM, SQM or WQM kernel, as per
Section~\ref{sec:quasi}.

As per Equation~\eqref{augstatedyn}, the augmented state-vector, ${\bf
  X}(t)$, is ${\bf X}(t)=\left(L(t),{\bf A}^T(t)\right)^T$ where ${\bf
  A}(t)$ are the eigenfunction weights corresponding to the periodic
latent force covariance functions, as per Equation~\eqref{augstateA}.
Unfortunately, the transition dynamics in Equation~\eqref{queuedyn}
are nonlinear.  However, if we assume that the mean, $\bar{L}(t_0)$,
of $L$, conditioned on the measurements up to time $t_0$ is a good
approximation for $L$ over the entire, yet small, interval $[t_0,\ t]$
then we can rewrite Equation~\eqref{queuedyn} as a locally linear
model,
\begin{eqnarray}
\frac{d L(t)}{dt}\approx-\frac{\Omega(t)}{1+\bar{L}(t_0)}\
L(t) +
\zeta(t)\ ,
\label{queuedyn2}
\end{eqnarray} 
over $[t_0,\ t]$.  This model then has the appropriate form for
inclusion within the Kalman filter. In our experiments predictions are
made over $2$ minute time intervals.  This time interval is chosen so
that Equation~\eqref{queuedyn2} is a stable local approximation to the
queue dynamics.  We shall call this KF algorithm \textbf{LFMwith} as
it contains a GP model of the arrival rate process.  We use maximum
likelihood to obtain the GP hyperparameters and the model parameters
which are the Mat\'ern output and input scales and the observation
noise variance.  The cycle period is fixed at 24 hours.

The efficacy of our customer queue model is evaluated by training the
model using data over three full consecutive Thursdays and then
tracking the mean queue length over the following Thursday.  The queue
length observations are made every 40 minutes during the training
period and every three hours during the fourth day tracking phase.
The longer tracking interval is specifically chosen to test the
predictive power of our model with sparse observations.  The efficacy
of our approach to LFM inference for even longer term predictions
(that is, day ahead predictions) is explored in Section~\ref{sec:appthermal}.

\begin{figure}[!ht]
\begin{center}
\begin{tabular}{cc}
\includegraphics[width=0.45\textwidth]{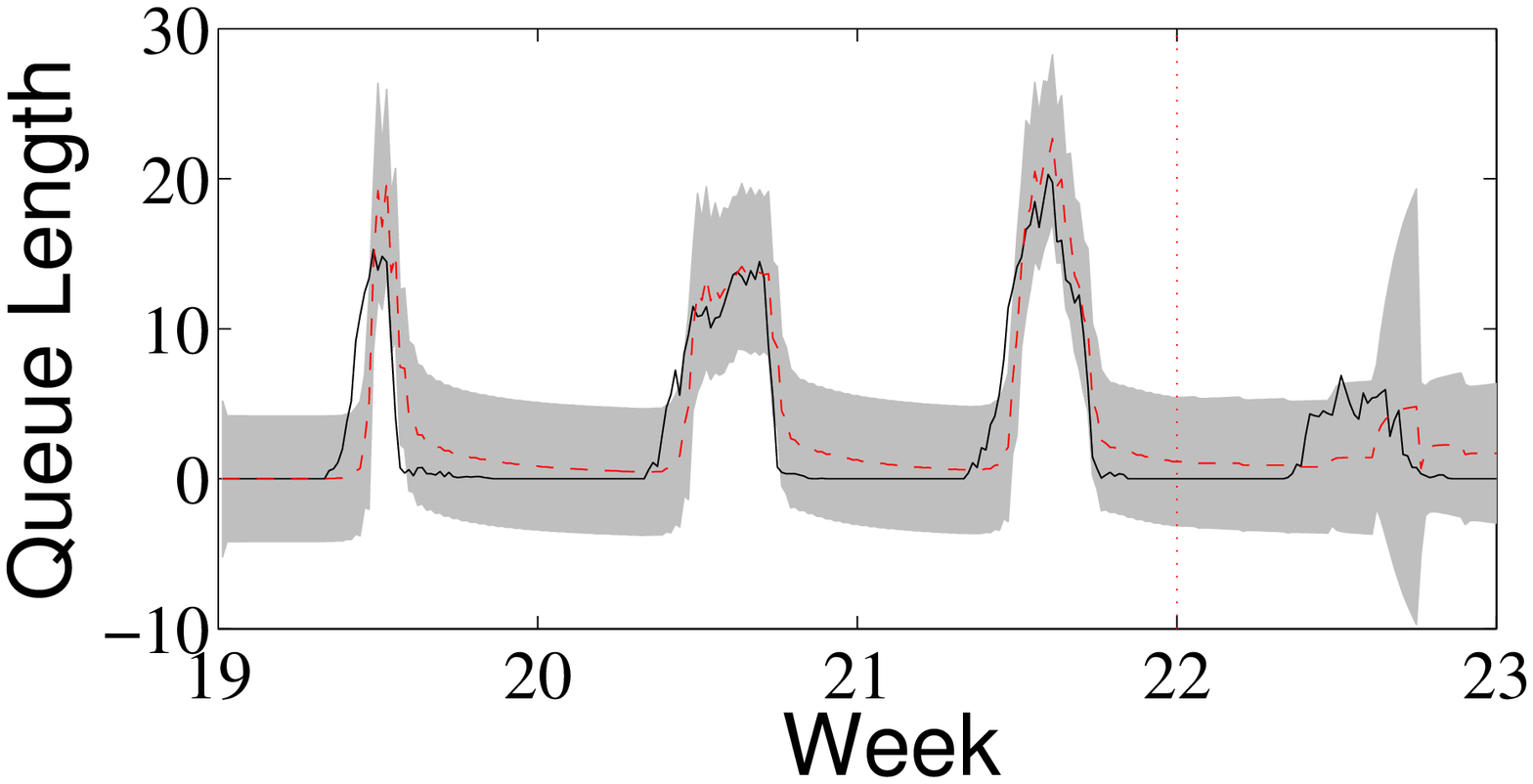}
& \includegraphics[width=0.45\textwidth]{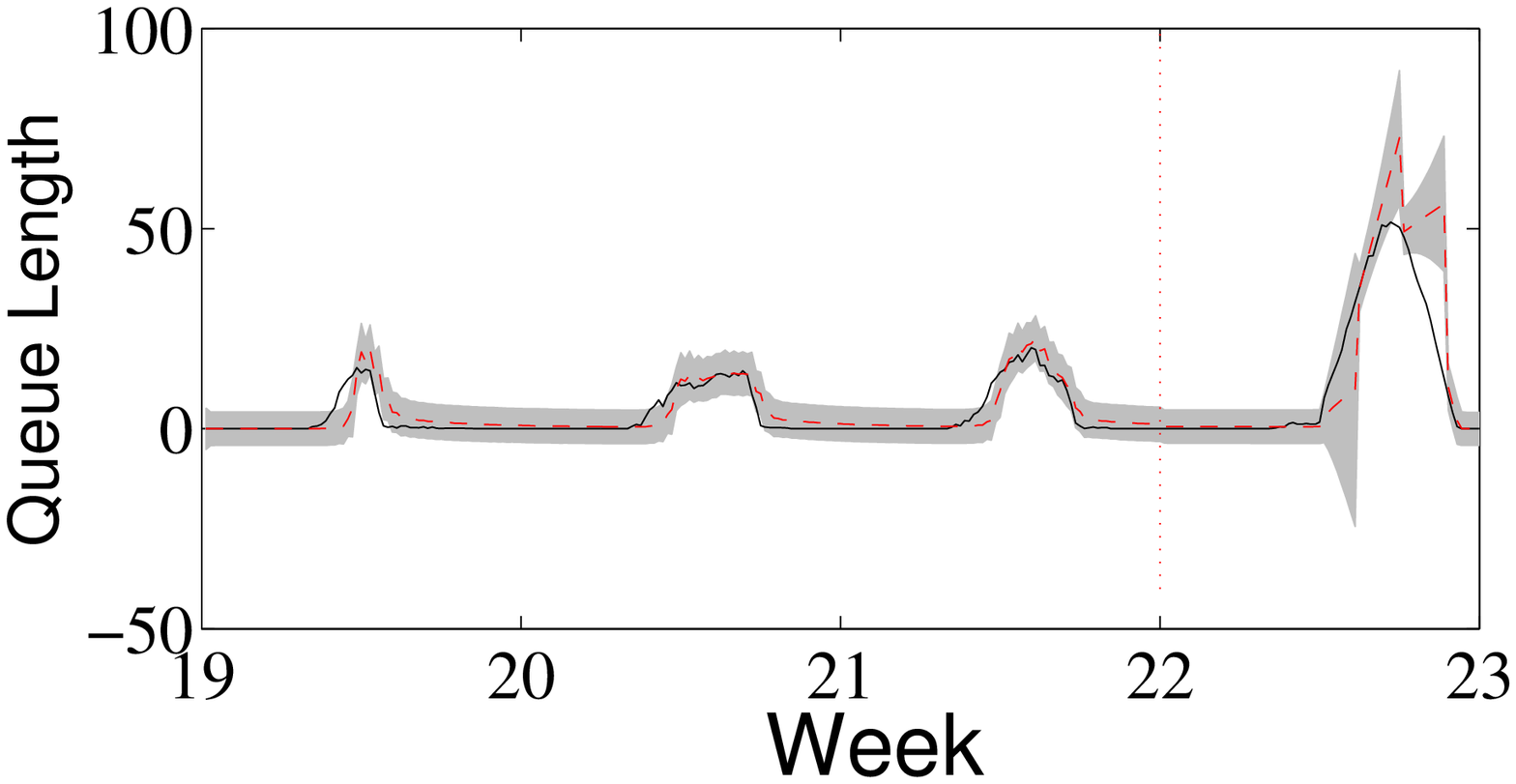}\\
(a) \textbf{Hart}: $\Omega$=10 & (b) \textbf{Hart}: $\Omega$=10, 15 and then 5\\ \ \\
\includegraphics[width=0.45\textwidth]{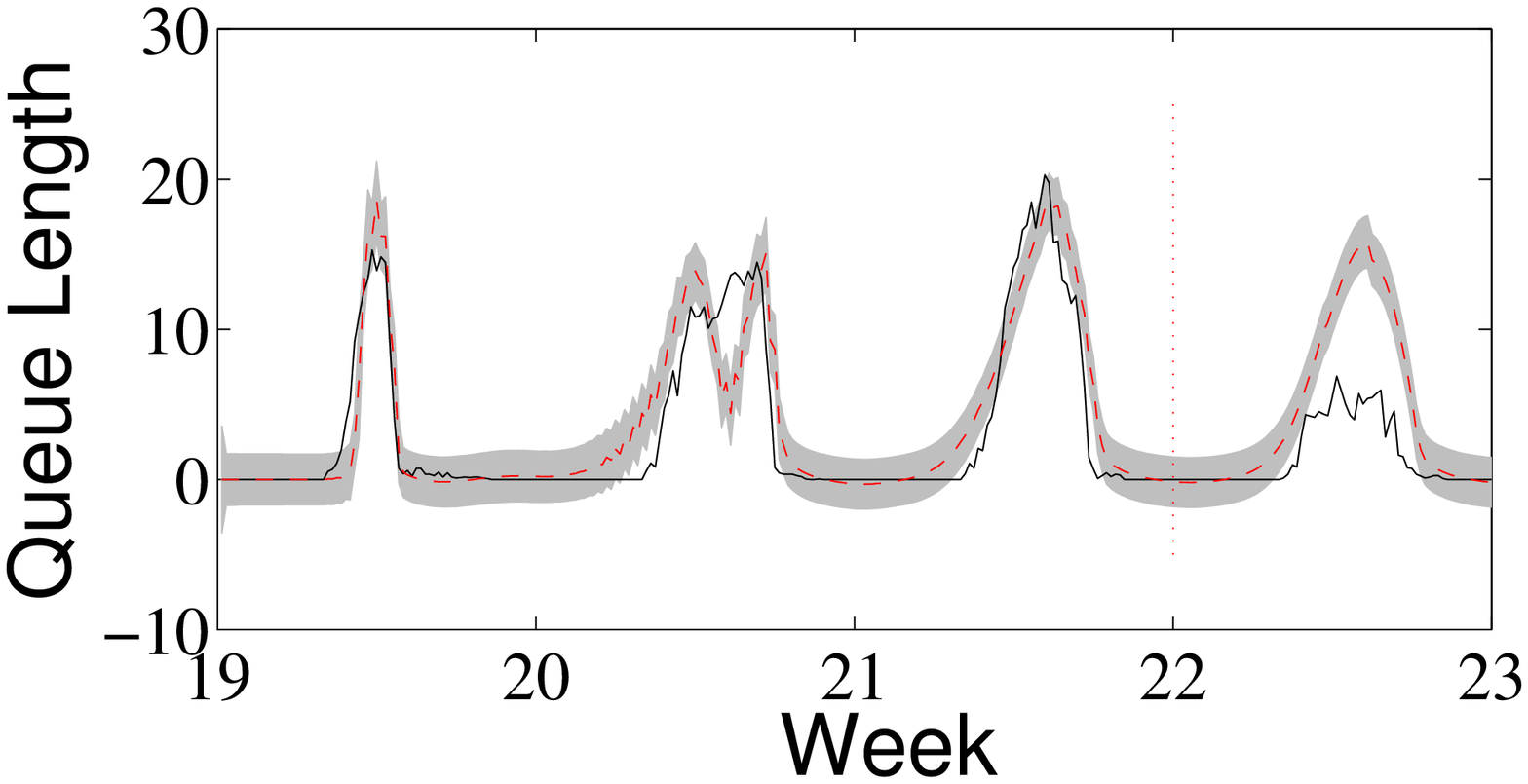}
& \includegraphics[width=0.45\textwidth]{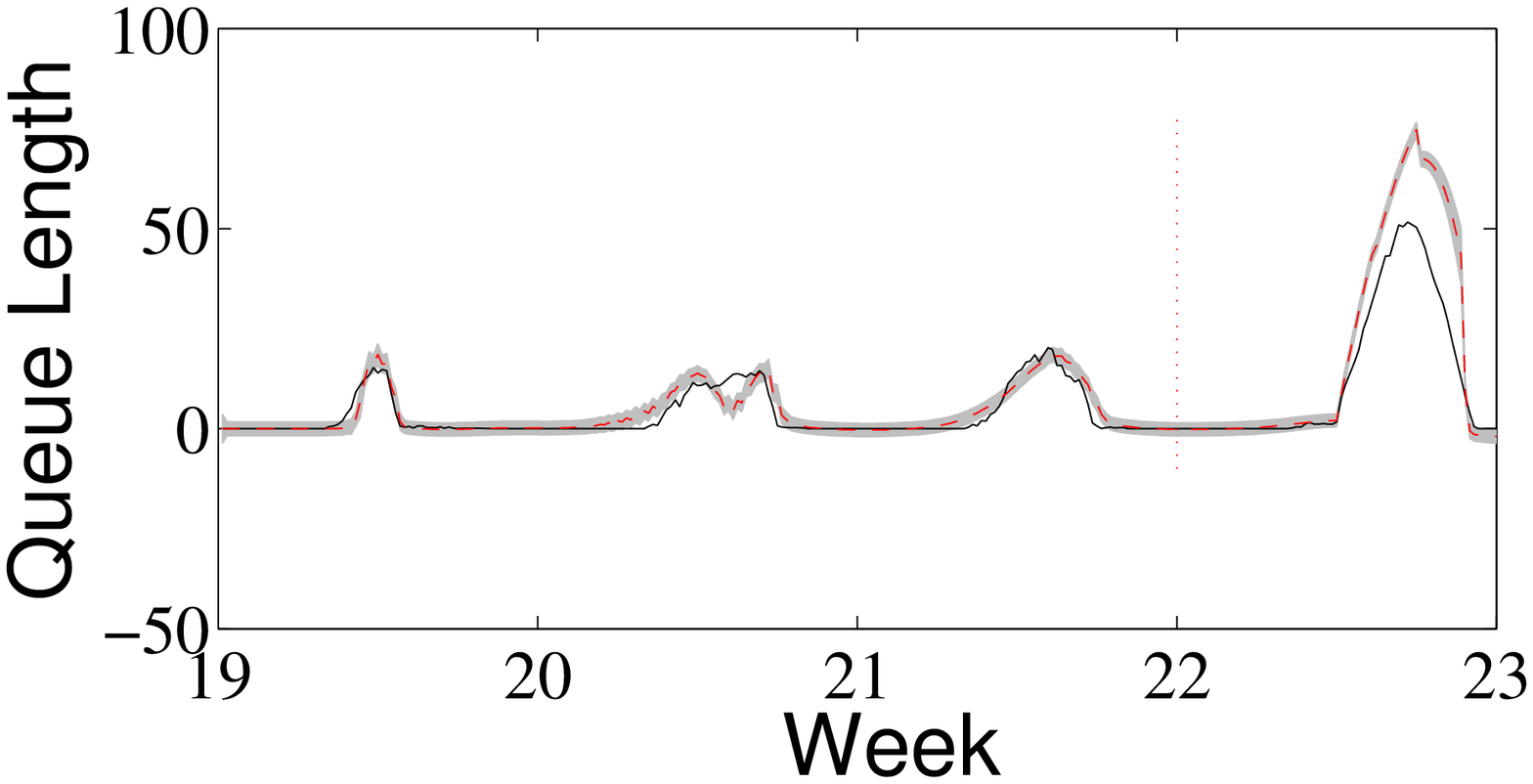}\\
(c) \textbf{LFMwith}: $\Omega$=10 & (d) \textbf{LFMwith}: $\Omega$=10, 15 and then 5\\ \ \\
\includegraphics[width=0.45\textwidth]{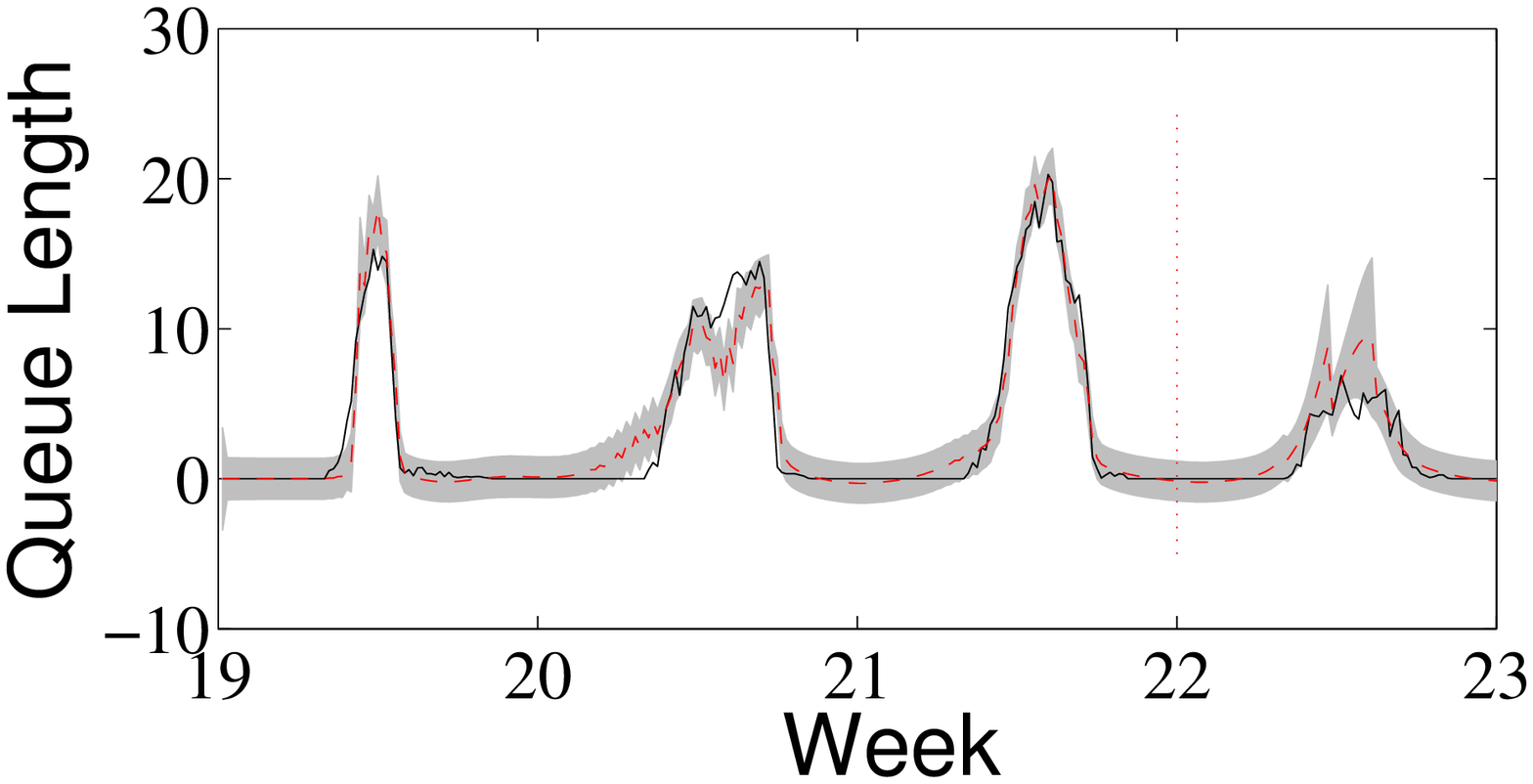}
& \includegraphics[width=0.45\textwidth]{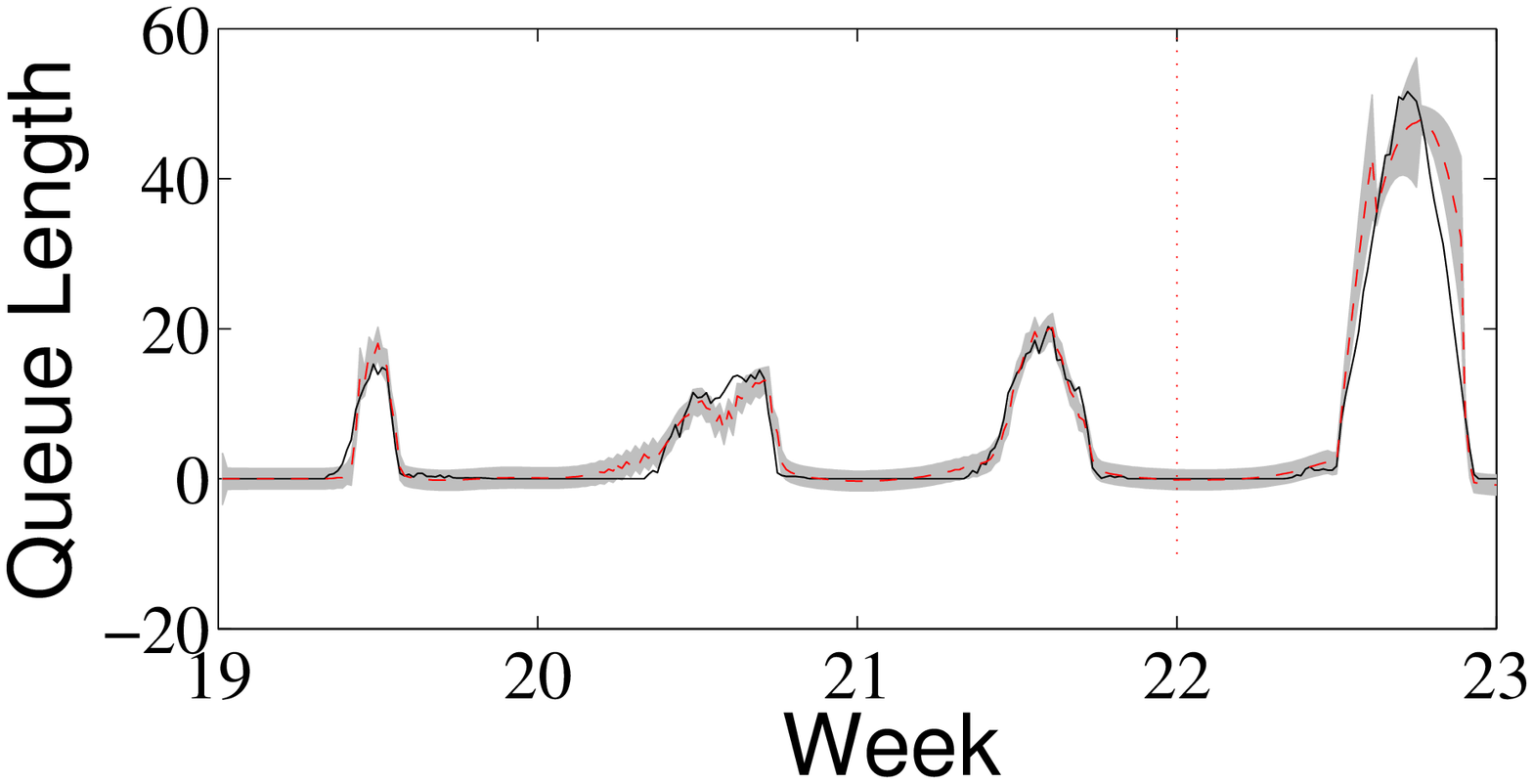}\\
(e) \textbf{LFMquasi (SQM)}: $\Omega$=10 & (f) \textbf{LFMquasi (SQM)}: $\Omega$=10, 15
and then 5
\end{tabular}
\end{center}
\caption{\label{queue_example} Call centre customer queue length over
  four consecutive Thursdays.  The first three days of data are used to train
  the model.  The fourth day is tracked. The 1st standard deviation
  confidence interval is shown (grey).  The solid black line shows the
  ground truth.  The left column of plots shows the results for a
  fixed service rate, $\Omega=10$, for both training and test phases.
  The right column of plots shows the results for a fixed training
  service rate, $\Omega=10$, and a variable test service rate of
  $\Omega=15$ for the first half of the fourth day and $\Omega=5$ for
  the second half.}
\end{figure} 

We also introduce four further algorithms to empirically demonstrate
the importance of using periodic and quasi-periodic latent force
models in our domain and to demonstrate the efficacy of our algorithm
at tracking customer queue lengths.  Three of these algorithms use
quasi-periodic latent force models. \textbf{LFMquasi (CQM)},
\textbf{LFMquasi (SQM)} and \textbf{LFMquasi (WQM)} algorithms use the
Continuous Quasi-periodic model, the Step Quasi-periodic model and the
Wiener-step periodic model respectively, described in
Section~\ref{sec:quasi}, to model the arrival rates.  These models are
identical to the periodic model used in \textbf{LFMwith} except that
the correlation between cycles is reduced by the quasi-periodic
kernel.  The most likely hyperparameters are used for the SQM, CQM and
WQM kernels.  The changepoints required by the SQM and WQM
quasi-periodic latent force models are set to midnight for all
cycles. We also implement Hartikainen's
algorithm \citep{hartikainen10} for sequential inference which uses
the M/M/1 model described above and a non-periodic first order
Mat\'ern kernel for the customer arrival rate process (\textbf{Hart})
to demonstrate the performance of a non-periodic model.

An example run of our algorithms is shown in
Figure~\ref{queue_example} and this shows the ground truth queue
lengths (in black) and first standard deviation estimates of the queue
lengths using \textbf{Hart}, \textbf{LFMwith} and \textbf{LFMquasi
  (SQM)} latent force models. This figure shows the tracked queue
length over four days.  The LFM parameters are learned using the first
three days of data only.  The fourth day is tracked without any
further learning of the LFM model parameters.  The left column of
plots shows the queue length estimates for a fixed service rate,
$\Omega=10$, applying to both training and tracking phases.  The right
column of plots shows the estimates for a fixed training service rate,
$\Omega=10$, and a variable test service rate of $\Omega=15$ for the
first half of the fourth day and $\Omega=5$ for the second half.  By
testing the algorithm with variable service rates, we are able to test
the efficacy of the algorithms at both reproducing the training data
and at making predictions in regimes not encountered during the
training period. Clearly the quasi-periodic model is the most
accurate, in this case, with a RMSE of $1.9$ compared to $2.3$ and
$4.6$ for \textbf{Hart} and \textbf{LFMwith}, respectively.

To fully test the accuracy of the inferred residual model we evaluated
the RMSE and expected log likelihood of the predicted average queue
length for each day and for each algorithm over $11$ days.  Firstly,
the service rate was held constant throughout at an arbitrary value of
$\Omega=10$.  The results are summarised in Table~\ref{queue_tab}.
Clearly, the RMSE is lower for the quasi-periodic models but their
expected log likelihoods are larger than the periodic model indicating
the superiority of the quasi-periodic models.

\begin{table}[ht]
  \caption{\label{queue_tab} Day ahead tracking: Queue length RMSE and
    expected log likelihood (ELL) for
    periodic, quasi-periodic and non-periodic arrival rate
    models. Both training and test epochs have the same fixed service rate $\Omega=10$.}
\begin{center}
\begin{tabular}{|c|c|c|}
\hline Method & RMSE & ELL \\ \hline\hline
LFMquasi (SQM) & $  4.6 \pm   2.2$ & $  -146 \pm     22$\\
LFMquasi (CQM) & $  4.4 \pm   1.3$ & $  -142 \pm     15$\\
LFMquasi (WQM) & $  {\bf 1.8 \pm   0.2}$ & $ {\bf -139 \pm     20}$\\
LFMwith & $  2.2 \pm   0.4$ & $  -276 \pm     60$\\
Hart & $ 10.6 \pm   5.9$ & $  -209 \pm     29$\\
\hline
\end{tabular}
\end{center}
\end{table}

In the final experiment in this section we demonstrate the ability of
our approach to make inferences in regimes where data is absent.  This
is a powerful and useful property of Gaussian process models.
Specifically, to plan future staffing requirements the call centre
manager needs to predict the impact that a novel service rate will
have on future queue lengths given predicted customer arrival rates.
However, the service centre may not have utilised this staffing
profile to date.  In this case, for illustrative purposes, we assume
that the staff profile to date has been constant during working hours
with a fixed service rate, $\Omega=10$.  However, the service manager
has noticed a significant queue of customers forming in the morning
and then relatively few customers arriving in the
afternoon. Consequently, the service manager contemplates employing a
variable staffing profile and hiring more staff during the first half
of the day, so that the service rate increases to $\Omega=15$, and
then retaining fewer staff in the afternoon, so that the service rate
drops to $\Omega=5$.

To determine the efficacy of our approach at predicting the impact of
variable staffing profiles given only data from constant staffing
profiles we repeated the experiment above with a fixed service rate,
$\Omega=10$ during training and a variable service rate during the
test period.  In this case, we chose a high service rate of
$\Omega=15$ for the first half of the test day and then a low rate,
$\Omega=5$ over the second half. The RMSE and expected loglikelihood
are shown in Table~\ref{queue_tab2}.  Again, the quasi-periodic models
have similar efficacy and produce the most accurate estimates of the
customer queue length in this case with an RMSE of $3.3$ compared to
$11.6$ and $5.7$ for \textbf{Hart} and \textbf{LFMwith}, respectively.
We note that, for both experiments, the \textbf{LFMquasi (SQM)},
\textbf{LFMquasi (CQM)}, \textbf{LFMquasi (WQM)} and \textbf{LFMwith}
used fewer than $28$, $28$, $20$, $28$ basis functions, respectively,
to represent the arrival rate process.~\footnote{The actual number of
  basis functions used varied between runs.}  Consequently, our LBM
Kalman filter approach to LFM tracking is computationally efficient.
 
\begin{table}[ht]
  \caption{\label{queue_tab2} Queue length RMSE and expected log
    likelihood (ELL) for
    periodic and quasi-periodic arrival rate models.  Training over
    three days with a fixed service rate $\Omega=10$.  The test day had a
    service rate of $\Omega=15$ for the first half of the day followed by
    $\Omega=5$ for the remainder.}
\begin{center}
\begin{tabular}{|c|c|c|}
\hline Method & RMSE & ELL \\ \hline\hline 
LFMquasi (SQM) & $  7.2 \pm   1.8$ & $  -183 \pm     21$\\
LFMquasi (CQM) & $ 15.2 \pm   4.0$ & $  -205 \pm     24$\\
LFMquasi (WQM) & $  {\bf 3.3 \pm   0.6}$ & $ {\bf -152 \pm     15}$\\
LFMwith & $  5.7 \pm   1.2$ & $  -725 \pm    301$\\
Hart & $ 11.6 \pm   1.1$ & $  -305 \pm     49$\\
\hline
\end{tabular}
\end{center}
\end{table}
 
In the next section we evaluate our approach to LFM inference for
longer term predictions than those considered in the call centre
application.  We shall demonstrate that our approach can exploit the
quasi-periodic nature of the latent force to project far forward in
time an accurate estimate of the force.  Consequently, we shall see
that our approach is effective at performing day ahead predictions of
temperatures in the home using differential thermal models and
non-parametric models of the residual heat within the home.

\section{MODELLING THE THERMAL DYNAMICS OF HOME HEATING}
\label{sec:appthermal}

In this section we apply our approach to LFM to the thermal modelling
problem outlined in Section~\ref{sec:intro}.  We assume that the
differential equation governing the thermal dynamics within a home is
given by,
\begin{eqnarray}
\frac{dT_\text{int}(t)}{dt}=\alpha(T_\text{ext}(t)-T_\text{int}(t))+
\beta E(t)+R(t)\ ,\label{thermalmodel}
\end{eqnarray}
where $T_\text{int}$ and $T_\text{ext}$ are the internal temperature
within the home and the onsite ambient external temperature
respectively in $^{\circ}\mathrm{C}$
\citep{madsenmodel,rogers11,ramchurn12}. $E(t)$ represents the
thermostat control output at time $t$ ($E(t) \in \{0, 1\}$), $\beta$
represents the thermal output of the heater and $\alpha$ is the
leakage coefficient to the ambient environment. In this model
$T_\text{ext}(t)$ and $E(t)$ are known latent forces for the LFM in
Equation~\eqref{thermalmodel}.  $R(t)$ is the residual generated by
latent forces which are not captured by the differential thermal
model, such as heat generated by solar warming and lags in the heating
system. These are completely unknown a priori but, since they are
expected to exhibit periodic behaviour, a periodic Mat\'ern kernel
prior is used to model them.

We assume that $T_\text{ext}(t)$ at times $t$ and $t_0$ can be
modelled by a non-periodic GP prior
$\tt{\text{Mat\'ern}(|t-t_0|,0.5,\sigma_\text{ext},l_\text{ext})}$.
We choose the Mat\'ern kernel as this imposes continuity in the
function but imposes no strong assumptions about higher order
derivatives.  However, our approach can be applied to Mat\'ern
functions of higher smoothness if required. The state vector, ${\bf
  X}$, as per Equation~\eqref{prediction}, comprises the internal
temperature, the external temperature and its derivative and
eigenfunction coefficient weights, ${\bf A}$, for our sparse basis
model of the residual as per Equation~\eqref{augstateA}. We model the
residual process by a periodic Mat\'ern kernel,
$\tt\text{Mat\'ern}(|\sin(\pi \tau /D)|,\nu,\sigma,l)$ with order
$\nu=1/2$, smoothness, $l$, and $D$ set to correspond to a daily
period.  Again, we choose the Mat\'ern for the same reasons as above.
The residual is represented via $J$ basis functions
$\left(\phi_1(\theta),\ldots, \phi_J(\theta)\right)$ corresponding to
Equation~\eqref{phi}, where $\theta$ is the periodic phase as
described in Section~\ref{sec:sslfm}.  The augmented state-vector,
${\bf X(t)}$, is ${\bf X}(t)=\left(T_\text{int}(t),T_\text{ext}(t),
  \frac{T_\text{ext}(t)}{dt},{\bf A}(t)^T\right)^T$ and the continuous
time dynamic model corresponding to Equation~\eqref{inhomo} is,
\begin{eqnarray*}
\frac{d{\bf X}(t)}{dt}= \begin{pmatrix}{\bf F}_a & {\bf m}(t)\\ {\bf 0} & {\bf
    F}_{\bf A}\end{pmatrix}{\bf
  X}(t)+{\bf L}_{\bf A}{\bf \omega}_{\bf A}(t)+\beta {\bf E}(t)\ ,
\end{eqnarray*} 
where ${\bf E}(t)=(E(t),0,\ldots,0)^T$ is the same size as ${\bf X}(t)$.

We will now describe the role of each term in the dynamic model.
Within the transition model, ${\bf F}_a$ captures the temperature
gradient components of Equation~\eqref{thermalmodel} and the Mat\'ern
driving forces for the external temperature,
\begin{eqnarray*}
{\bf F}_a=\begin{pmatrix} -\alpha & \alpha & 0\\
0 & 0 & 1 \\
0 & -\rho_\text{ext}^2 & -2\rho_\text{ext}
\end{pmatrix}\ ,
\end{eqnarray*}
with $\rho_\text{ext}=2/l_\text{ext}$.  The derivative of the external
temperature is represented in the state vector so that the Mat\'ern
latent force kernel can be encoded within the Kalman filter as
summarised in Section~\ref{sec:statebasedapproaches} and described in
detailed in \citet{hartikainen10}.  We set the order of this Mat\'ern
covariance function to $\nu=3/2$ as the external temperature process
is relatively smooth.  The matrix ${\bf m}$ captures the residual heat
contribution to the internal temperate and depends on the choice of
the residual model covariance function, $K$, in Equation~\eqref{m}.
Further, the matrix ${\bf F}_A$ models the dynamics of each periodic
residual heat process and this also depends on the choice of residual
model covariance function, as per Equation~\eqref{processdyn}.  The
corresponding discrete form of the Kalman filter, as per
Equation~\eqref{Xavar}, is evaluated over $10$ minute time intervals,
$[t_0,\ t]$.  This time interval is chosen to coincide with the heater
on/off control cycle.

The Kalman filter is initialised with known current temperature
values.  The initial covariance is block diagonal with a diagonal
matrix over the temperature components \citep[including the solution to the
appropriate Riccati equation for the external temperature Mat\'ern
model presented in][]{hartikainen10} and a diagonal covariance
over the residual model weights corresponding to the periodic Mat\'ern
residual process.  The covariance for the model weights is obtained
using the periodic Mat\'ern prior and corresponding eigenfunctions as
described in Section~\ref{sec:statebasedapproaches}.
 
\begin{figure}[ht]
\centerline{
\includegraphics[width=1.1\textwidth]{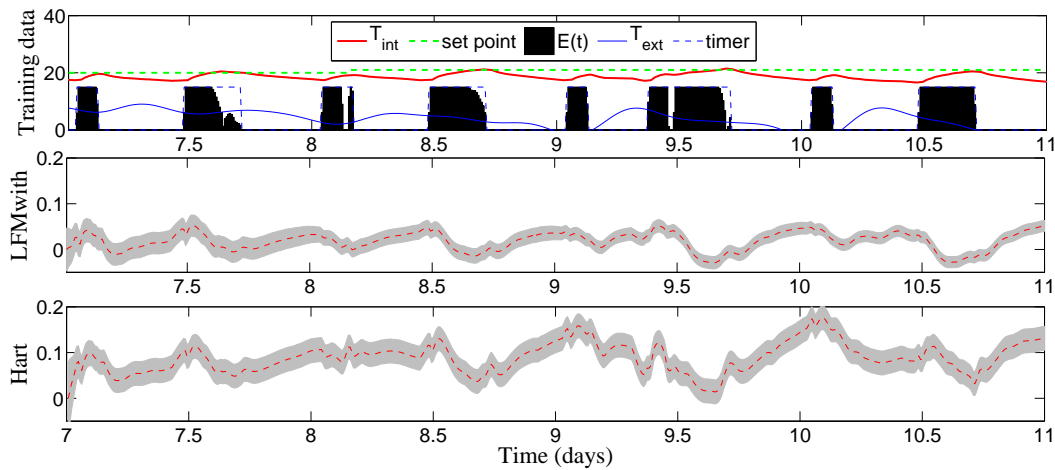}
}
\caption{\label{train} Internal and external temperature, thermostat
  set-point and heater activation for a four day training period
  (upper).  Also shown is the residual sequentially inferred using
  \textbf{LFMwith} (central) and the \textbf{Hart} (lower) algorithms.
  The 1st standard deviation confidence interval is shown (grey).}
\end{figure}

When tracking the internal room temperature we know the state of the
heater, that is, whether it is ``on'' (that is, $E(t)=1$) or ``off''
(that is, $E(t)=0$), at each point in time.  However, when predicting
the internal room temperature a full day ahead the times at which the
heater will switch on or off will not be known in advance.
Uncertainty in the future controller behaviour arises because the
heater behaviour depends on the internal room temperature and the
predicted internal room temperature will be uncertain.  The heater
will be on if the room temperature is below the set-point or off if
above the set-point.  In order to accommodate the uncertainty in the
heater switching process and, as the control output is binary, then
prediction is performed using the Rao-Blackwellised Particle filter
\citep[RBPF,][]{doucet00}.  The RBPF uses a set of particles to
represent the many possible states of the system (the internal
temperature and residual).  The corresponding on/off control output is
determined for each particle from the value of the internal
temperature associated with that particle and the set point.  For each
particle the Kalman filter is used to predict the room temperature
conditioned on the control output for that particle which is held
constant for each 10 minute interval.  For each 10 minute interval
there are RBPF particles corresponding to the control output being
``on'' or ``off'' over that interval.  Each particle also has a prior
Gaussian distribution over ${\bf X}$ and the Kalman filter is used to
predict the state ${\bf X}$ at the end of the current interval
conditioned on the binary value of the heater for that particle.  A
new set of particles is then generated by taking each particle in
turn, sampling the posterior of the internal temperature, conditioning
the posterior on that internal temperature sample and then assigning
the heater state according to the controller (set point minus the
internal temperature when the heater is primed).  This procedure is
iterated to predict over the entire day ahead. With $P$ particles and
cardinality $C$ of ${\bf X}$, the complexity of the prediction phase
scales as $\mathcal{O}(C^2 TP)$ over $T$ time steps.  Following
nomenclature in the call centre theory application in
Section~\ref{sec:appqueue}, we shall call this RBPF algorithm
\textbf{LFMwith$^+$} as it contains a GP model of the residual heat
process.  However, we have added the superscript `{\tt +}' to denote
that LFM inference is performed by the RBPF. We use maximum likelihood
to obtain the model parameters for the thermal model, $\{\alpha,\
\beta\}$, the GP hyperparameters, $\{\sigma,\ l,\ \sigma_\text{ext},\
l_\text{ext}\}$ and the observation noise variance.  We note that, if
the set-point process is also uncertain but a distribution over the
future set-point process is known then the RBPF particles can be drawn
from the on/off control distribution and the set-point distribution.
We do not examine the case of uncertain set-point values in this
paper.
 
We also implement four further algorithms to empirically demonstrate
the importance of using a periodic residual model in our domain and to
demonstrate the efficacy of our algorithm at predicting internal
temperatures.  Three of these algorithms use quasi-periodic latent
force models. \textbf{LFMquasi (SQM)$^+$}, \textbf{LFMquasi (CQM)$^+$}
and \textbf{LFMquasi (WQM)$^+$} algorithms use the step quasi-periodic
model, the continuous quasi-periodic model and the Wiener
quasi-periodic model, respectively, described in
Section~\ref{sec:quasi}, to model the residual driving forces.  These
models are identical to the periodic model used in
\textbf{LFMwith$^+$} except that the correlation between cycles is
reduced by the quasi-periodic kernel.  Again, these models use the
same quasi-periodic covariance functions as their counterparts in the
call centre application in Section~\ref{sec:appqueue} and, again, we
have added the superscript `{\tt +}' to denote that LFM inference is
performed by the RBPF.  A fifth algorithm, \textbf{LFMwithout$^+$},
assumes that there is no residual heat within the home.  This
algorithm is an instance of \textbf{LFMwith$^+$} with no periodic
latent force basis functions in the state vector.  We also implement
Hartikainen's algorithm \citep{hartikainen10} for sequential inference
which uses the thermal model described above and a non-periodic
Mat\'ern kernel for the residual (\textbf{Hart$^+$}).  To accommodate
the binary thermostat controller within \textbf{Hart$^+$} we use the
RBPF, as described above, but with Hartikainen's Kalman filter
formalism.

We also implement a recent version of the resonator model
\citep{solin13} which we call the \textbf{Resonator$^+$}.  The
resonators are defined via the second order differential equation, as
per Equation~\eqref{resonator2}, which includes a decay term with
fixed frequency and decay coefficients.  We chose to implement this
version of the resonator model as opposed to the time varying
frequency version \citep{sarkka12} as this version of the resonator
model is completely developed in the literature and inferring the
coefficients using maximum likelihood techniques has been thoroughly
tested \citep{solin13}. In order to undertake a fair comparison between
the performance of the resonator model and our eigenfunction
approaches we choose the number of resonators and eigenfunctions
to be the same.  Further, we infer the most likely resonator
frequencies and decay coefficients using the same Nelder-Mead
optimisation algorithm implemented in our eigenfunction approach.  As
the residual process is quasi-periodic with period, $D$ (corresponding
to one day), we initialise the resonator frequencies to be distinct
and contiguous multiples of $1/D$. As with all the LFM algorithms
above, day ahead predictions with the resonator model are performed by
the RBPF.

We collected two data sets from two different homes in January 2012
recording the internal temperature, $T_\text{int}$, the external
temperature, $T_\text{ext}$, the thermostat set point and the heater
activity, $E$, at one minute intervals. Each data set comprises
fourteen consecutive days of data.  We label these data sets {\tt
  data1} and {\tt data2}.  For each home four complete consecutive
days of the data set are chosen to train each algorithm.  We then
predict the internal temperature, $T_\text{int}(t)$, over the next
full day.  With $14$ days of data for each dataset, $10$ full day
predictions can be made for each dataset with each
algorithm.  Note that both data sets have thermostat set point changes
that require predictions to be made in regimes in which the algorithms
have not been trained.

\begin{figure}[ht]
\begin{center}
\begin{tabular}{ccc} 
LFMwithout$^+$ & LFMwith$^+$\\
\includegraphics[width=0.35\textwidth]{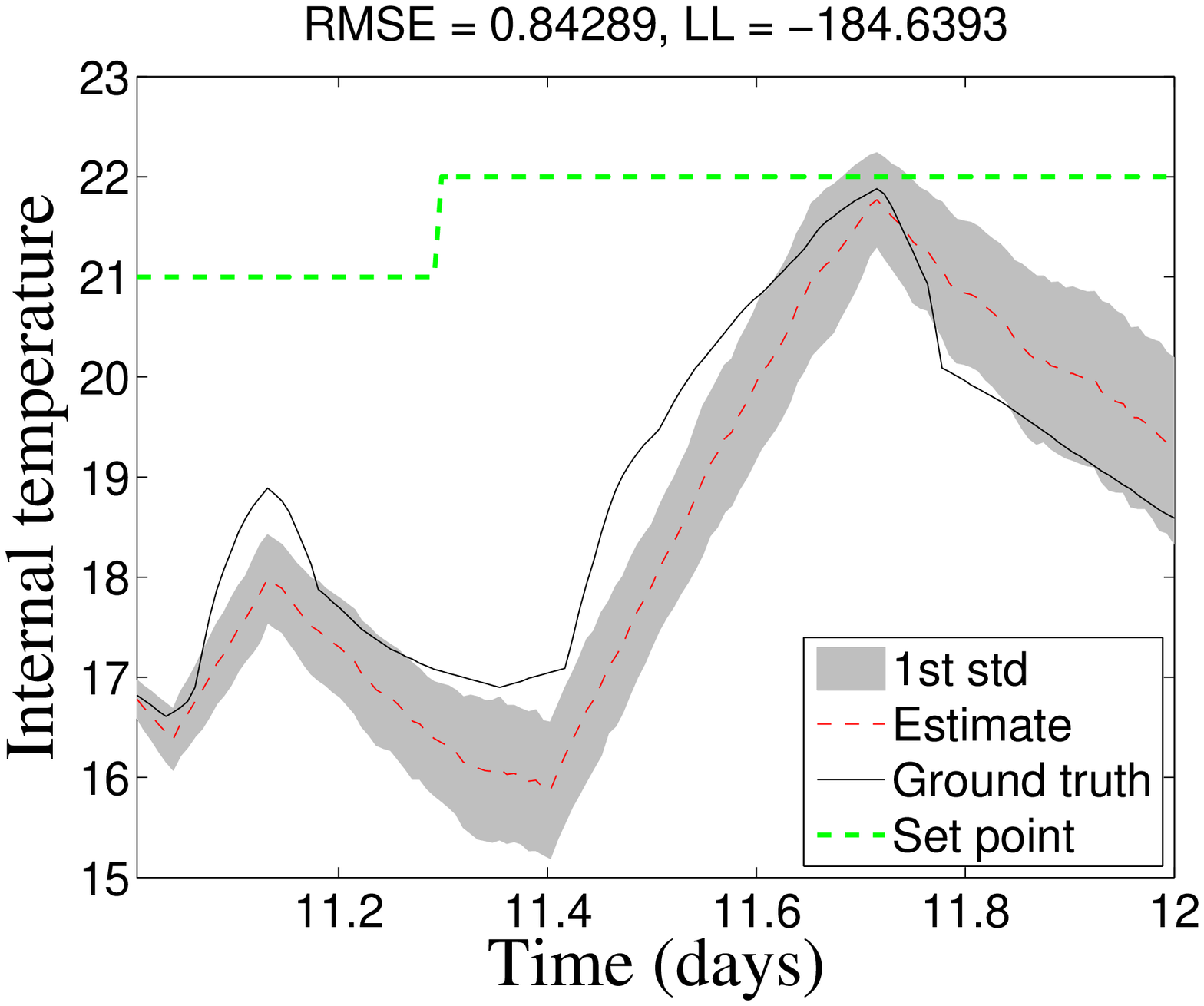} &
\includegraphics[width=0.35\textwidth]{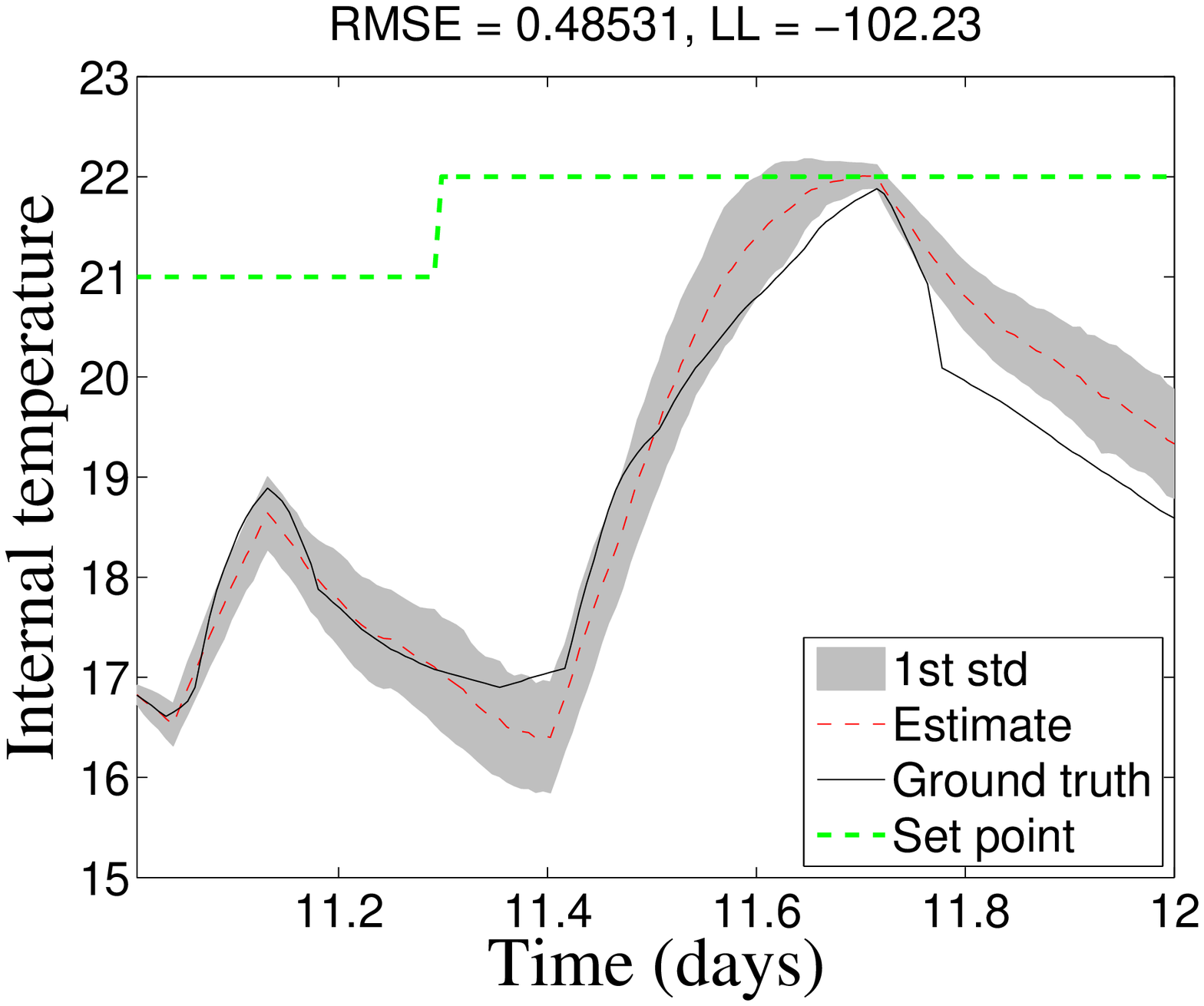}\\
\  \\ 
LFMquasi (SQM)$^+$ & LFMquasi (CQM)$^+$\\
\includegraphics[width=0.35\textwidth]{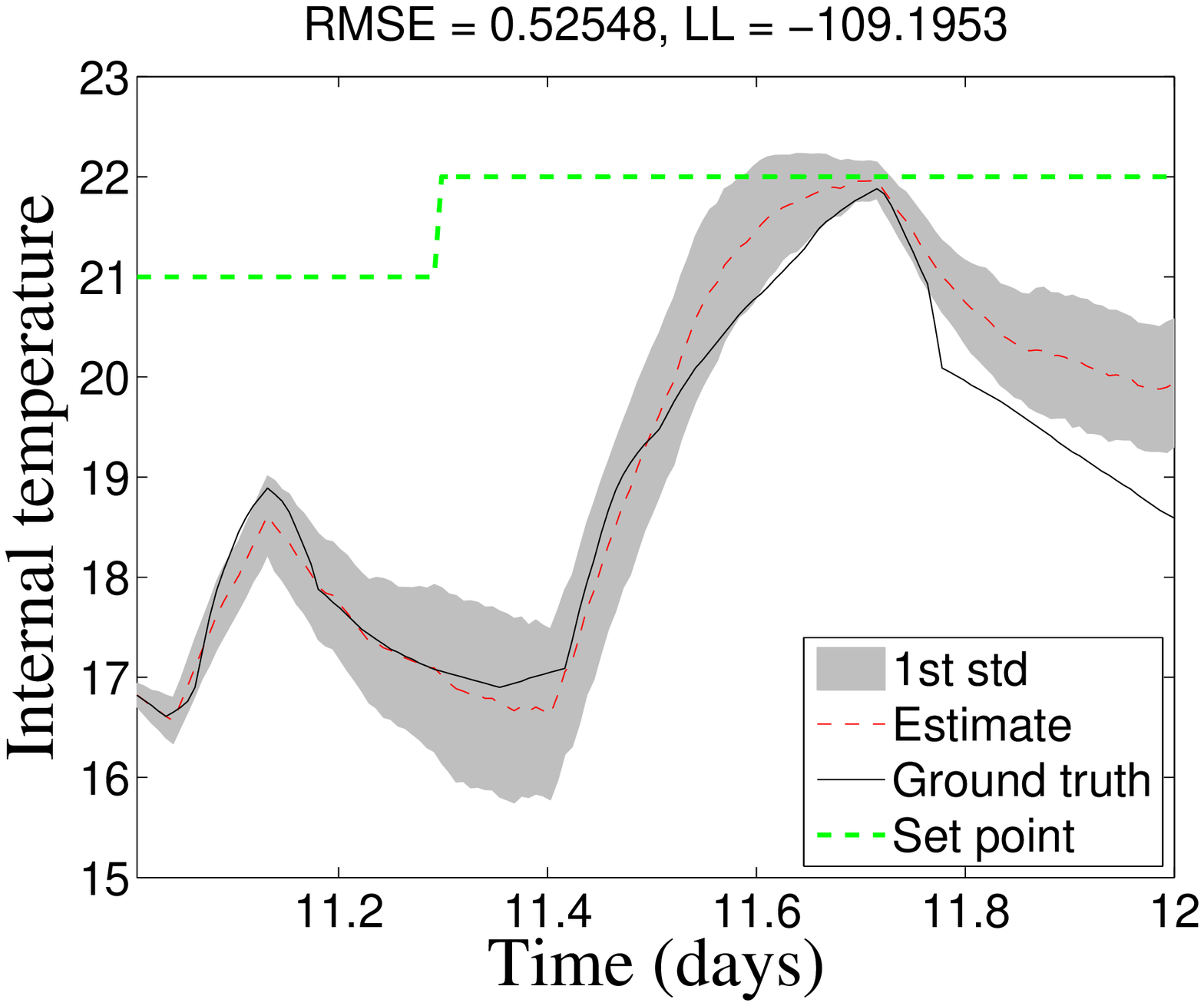} &
\includegraphics[width=0.35\textwidth]{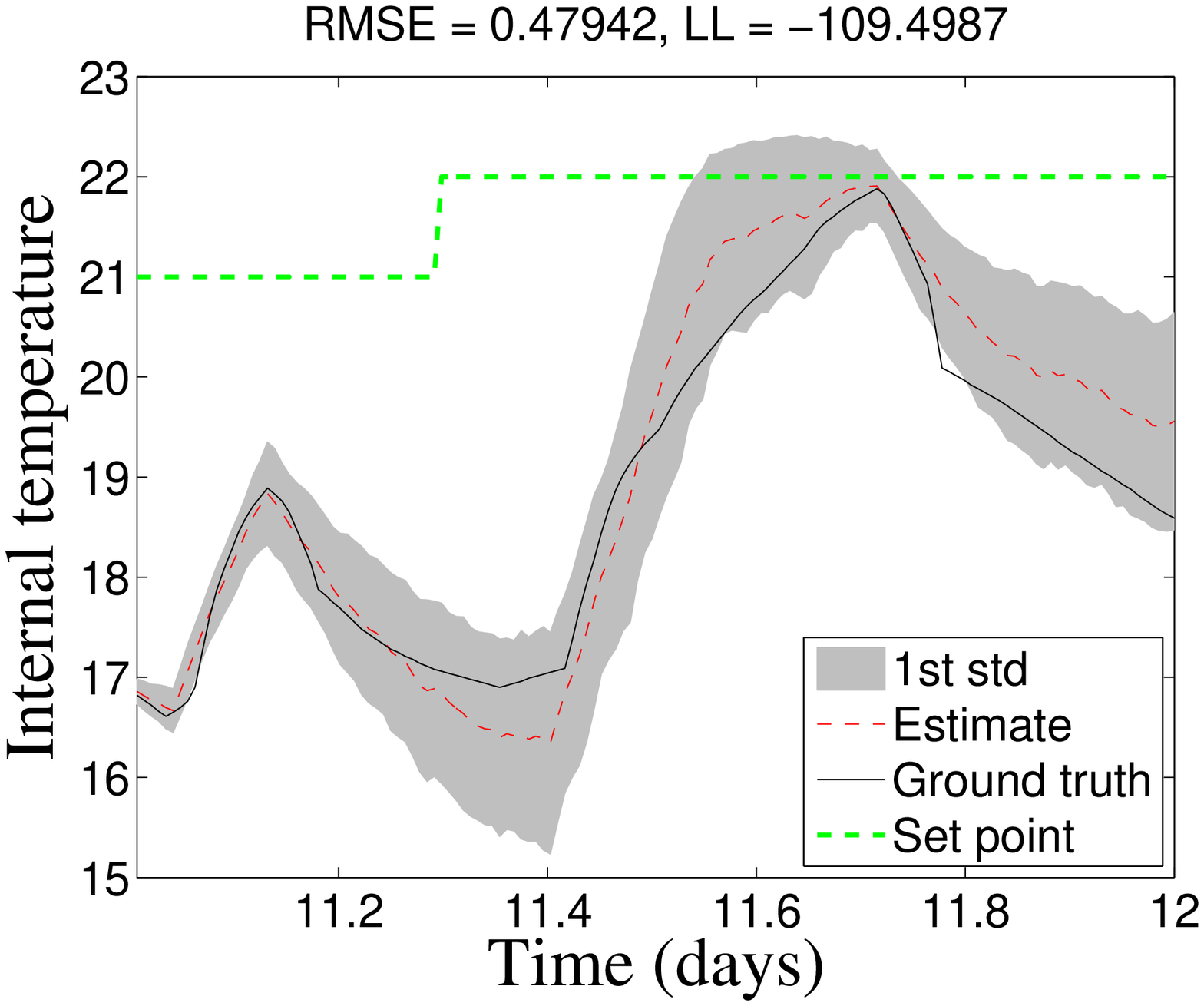}\\
\ \\ 
Hart$^+$ & Resonator$^+$\\
\includegraphics[width=0.35\textwidth]{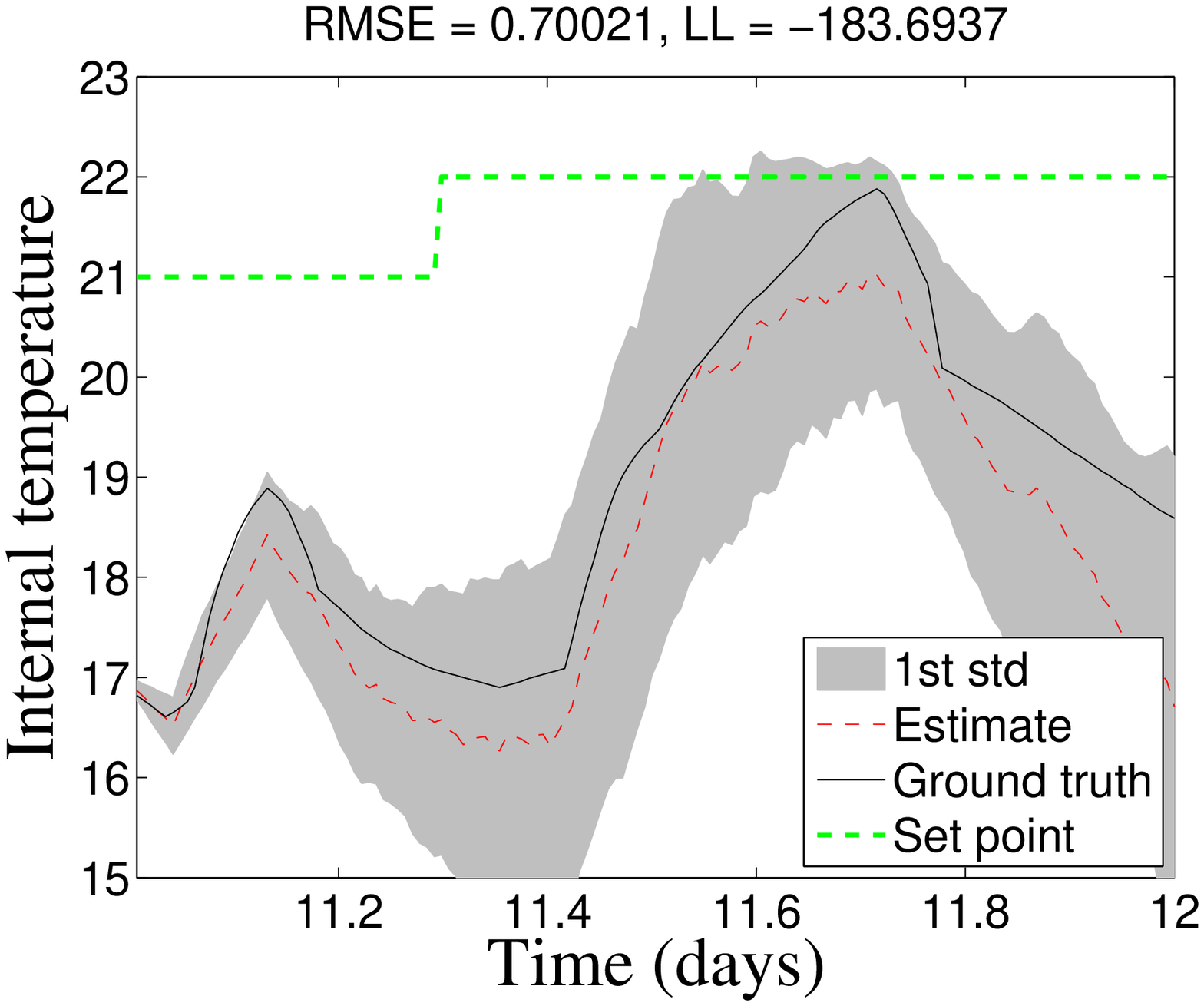} &
\includegraphics[width=0.35\textwidth]{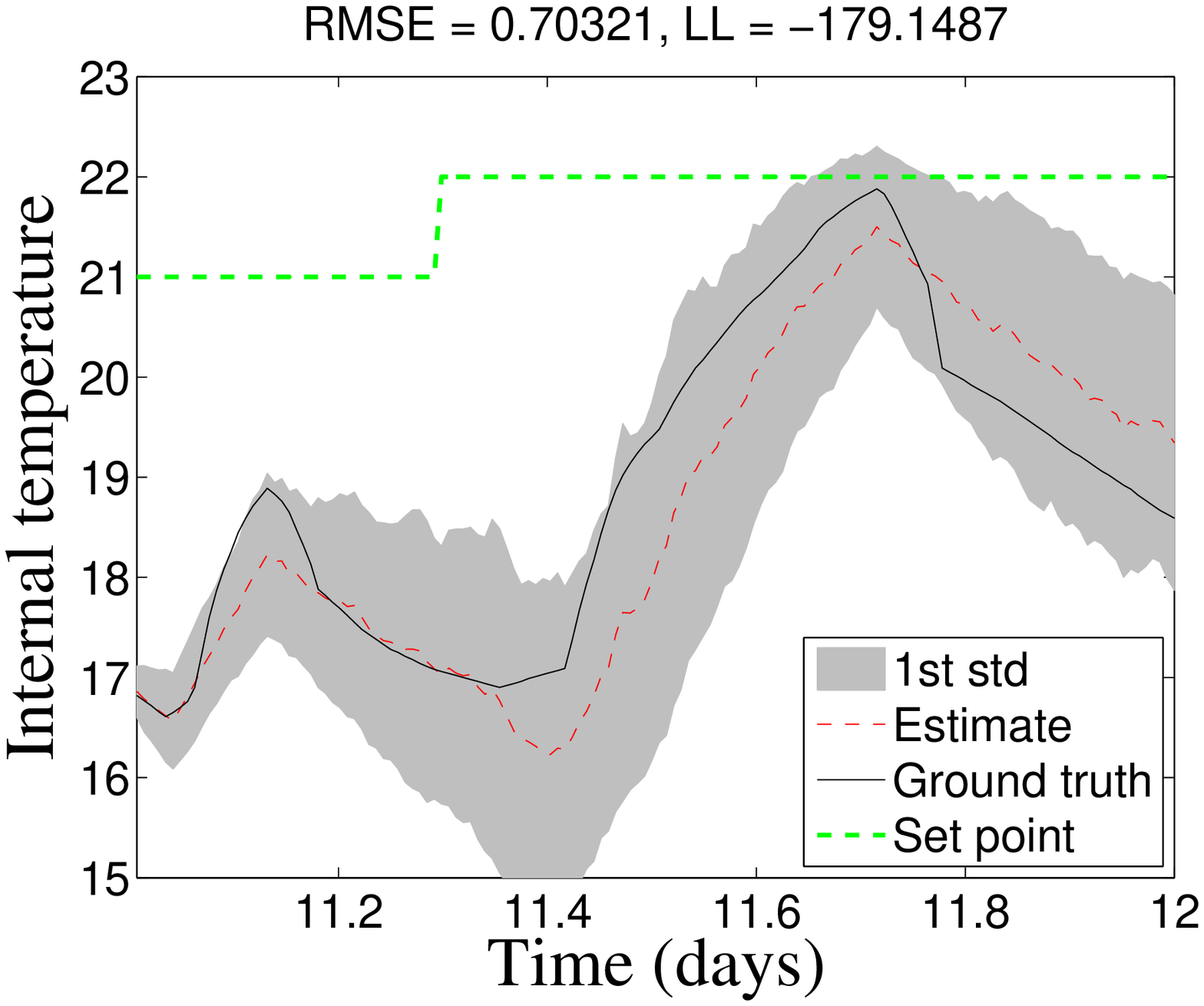}
\end{tabular}
\end{center}
\caption{\label{test} Internal temperature prediction compared to
  actual measured value using the \textbf{LFMwithout$^+$} (top left),
  the \textbf{LFMwith$^+$} (top right), the \textbf{LFMquasi
    (SQM)$^+$} (middle left), the \textbf{LFMquasi (CQM)$^+$} (middle
  right), the \textbf{Hart$^+$} (bottom left) and the
  \textbf{Resonator$^+$} (bottom right) algorithms.  The 1st standard
  deviation confidence interval is shown (grey).  Also shown is the
  thermostat set point (green).}
\end{figure}
 
We shall first illustrate the efficacy of the three algorithms on a
single example of the training and prediction process before
presenting a statistical comparison of the algorithms over the full
data set.  Figure~\ref{train} shows four days of training data from
{\tt data1}.  The central and bottom plots show the residual over the
two day period inferred by \textbf{LFMwith$^+$} and by \textbf{Hart$^+$}.  The
\textbf{Hart$^+$} plot shows that, although the residual exhibits some
daily periodicity, the cycle is imperfect.  However, the inferred
residual for \textbf{LFMwith$^+$} is more certain than that for
\textbf{Hart$^+$} as the periodic residual model in \textbf{LFMwith$^+$}
shares information between cycles. Consequently, the predictions drawn
using \textbf{LFMwith$^+$} are more accurate than those from \textbf{Hart$^+$}
as we will see subsequently.  In addition in Figure~\ref{train} in the
plot of the residual for \textbf{LFMwith$^+$}, we observe that this
residual tries to compensate for the errors in the daily fall in
temperature at the start of each day between times $7.0$ and $7.2$,
$8.0$ and $8.2$, $9.0$ and $9.2$ and again between $10.0$ and $10.2$.
We will show that these effects can have a significant impact on the
day ahead prediction of the internal temperature.  Although the
residual model for \textbf{LFMwith$^+$} is less certain than that for
\textbf{Hart$^+$}, it captures the residual errors that arise due to using
the simple thermal model in Equation~\eqref{thermalmodel}.  For
instance, at the start of each day, when the heating comes on, the
residual for \textbf{LFMwith$^+$} shows a sharp spike, which represents a
thermal lag in the physical process: in our homes a boiler heats up
water, which, as it flows through radiators, indirectly heats up the
air inside.  The residual for \textbf{Hart$^+$} however, is unable to
accurately model this lag.

Figure~\ref{test} shows the day-ahead prediction of the temperature
for the day immediately following the training days in
Figure~\ref{train} using all seven algorithms.  The \textbf{LFMquasi
  (CQM)$^+$} prediction of the internal temperature has the smallest
RMSE and one of the largest log likelihoods.  This indicates that the
underlying model is much more accurate than those of the other
approaches.  The RMSE for the example in Figure~\ref{test} is shown in
Table~\ref{RMSE}.
\begin{table}[ht]
  \caption{\label{RMSE} Internal temperature prediction RMSE 
    of real home data {\tt data1} over day 11 for
    non-periodic, quasi-periodic, periodic and no residual models.}
\begin{center}
\begin{tabular}{|c|c|}
\hline Method & RMSE\\\hline\hline
LFMquasi (SQM)$^+$ & 0.53\\ 
LFMquasi (CQM)$^+$ &{\bf 0.48}\\ 
LFMwith$^+$ & 0.49\\
LFMwithout$^+$ & 0.84\\
Hart$^+$ & 0.70\\
Resonator$^+$ & 0.70\\\hline
\end{tabular}
\end{center}
\end{table}

To fully test the accuracy of the inferred residual model we evaluated
the RMSE and the expected loglikelihood of the predicted temperature
for each day and for each algorithm over the $10$ days for both homes
for which predictions were generated.  The example in
Figures~\ref{train} and~\ref{test} corresponds to data set {\tt
  data1}, day $11$.  Table~\ref{thermal_tab} presents the expected
RMSE and the expected log likelihood of the predicted internal
temperatures for each home.  LFMs with periodic and quasi-periodic
eigenfunction residual models have both the best RMSE and expected log
likelihoods for {\tt data1} and {\tt data2}.  The best periodic model
overall with a mean RMSE of $0.52 \pm 0.05$ across both datasets and
an expected loglikelihood of $-106 \pm 13$ is the
\textbf{LFMwith$^+$}.  The \textbf{Resonator$^+$} model has a lower
consistency with an expected log likelihood of $-1373 \pm 1027$ and it
also has a higher overall RMSE at $1.15 \pm 0.22$.~\footnote{The
  relative performance of the eigenfunction and resonator models
  depends on how well the model parameters are learned.  Of course,
  changing the parameter inference mechanism could effect the
  performance measures reported in this paper.  Given this, we
  endeavoured to extract the best performance from each model.}  The
\textbf{LFMwithout$^+$} model is weak as it is unable to accurately
explain the training data without a residual model. Furthermore,
although the \textbf{Hart$^+$} approach has a very precise residual
model, as shown in Figure~\ref{train}, its predictions are weak since
it is unaware that the residual is periodic.  The non-periodic
Mat\'ern kernel, that is used by \textbf{Hart$^+$}, is unable to make
accurate long term predictions of the residual since the learned input
length scale for the residual is short.  We note that the
\textbf{LFMquasi (WQM)$^+$} performs relatively badly on these
datasets whereas the same algorithm performs well in the call centre
application in Section~\ref{sec:appqueue}.  The reason for this is
that the \textbf{LFMquasi (WQM)$^+$} best models quasi-periodicity
when the output scale of the residual changes between periods.  Since
the output scale for the heat residual does not vary from day to day
then this model gives a poor fit.  However, the scale of the queue
length varies significantly from day to day within the call centre
application and this is best modelled via the \textbf{LFMquasi
  (WQM)$^+$}.  We note that the \textbf{LFMwith$^+$}, \textbf{LFMquasi
  (WQM)$^+$}, \textbf{LFMquasi (SQM)$^+$} and \textbf{LFMquasi
  (CQM)$^+$} each used fewer than $24$ significant basis functions to
represent the residual process.  Consequently, our LBM Kalman filter
approach to LFM prediction is computationally efficient.
 \begin{table}[ht]
   \caption{\label{thermal_tab} Day ahead prediction (real home data):
     RMSE and expected log likelihood (ELL) for
     non-periodic, quasi-periodic, periodic and no residual models.}
\begin{center}
\hspace*{-1.2cm}
\begin{tabular}{|c|c|c|c|c|c|c|}
\hline \multirow{2}{*}{Method} & \multicolumn{2}{|c|}{data1} &
\multicolumn{2}{|c}{data2} & \multicolumn{2}{|c|}{Overall}\\ \cline{2-7}
 & RMSE & ELL & RMSE & ELL & RMSE & ELL\\ \hline\hline      
LFMquasi (SQM)$^+$ & $0.73 \pm 0.22$ & $-133 \pm  24$ & $0.59 \pm
0.07$ & ${\bf -90 \pm   8}$ & $0.67 \pm 0.13$ & $-116 \pm  15$\\
LFMquasi (CQM)$^+$ & $0.85 \pm 0.15$ & $-166 \pm  17$ & $0.89 \pm 0.19$ & $-144 \pm  20$ & $0.87 \pm 0.11$ & $-157 \pm  13$\\
LFMquasi (WQM)$^+$ & $1.15 \pm 0.14$ & $-260 \pm  17$ & $0.94 \pm 0.17$ & $-179 \pm  27$ & $1.06 \pm 0.11$ & $-227 \pm  18$\\
LFMwith$^+$ & ${\bf 0.51 \pm 0.06}$ & ${\bf -104 \pm  19}$ & ${\bf
  0.55 \pm 0.09}$ & $-108 \pm  19$ & ${\bf 0.52 \pm 0.05}$ & ${\bf -106 \pm  13}$\\
LFMwithout$^+$ & $0.59 \pm 0.08$ & $-156 \pm  40$ & $0.65 \pm 0.11$ & $-121 \pm  21$ & $0.61 \pm 0.06$ & $-142 \pm  25$\\
Hart$^+$ & $1.03 \pm 0.17$ & $-183 \pm  21$ & $0.75 \pm 0.17$ & $-130 \pm  20$ & $0.92 \pm 0.12$ & $-162 \pm  16$\\
Resonator$^+$ & $1.39 \pm 0.32$ & $-2122 \pm 1702$ & $0.79 \pm 0.22$ &
$-250 \pm 110$ & $1.15 \pm 0.22$ & $-1373 \pm 1027$\\
\hline
\end{tabular}
\end{center}
\end{table}
     
We also evaluated the algorithms on the data when tracking the
internal temperature over a day.  We note, when tracking, the heater
output is known at each time instant and, thus, it is not necessary to
use the RBPF whose sole purpose is to accommodate uncertainty in the
binary heater output.  Thus, each LFM is now implemented through a
standard Kalman filter.  The LFM models were trained over four
consecutive days as described above, but, in this case, the day ahead
internal temperatures were filtered using measurements obtained every
$100$ minutes.  Table~\ref{thermal_tab_track_real} presents the
expected RMSE and the expected log likelihood of the internal
temperatures for each home.
\begin{table}[ht]
  \caption{\label{thermal_tab_track_real} Tracking a day ahead: RMSE and expected
    log likelihood (ELL) for non-periodic, quasi-periodic, periodic and no residual models.}
\begin{center}
\hspace*{-1.2cm}
\begin{tabular}{|c|c|c|c|c|c|c|}
\hline \multirow{2}{*}{Method} & \multicolumn{2}{|c|}{data1} &
\multicolumn{2}{|c}{data2} & \multicolumn{2}{|c|}{Overall}\\ \cline{2-7}
 & RMSE & ELL & RMSE & ELL & RMSE & ELL\\ \hline\hline     
LFMquasi (SQM) & $0.19 \pm 0.01$ & $ 85 \pm  13$ & $0.28 \pm 0.04$ & $
12 \pm  25$ & ${\bf 0.22 \pm 0.02}$ & ${\bf 56 \pm  15}$\\
LFMquasi (CQM) & $0.19 \pm 0.02$ & $ 63 \pm  12$ & ${\bf 0.27 \pm
  0.04}$ & $ -9 \pm  26$ & ${\bf 0.22 \pm 0.02}$ & $ 34 \pm  15$\\
LFMquasi (WQM) & $0.26 \pm 0.06$ & $ 51 \pm  30$ & $0.29 \pm 0.06$ &
$-23 \pm  44$ & $0.27 \pm 0.04$ & $ 21 \pm  26$\\
LFMwith & ${\bf 0.18 \pm 0.02}$ & ${\bf 87 \pm  11}$ & $0.32 \pm 0.04$ & $-41 \pm  29$ & $0.24 \pm 0.02$ & $ 35 \pm  21$\\
LFMwithout & $0.22 \pm 0.03$ & $ 48 \pm  16$ & $0.29 \pm 0.04$ & $  6 \pm  25$ & $0.25 \pm 0.02$ & $ 31 \pm  14$\\
Hart & $0.21 \pm 0.02$ & $ 78 \pm  15$ & ${\bf 0.27 \pm 0.05}$ & ${\bf 26 \pm  24}$ & $0.23 \pm 0.02$ & $ 55 \pm  14$\\
Resonator & $0.82 \pm 0.34$ & $-190 \pm  87$ & $0.81 \pm 0.35$ & $-343 \pm 225$ & $0.82 \pm 0.24$ & $-251 \pm 101$\\
\hline
\end{tabular}
\end{center}
\end{table}

To determine the efficacy of the algorithms under more pronounced
residual forces we simulated the heater output and, consequently, the
internal temperature for residual heat drawn from a crisp quasi-periodic
Mat\'ern Gaussian process.  We drew the residual process from the
step-quasi model (SQM) as this model was a good representation of the
real data as demonstrated in Table~\ref{thermal_tab}.  We then inferred
the internal temperature process using Equation~\eqref{thermalmodel}.
Although we found that all three quasi-periodic models exhibited similar
RMSE performance for day ahead tracking, the SQM model showed
significant performance improvement over all other models when
predicting a day ahead.

\begin{figure}[ht]
\begin{center}
\begin{tabular}{ccc}
LFMwithout & LFMwith\\
\includegraphics[width=0.35\textwidth]{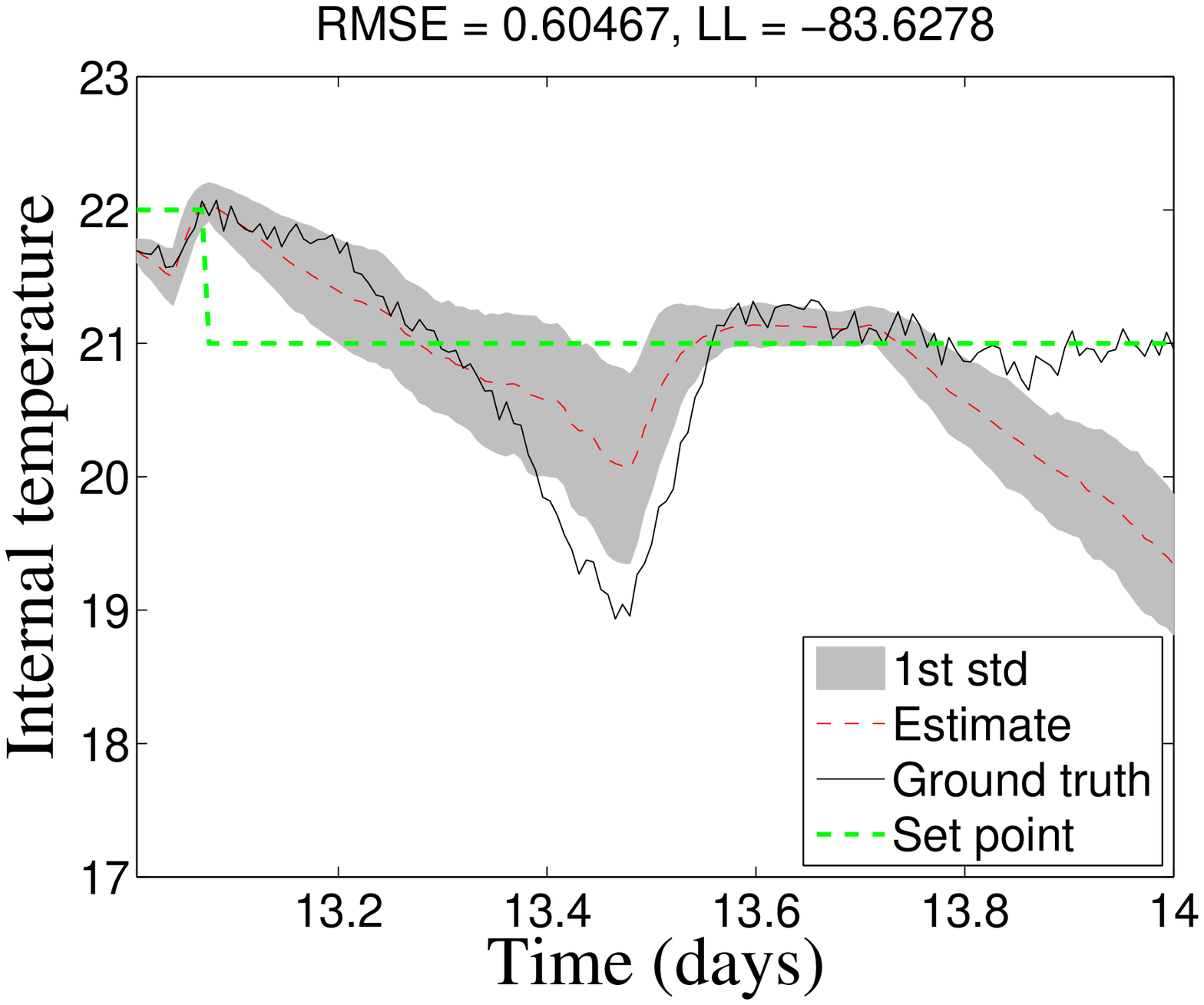} &
\includegraphics[width=0.35\textwidth]{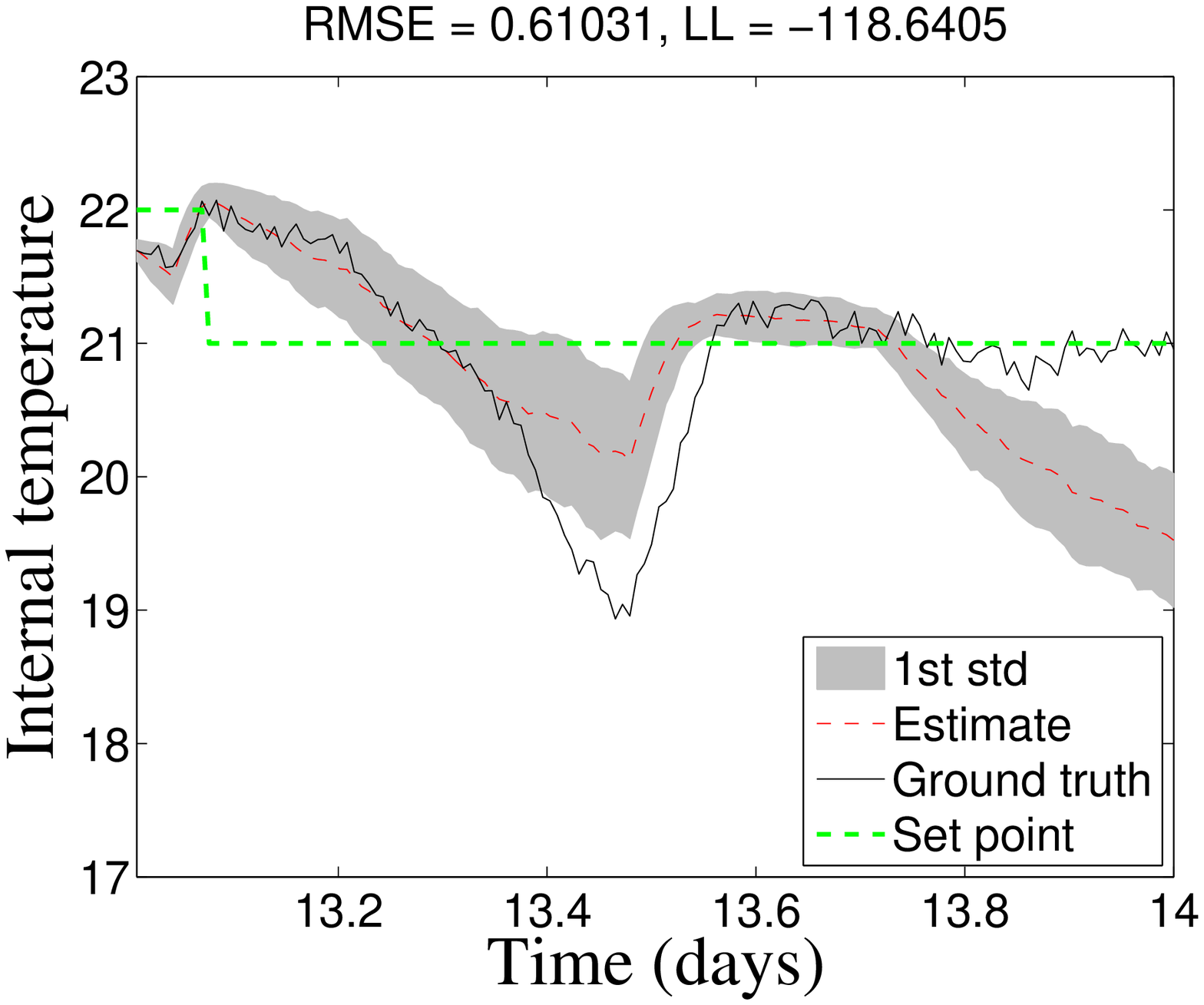}\\
\  \\
LFMquasi (SQM) & LFMquasi (CQM)\\
\includegraphics[width=0.35\textwidth]{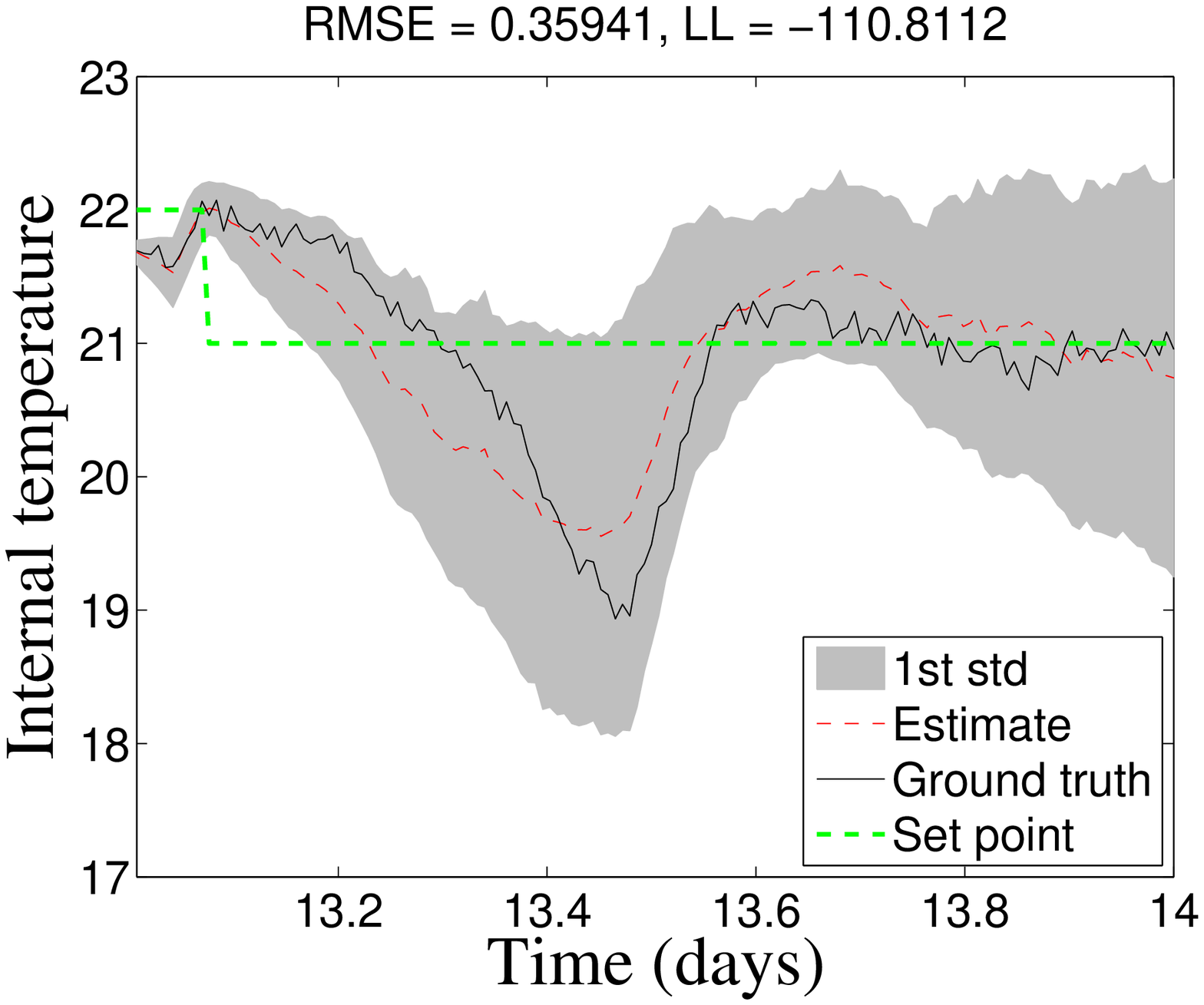} &
\includegraphics[width=0.35\textwidth]{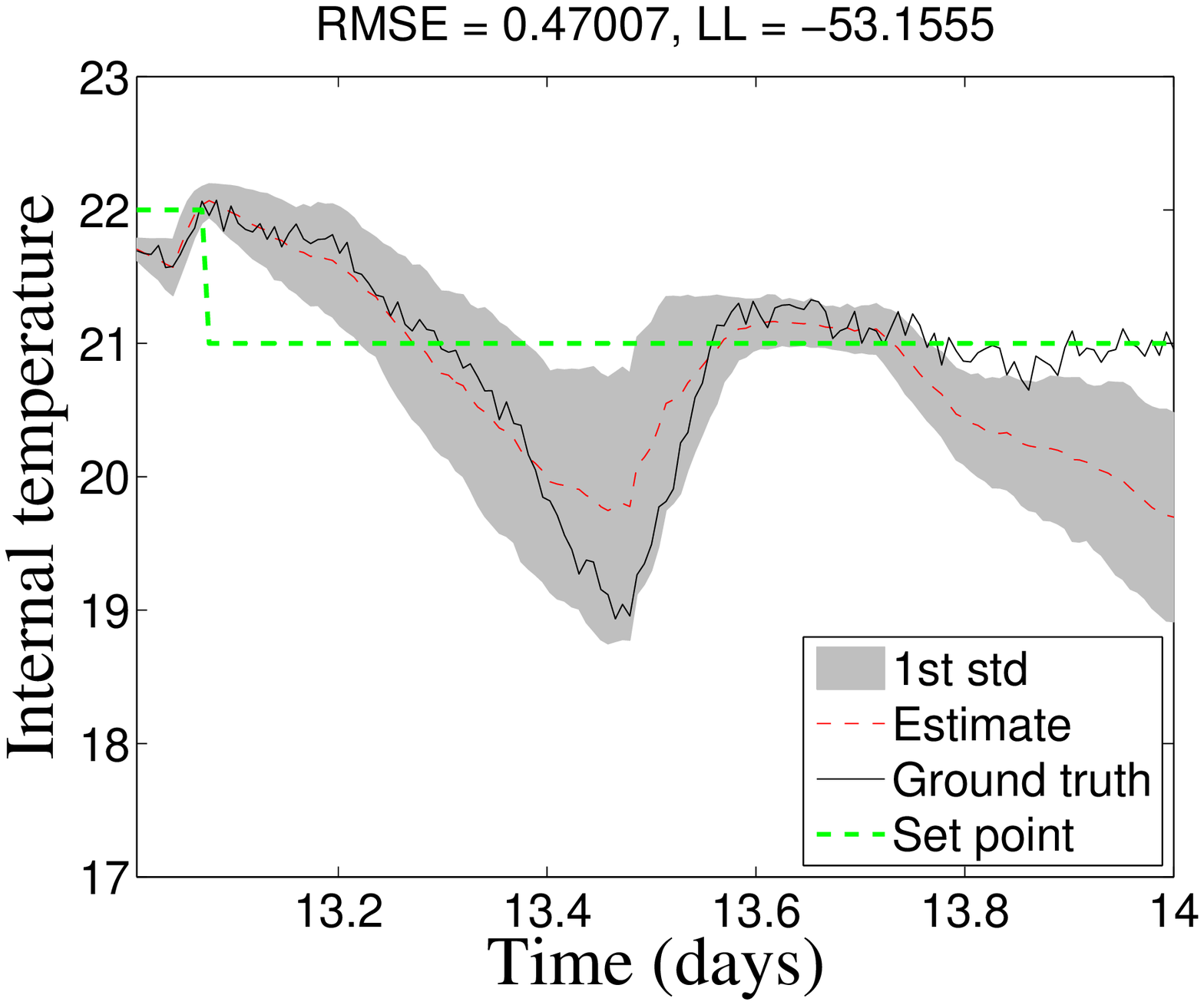}\\
\ \\
Hart & Resonator\\
\includegraphics[width=0.35\textwidth]{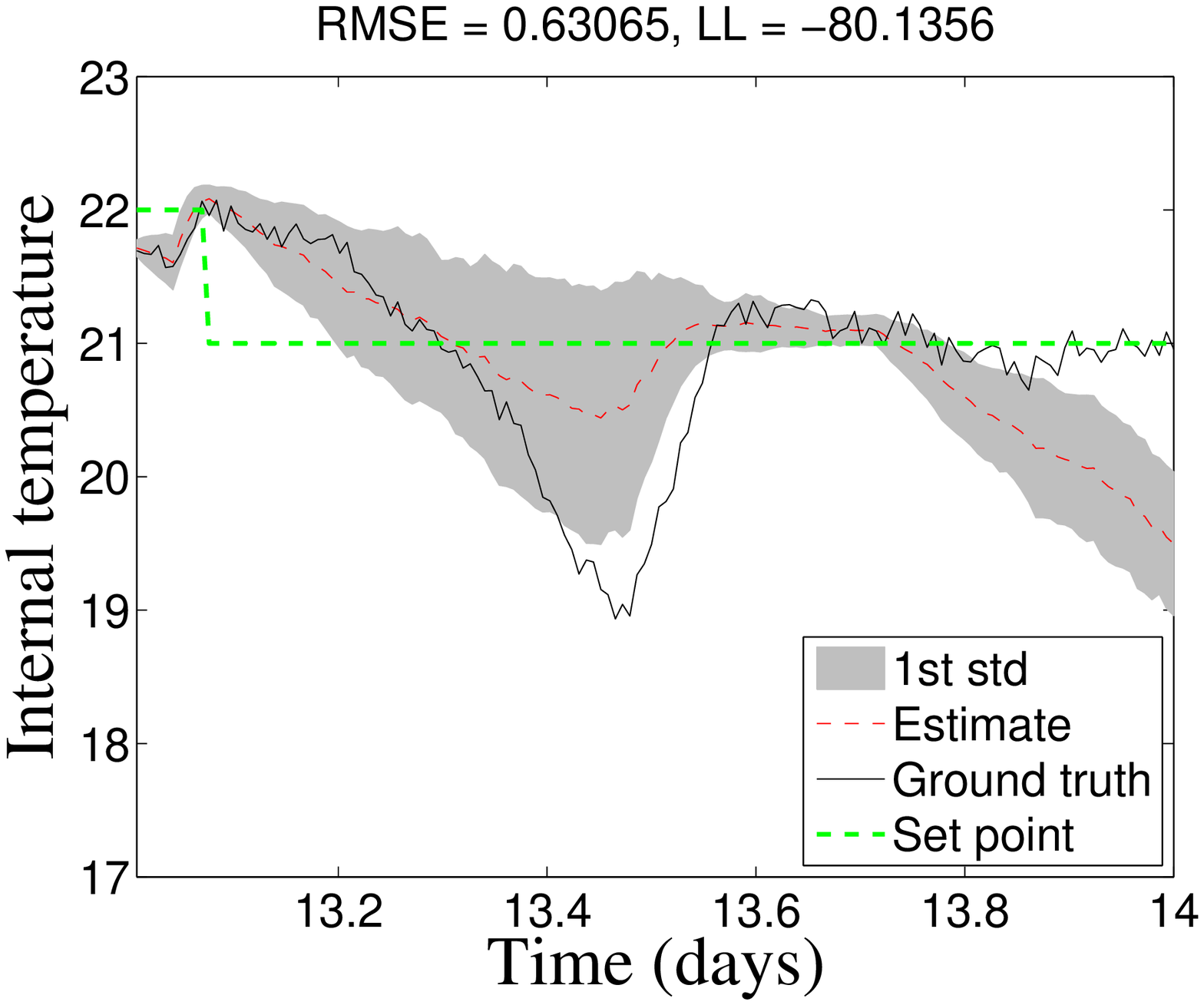} &
\includegraphics[width=0.35\textwidth]{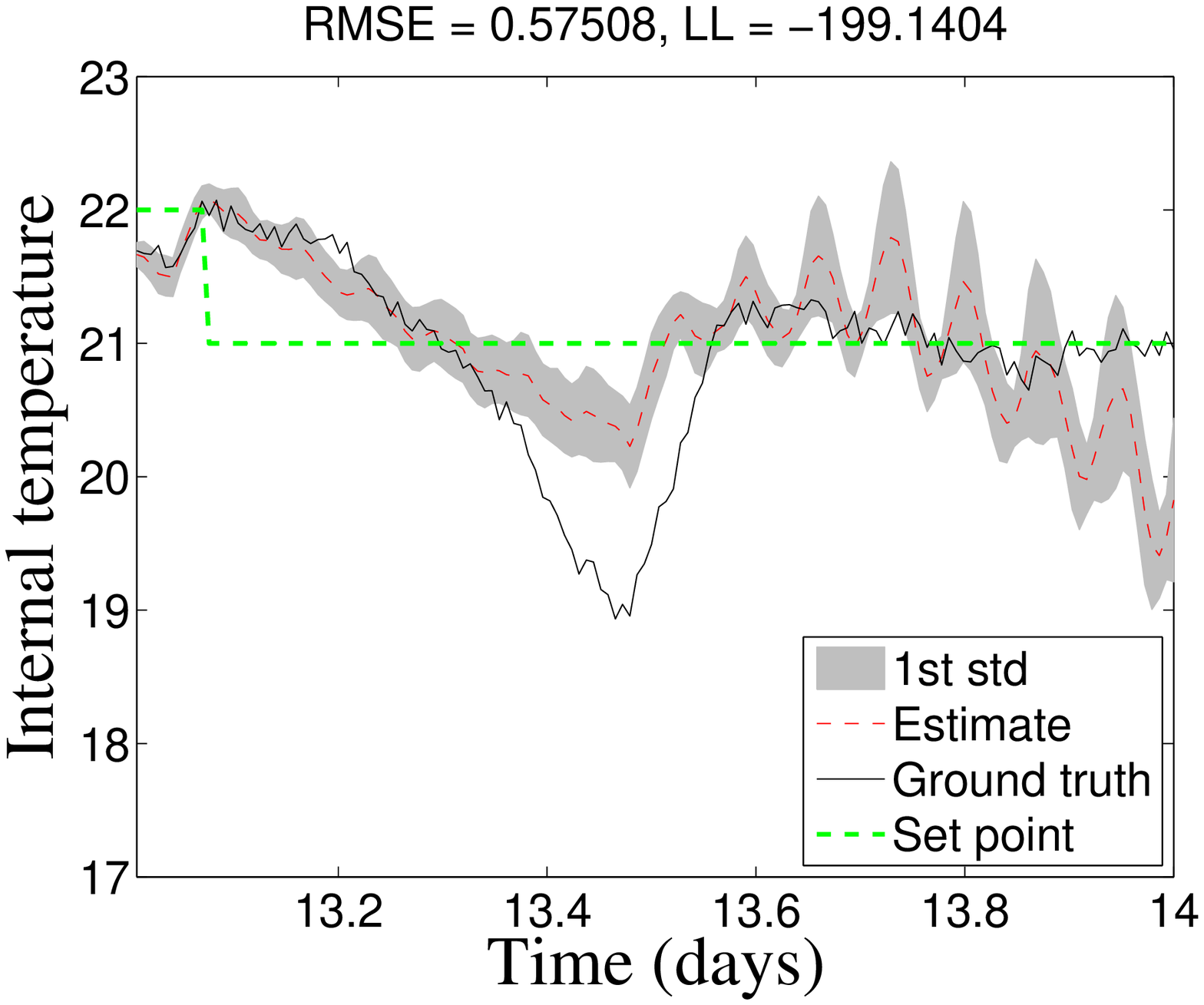}
\end{tabular}
\end{center}
\caption{\label{extreme_test_pred} Predicted internal temperature
  compared to simulated value using the \textbf{LFMwithout} (top
  left), the \textbf{LFMwith} (top right), the \textbf{LFMquasi
    (SQM)} (middle left), the \textbf{LFMquasi (CQM)} (middle
  right), the \textbf{Hart} (bottom left) and the
  \textbf{Resonator} (bottom right) algorithms.  The 1st standard
  deviation confidence interval is shown (grey).  Also shown is the
  thermostat set point (green).}
\end{figure}
Example estimates for each prediction algorithm are shown in
Figure~\ref{extreme_test_pred}.  We re-evaluated the filter algorithms
on this simulated data and the results are presented in
Table~\ref{thermal_tab_simulated}.  The \textbf{ LFMquasi (SQM)$^+$}
exhibits the lowest RMSE and highest loglikelihood overall with values
$1.00\pm 0.16$ and $-190\pm 28$, respectively, which isn't surprising
as the alternative approaches all use incorrect models for the
residual.  However, specifically the \textbf{LFMquasi(SQM)$^+$} is
significantly more accurate and consistent than the
\textbf{Resonator$^+$} model, which has an RMSE and expected
likelihood of $1.43\pm 0.19$ and $-279 \pm 54$, respectively.
Consequently, despite the flexibility of the resonator model, it is
unable to capture the dynamics of the SQM generated residual as it has
not been informed of the prior nature of the residual and further, is
unable to recover this information from the data. Clearly, encoding
the appropriate prior model for the residual is critical for tracking
the internal temperature accurately.
\begin{table}[ht]
  \caption{\label{thermal_tab_simulated} Day ahead prediction
    (partially simulated home data): RMSE and expected log
    likelihood (ELL) for simulated
    non-periodic, quasi-periodic, periodic, Resonator and no residual models.}
\begin{center}
\hspace*{-1.2cm}
\begin{tabular}{|c|c|c|c|c|c|c|}
\hline \multirow{2}{*}{Method} & \multicolumn{2}{|c|}{data1} &
\multicolumn{2}{|c}{data2} & \multicolumn{2}{|c|}{Overall}\\ \cline{2-7}
 & RMSE & ELL & RMSE & ELL & RMSE & ELL\\ \hline\hline     
LFMquasi (SQM)$^+$ & ${\bf 0.75 \pm 0.24}$ & ${\bf -161 \pm  30}$ &
${\bf 1.12 \pm 0.20}$ & ${\bf -204 \pm  40}$ & ${\bf 1.00 \pm 0.16}$ &
${\bf -190 \pm  28}$\\
LFMquasi (CQM)$^+$ & $1.37 \pm 0.26$ & $-288 \pm  59$ & $1.37 \pm 0.23$ & $-258 \pm  48$ & $1.37 \pm 0.17$ & $-268 \pm  37$\\
LFMquasi (WQM)$^+$ & $1.43 \pm 0.40$ & $-236 \pm  39$ & $1.46 \pm 0.27$ & $-241 \pm  27$ & $1.45 \pm 0.21$ & $-239 \pm  21$\\
LFMwith$^+$ & $1.02 \pm 0.34$ & $-338 \pm 198$ & $1.23 \pm 0.25$ & $-377 \pm 116$ & $1.16 \pm 0.19$ & $-364 \pm  97$\\
LFMwithout$^+$ & $1.57 \pm 0.38$ & $-376 \pm 128$ & $1.70 \pm 0.36$ & $-362 \pm 101$ & $1.66 \pm 0.26$ & $-367 \pm  77$\\
Hart$^+$ & $1.48 \pm 0.37$ & $-247 \pm  36$ & $1.40 \pm 0.22$ & $-294 \pm  80$ & $1.43 \pm 0.19$ & $-279 \pm  54$\\
Resonator$^+$ & $1.48 \pm 0.37$ & $-247 \pm  36$ & $1.40 \pm 0.22$ & $-294 \pm  80$ & $1.43 \pm 0.19$ & $-279 \pm  54$\\
\hline
\end{tabular}
\end{center}
\end{table}
  
\begin{figure}[ht]
\begin{center}
\begin{tabular}{ccc} 
\includegraphics[width=0.45\textwidth]{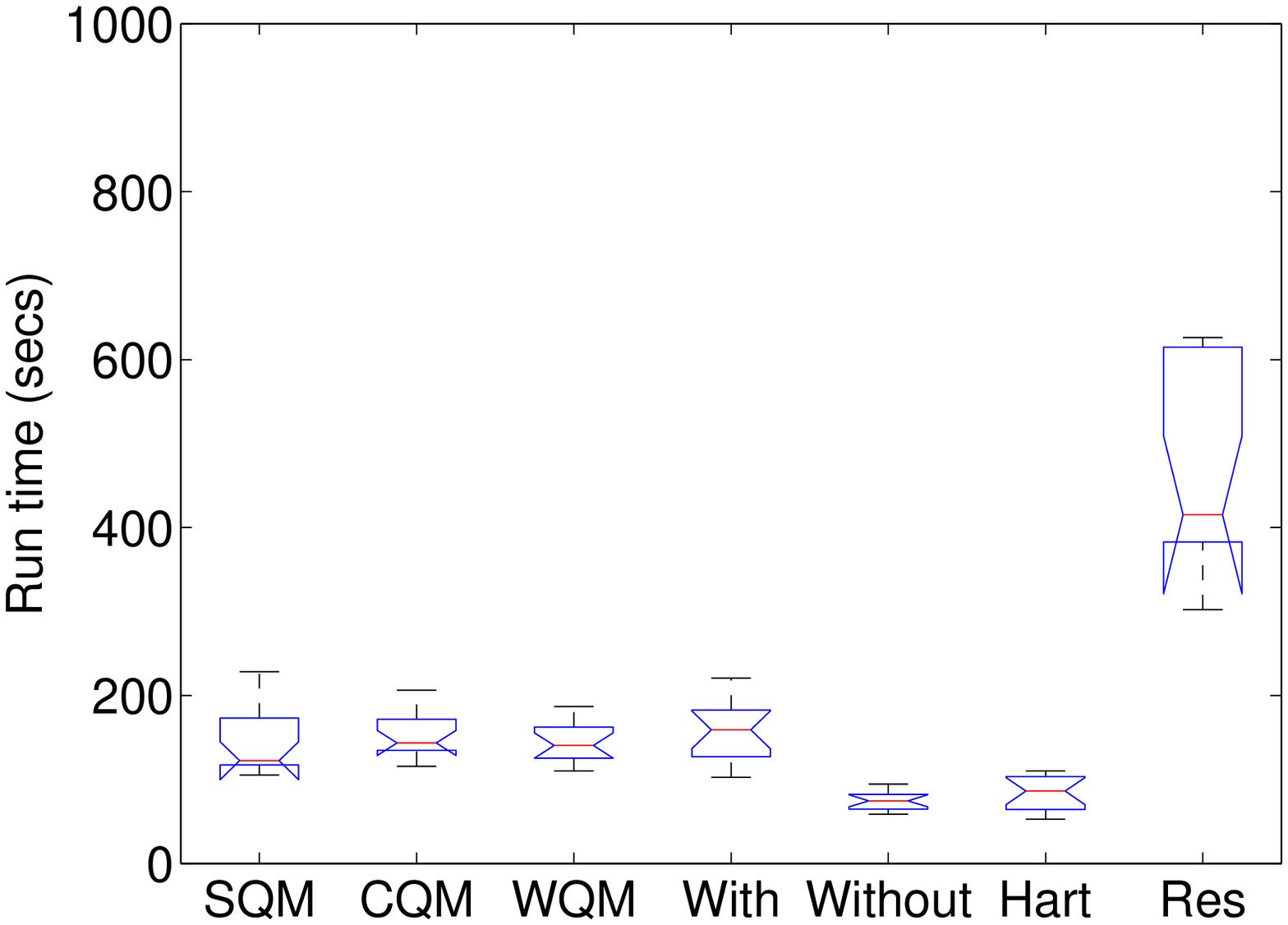}
& &
\includegraphics[width=0.45\textwidth]{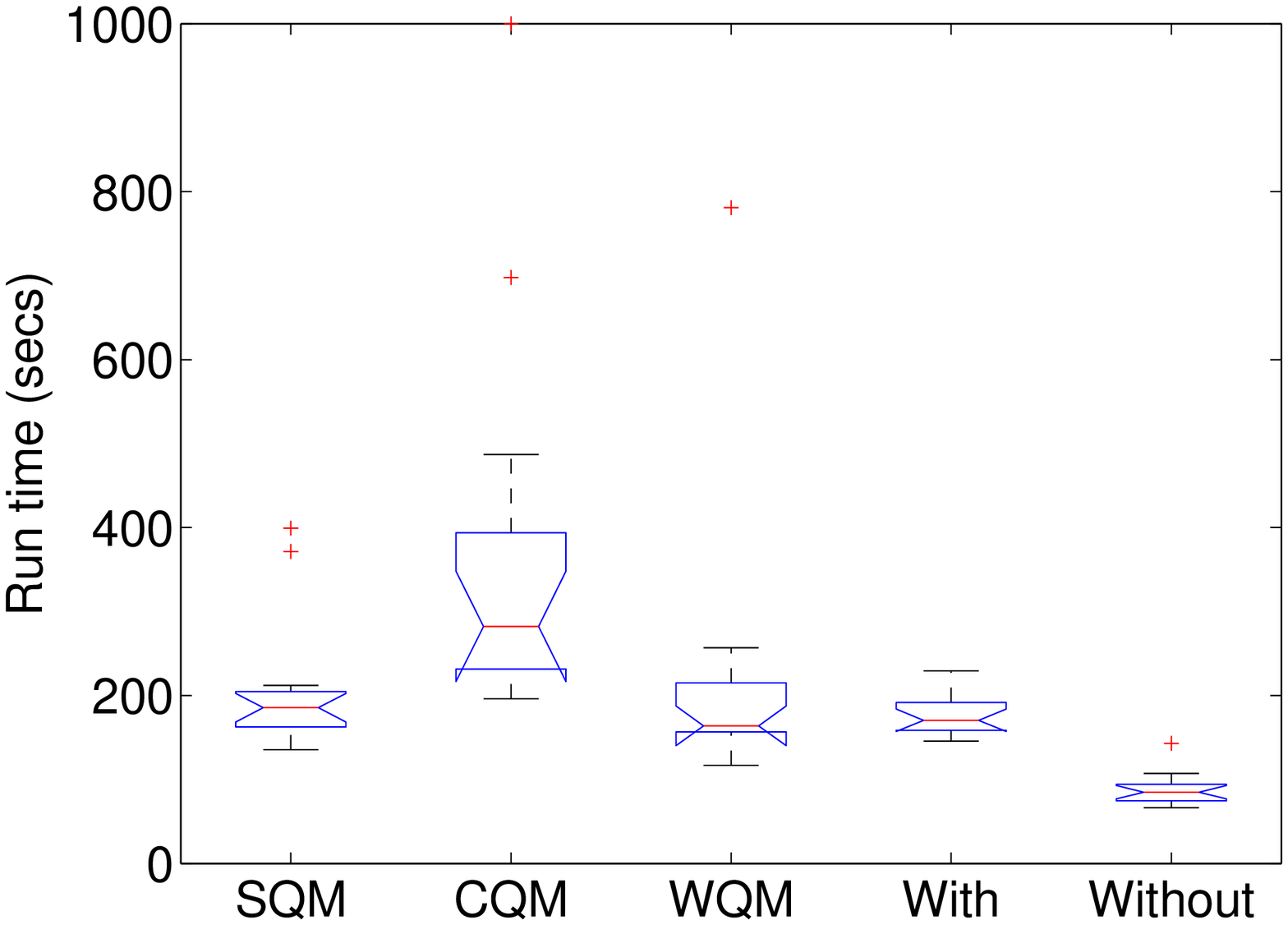}\\
(a) Single output thermal model & & (b) Multi-output thermal model
\end{tabular}
\end{center}
\caption{\label{run_times} Empirical distribution of run times (in seconds) for each LFM algorithm for both the single
  output and multi-output latent force models.  The time for a sample
  run includes the time to train the model, predict a day ahead
  and also track a day ahead.}
\end{figure}

We also compared the run times for each algorithm.~\footnote{The run
  times were determined using a Macbook Pro with a 2.4 GHz Intel Core
  i7 processor and 8 GB of memory.}  We collected the time it took to
train each model on four days of data, predict an entire day ahead and
then track the internal temperature over that day.  For each run the
resonator model and \textbf{LFMquasi (SQM)$^+$} used exactly the same
number of resonators and eigenfunctions, respectively.  The resonator
model used between $19$ and $21$ resonators during the
experiment. Further, the resonator model was provided with a bias term
to accomodate non-zero mean residuals.  Figure~\ref{run_times}(a)
shows a box plot of the single output algorithm run times.  The
resonator algorithm is clearly the slowest as the model inference for
the resonator requires a search over a space of frequency and decay
coefficients.  A detailed breakdown and comparison of the
computational costs of the eigenfunction model and resonator model is
presented in Section~\ref{sec:compcompl}.

Finally, we demonstrate the efficacy of our approach at modelling a
multi-output system and consider an extension to the thermal model
that incorporates the effect of a building's {\em envelope} as
proposed in \cite{madsenmodel}.  The building's envelope comprises
mainly the walls which act as a thermal reservoir and delay the heat
transfer between the inside and the outside of the building.  The
multi-output model is represented by a system of coupled differential
equations,
\begin{eqnarray} \label{eq:titemodel}
\frac{dT_{int}(t)}{dt}&=& \alpha \left(T_{env}(t) - T_{int}(t)\right)
+ \beta E(t) + R(t)\ ,\\
\frac{dT_{env}(t)}{dt}&=& \Gamma \left(T_{int}(t) - T_{env}(t)\right)
+ \Psi \left(T_{ext}(t) - T_{env}(t)\right)\ .
\end{eqnarray}
Here, $T_{int}$ and $T_{env}$ are the internal temperature within the
home and the temperature of a building's envelope,
respectively. $T_{env}$ is not directly observed, and has to be
inferred from the data. The parameters in this model include: i)
$\beta$, which represents the thermal output of the heater, ii)
$\alpha$, which regulates the convective heat transfer from the
internal ambient air to the envelope, iii) $\Gamma$, which weights the
convective heat transfer from the envelope to the ambient air and iv)
$\Psi$, which represents the leakage coefficient to the ambient
environment. In this model $T_{ext}(t)$ and $E(t)$ are the latent
forces.

\begin{figure}[ht]
\begin{center}
\begin{tabular}{ccc}
LFMwithout$^+$ & LFMwith$^+$\\
\includegraphics[width=0.35\textwidth]{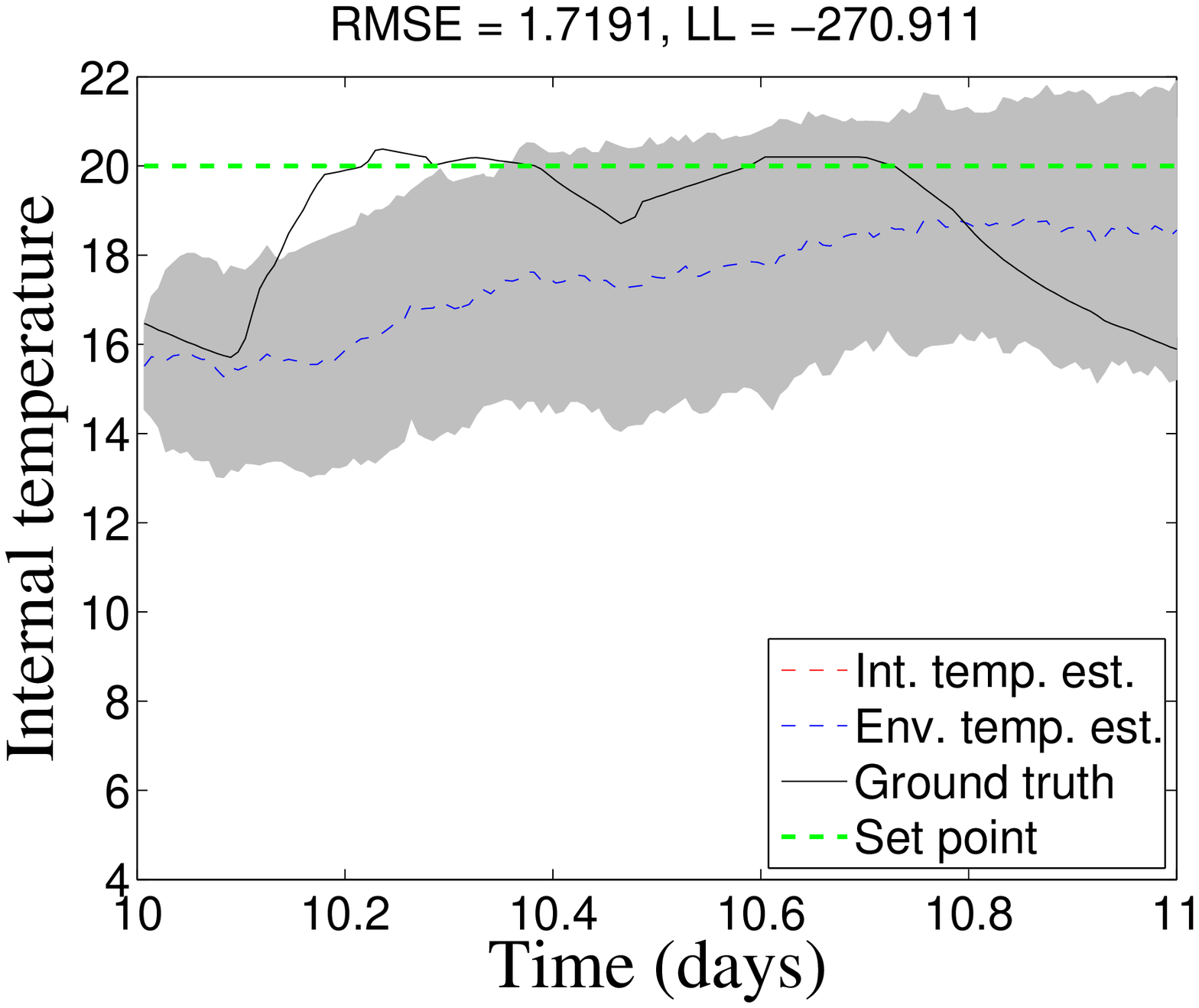} &
\includegraphics[width=0.35\textwidth]{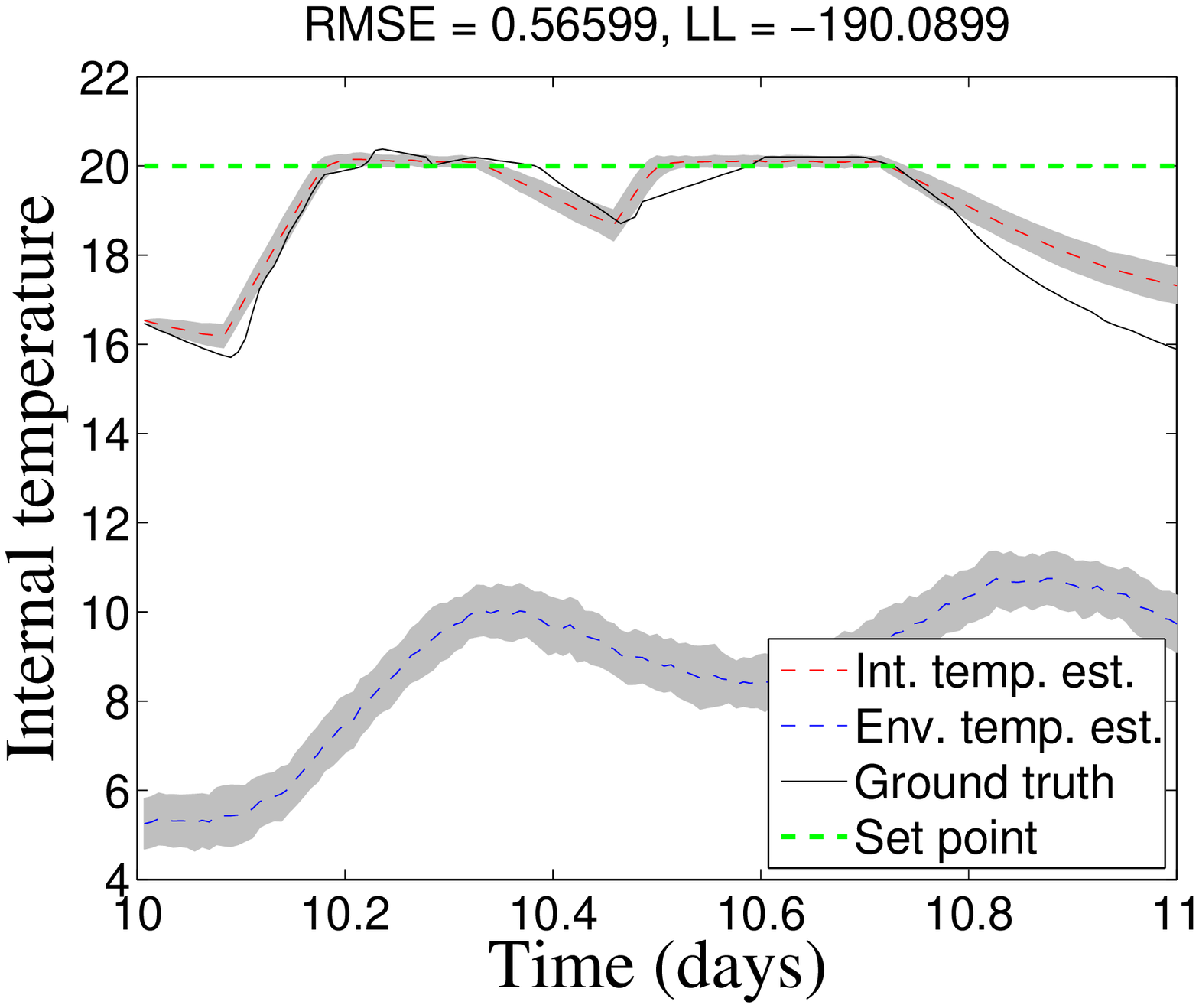}\\
\  \\
LFMquasi (SQM)$^+$ & LFMquasi (CQM)$^+$\\
\includegraphics[width=0.35\textwidth]{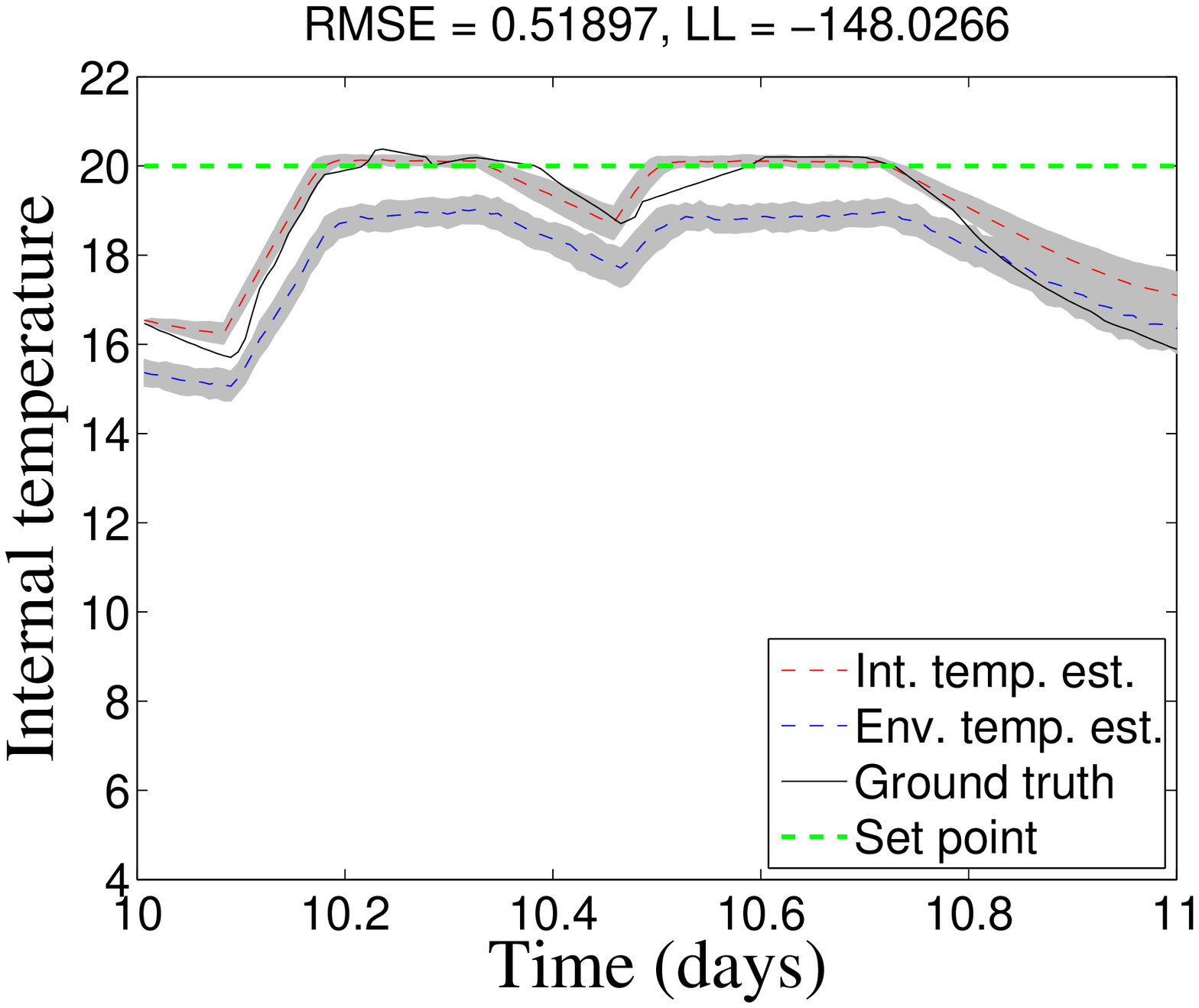} &
\includegraphics[width=0.35\textwidth]{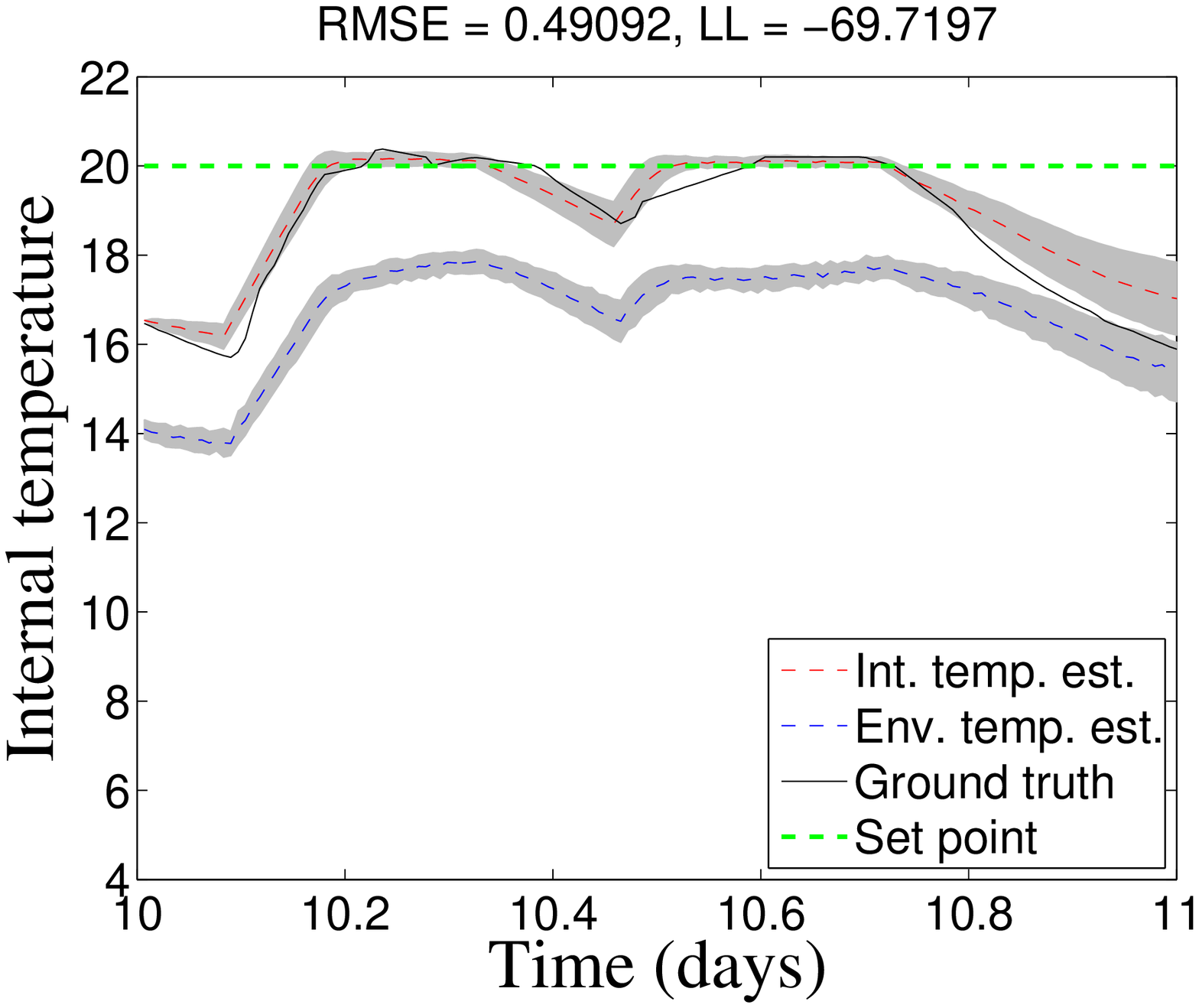}
\end{tabular}
\end{center}
\caption{\label{test_env} Internal temperature predictions compared to
  actual real value using the \textbf{LFMwithout$^+$} (top left), the
  \textbf{LFMwith$^+$} (top right), the \textbf{LFMquasi (SQM)$^+$}
  (bottom left), the \textbf{LFMquasi (CQM)$^+$} (bottom right)
  algorithms.  The 1st standard deviation confidence interval is shown
  (grey).  Also shown is the thermostat set point (green).}
\end{figure}

To infer the internal temperature of the building using the envelope
model we introduce $T_{env}$ to the Kalman filter state vector and two
further parameters, $\Gamma$ and $\Psi$, whose most likely values are
inferred from the training data.  We repeated the experiments on the
real data above but this time using the envelope model.
Figure~\ref{test_env} shows example day-ahead predictions of the
internal and envelope temperatures for day $10$ in dataset {\tt
  data1}.

Table~\ref{thermal_tab_multi} presents the expected RMSE and the
expected log likelihood of the predicted internal temperatures for
each home. The best algorithm is the \textbf{LFMquasi (SQM)$^+$} which
uses a quasi-periodic residual model.  Referring to the performance of
the single-output model in Table~\ref{thermal_tab} it is interesting
to note that the addition of the envelope, as proposed in
\cite{madsenmodel}, improves the overall performance of all the
algorithms.
\begin{table}[ht]
  \caption{\label{thermal_tab_multi} Predicting internal temperature a
    day ahead using the multi-output model: RMSE and expected
    log likelihood (ELL) for quasi-periodic and periodic models
    on real thermal data from two homes.}
\begin{center}
\hspace*{-1.2cm}
\begin{tabular}{|c|c|c|c|c|c|c|}
\hline \multirow{2}{*}{Method} & \multicolumn{2}{|c|}{data1} &
\multicolumn{2}{|c}{data2} & \multicolumn{2}{|c|}{Overall}\\ \cline{2-7}
 & RMSE & ELL & RMSE & ELL & RMSE & ELL\\ \hline\hline    
LFMquasi (SQM)$^+$ & ${\bf 0.19 \pm 0.02}$ & ${\bf 80 \pm  12}$ &
${\bf 0.27 \pm 0.04}$ & $  7 \pm  19$ & ${\bf 0.22 \pm 0.02}$ & ${\bf 51 \pm  14}$\\
LFMquasi (CQM)$^+$ & $0.29 \pm 0.07$ & $ 12 \pm  30$ & $0.28 \pm 0.04$ & $ -8 \pm  18$ & $0.29 \pm 0.05$ & $  4 \pm  19$\\
LFMquasi (WQM)$^+$ & $0.20 \pm 0.02$ & $ 48 \pm  27$ & ${\bf 0.27 \pm 0.05}$ & $ -1 \pm  32$ & $0.23 \pm 0.02$ & $ 29 \pm  21$\\
LFMwith$^+$ & $0.21 \pm 0.02$ & $ 56 \pm  19$ & $0.32 \pm 0.03$ & $-26 \pm  23$ & $0.25 \pm 0.02$ & $ 23 \pm  18$\\
LFMwithout$^+$ & $0.27 \pm 0.05$ & $ 24 \pm  26$ & $0.28 \pm 0.05$ & $
{\bf 10 \pm  23}$ & $0.27 \pm 0.04$ & $ 19 \pm  18$\\
\hline
\end{tabular}
\end{center}
\end{table}
However, of key importance for the application of our approach to
multi-output latent force models in general is the RBPF run times for
this model and how they compare with the single output case.
Figure~\ref{run_times}(b) shows a box plot of the run times (the total
time to train the RBPF, predict a day ahead and also track a day
ahead) for the multi-output case.  The run times compare favourably
with the run times for the single output case despite the fact that
the multi-output model requires two extra parameters.  The increased
computational cost for the multi-output model is due to the extra
parameters in the multi-output model and this cost would be present if
the standard Gaussian process inference equations, as
Equations~\eqref{GP1} and~\eqref{GP2}, were used in place of the
Kalman filter.  In general, the computational complexity of the Kalman
filter scales quadratically with the size of the state vector and so
multiple output processes can be accommodated efficiently.  We note
that our approach has a linear cost when conditionally independent
measurements of multiple processes are incorporated.  This contrasts
with the standard Gaussian process inference equations which have a
cubic cost in the number of processes and measurements from each
process due to the need to invert a covariance matrix over all the
processes.

In both the home heating application and the call centre application
the eigenfunction-based models demonstrated the best performance, with
improved RMSE and expected loglikelihood over non-residual models,
non-periodic models and the resonator model.  Further, the
quasi-periodic residual models were shown to outperform perfectly
periodic models on problems for which regular human behaviours, such
as queuing as customers or heating homes through cooking or switching
on the heating, have some influence.  We noted that the WQM model had
the best performance on the call centre application but the SQM
exhibited the best predictive performance on the thermal modelling application.
The WQM performed well on the call centre application because that
application included residual forces, in this case arrival rates,
which varied in amplitude from day to day.  The SQM succeeded in the
thermal modelling application because the residual heat profile varied
slightly from day to day whilst maintaining a constant overall
amplitude.

\section{CONCLUSIONS}
\label{sec:conclusions}

We have derived a novel and principled Bayesian approach to latent
force modelling which accommodates both periodic and non-periodic
forces. This approach can be incorporated within computationally
efficient, iterative state-space approaches to inference. We are the
first to demonstrate that eigenfunctions can be used to model periodic
forces within state-space approaches to LFM inference and we offer the
only principled approach to incorporating periodic covariance
functions within a state-space approach to inference with LFMs. We use
the approach in \citet{hartikainen10} for modelling non-periodic
kernels and eigenfunction basis functions for modelling periodic
kernels within a state-space approach to inference.  We demonstrated
that our eigenfunction approach out-performs the sparse spectrum
Gaussian process regression (SSGPR) approach developed by
\citet{gredilla11}.  Further, we demonstrated the close link between
the eigenfunction model and the resonator model proposed by
\citet{sarkka12}.  Consequently, we are the first to demonstrate how
any periodic covariance function can be encoded within the resonator
model using the covariance function's eigenfunctions.  We are also the
first to demonstrate that eigenfunctions can be represented via the
resonator model within Kalman filters if required.  Thus, we have
proposed, in this paper, the only two approaches to date that are able to
incorporate all types of Gaussian periodic model priors within a
state-space approach to LFM inference.  These priors include
stationary periodic, non-stationary periodic and quasi-periodic
covariance functions.

We have applied our approach to two applications: call centre customer
queues and thermal modelling of homes. In detail, within the call
centre application, customer arrival rates were modelled as driving
forces through a differential model approximation of the Poisson
arrival process.  Both periodic and quasi-periodic models were
developed to model the arrival rates of customers.  The periodic
models improve on the non-periodic model by as much as $83\%$ in the
root-mean-squared error.  In the home heating application we modelled
the thermal dynamics of homes where the physics of the energy exchange
process is known but some of the heat generating processes are not
known in advance.  Our approach can learn the unknown heat dynamics
from data and is able to accurately predict internal temperatures $24$
hours ahead.  Again, both periodic and quasi-periodic models were
developed but, in this case, to model residual heat within the home.
In this case the periodic models improve on the non-periodic model by
reducing the RMSE by as much as $28\%$. Overall, the quasi-periodic
models produced the lowest mean-squared-error and the highest expected
log likelihood. Further, the eigenfunction model demonstrated improved
performance over the resonator model.  In the thermal application the
eigenfunction models improve on the resonator by reducing the RMSE by
as much as $74\%$.

In both the thermal modelling application and the call centre
application the periodic residual models demonstrated the best
performance, with improved RMSE and expected loglikelihood over
non-residual models, non-periodic models and the resonator model.
Further, the quasi-periodic residual models were shown to outperform
perfectly periodic models in the presence of regular human behaviours,
such as customer queues and heating homes through cooking or switching
on the heating. We noted that the WQM model had the best performance
for the call centre application but the SQM exhibited the best performance
on the thermal application.  The WQM performed well on the call centre
application because that application included residual forces, in this
case arrival rates, which varied in amplitude from day to day.  The
SQM succeeded in the thermal application because the residual heat
profile varied slightly from day to day whilst maintaining a constant
overall amplitude.

Both applications deployed state-space approaches to LFMs and both
applications utilised the eigenfunction representation of the periodic
latent forces acting on the system.  These applications demonstrated
the efficacy of our approach on both long term predictions and
tracking problems.  The applications demonstrated LFM inference on
both linear (home heating) and non-linear (call centre)
problems; on latent forces with constant output scale (home
heating) and variable output scale (call centre) and on purely
Gaussian models (call centre) and models involving both Gaussian and
binomial variables (thermal).  We also demonstrated both single
output Gaussian process and multi-output Gaussian process regression
in the home heating application.

As we noted in the Appendix, the eigenfunction is the optimal RMSE
basis model for any covariance function.  However, our approach uses
only the eigenfunctions derived from the covariance function prior.
Consequently, an optimal J-dimensional model should adapt its basis
functions to the data set and the eigenfunctions of the posterior
covariance function should be used.  We believe it is possible to
extend our approach to accommodate adaptable eigenfunctions and this
will be the focus of further work. Further, the Kalman formalism
expressed in this paper lends itself immediately to control problems
and we intend to investigate our approach to LFM inference within
model-based predictive control.  This research will be of particular
value to domains in which some physical knowledge of the process is
known (and expressible via differential equations) and nonparametric
models can be used to express the latent forces.  We will explore the
relative merits of expressing control problems directly via the
Gaussian process prior as in, for example, \citet{azman08}, and via the
Markovian formalism advocated in this paper.

\section*{ACKNOWLEDGEMENTS}
This work was funded in the UK by the EPSRC ORCHID programme grant
(EP/I011587/1) and the EPSRC `Intelligent Agents for Home Energy
Management' project (EP/I000143/1), and in Kingdom of Saudi Arabia by
the Deanship of Scientific Research (DSR), King Abdulaziz University,
Jeddah (9-15-1432-HiCi).

\bibliography{references}

\appendix
\section{DISCRETE JUMP MARKOV PROCESSES FOR NON-STATIONARY COVARIANCE FUNCTIONS}
\label{app:discretejump}
We describe how the Step Quasi model (SQM) and the Wiener-step Quasi
model (WQM) can be incorporated within the discrete time Kalman
filter.  

Suppose that either $a\sim\mathcal{GP}(0,K^{\text{SQM}})$ or
$a\sim\mathcal{GP}(0,K^{\text{WQM}})$, where $K^{\text{SQM}}$ and
$K^{\text{WQM}}$ are the covariance functions for the SQM and WQM,
respectively. Consider a changepoint, $\tau$ and some earlier time
$\tau_-$ close to $\tau$ such that $\tau>\tau_-$.  We assume that,
\begin{eqnarray} 
a(\tau)=G a(\tau_-)+\chi(\tau)
\label{aaa}
\end{eqnarray} 
where $G$ is the Kalman filter process model and
$\chi(\tau)\sim\mathcal{N}(0,Q)$.  We will now see that $G$ and $Q$
can be expressed in terms of the kernels, $K^\text{SQM}$ and
$K^\text{WQM}$ at the changepoint $\tau$.  Recall
$E[a(t)]=E[\chi(t)]=0$, $E[a(t)\chi(t)]=0$ and
$K(t,t^\prime)=E[a(t)a(t^\prime)]$ for all $t$ and $t^\prime$.  Thus,
by squaring both sides of Equation~\eqref{aaa} and then taking the
expectation we get the variance, $K(\tau,\tau)$, of $a(\tau)$,
\begin{eqnarray} 
K(\tau,\tau)=G K(\tau_-,\tau_-) G+Q\ .
\label{quasiP}
\end{eqnarray} 
Also, multiplying Equation~\eqref{aaa} throughout by
$a(\tau_-)$ before taking the expectation gives the covariance between
$a(\tau)$ and $a(\tau_-)$,
\begin{eqnarray}
K(\tau,\tau_-)=G K(\tau_-,\tau_-)\ .
\label{quasiC}
\end{eqnarray}
Specifically, from Equation~\eqref{sqmdef}, the SQM variance,
$K^\text{SQM}(\tau,\tau)=\sigma^2$ and
$K^\text{SQM}(\tau,\tau_-)=\sigma^2\exp(-1/l)$ as
$C(\tau)-C(\tau_-)=1$ in Equation~\eqref{sqmdef} across a
single changepoint.  Thus, using Equations~\eqref{quasiP}
and~\eqref{quasiC}, the process model, $G$,
and process noise variance, $Q$, for the SQM are,
\begin{eqnarray*}
G_\text{SQM}=\exp\left(-\frac{1}{l}\right)\ ,\hspace*{2cm} Q_\text{SQM}=\sigma^2\left(1-\exp\left(-\frac{2}{l}\right)\right)\ .
\end{eqnarray*}
Also, by Equation~\eqref{wqmdef} the WQM variances,
$K^\text{WQM}(\tau,\tau)=\xi_0+C(\tau)\xi$,
  $K^\text{WQM}(\tau_-,\tau_-)=\xi_0+C(\tau_-)\xi$ and covariance,
    $K^\text{WQM}(\tau,\tau_-)=\xi_0+C(\tau_-)\xi$.  Thus, by
    Equations~\eqref{quasiP} and~\eqref{quasiC}, the process model,
    $G$, and process noise variance, $Q$, for the WQM are,
\begin{eqnarray*}
G_\text{WQM}=1\ ,\hspace*{2cm} Q_\text{WQM}=\xi
\end{eqnarray*}
as $C(\tau)-C(\tau_-)=1$. 

\section{COMPARISON OF EIGENFUNCTION AND RESONATOR MODELS}
\label{app:eigres}
In this section we assert that the eigenfunction basis model advocated
in this paper is optimal in that it minimises the mean squared error
for all possible J-dimensional basis models and thus establish the
eigenfunction approach as the preferred approach.  We shall then
develop the theoretical link between the eigenfunction basis model and
the {\em resonator model} \citep{sarkka12,hartikainen12,solin13} which
is the most significant alternative approach to modelling periodic
forces in LFMs.  Consequently, we will demonstrate that the resonator
model parameters can be chosen so that the resonator basis is
equivalent to the eigenfunction basis.  As a corollary we propose a novel
mechanism for encoding periodic covariance function priors in the
resonator model.

\subsection{Establishing the Link Between the Resonator Basis and
  Eigenfunctions}
\label{sec:link}
\label{sec:optimal}
A {\em J-dimensional linear model} is a linear combination of $J$ basis
functions.  Both the eigenfunction model, as per
Equation~\eqref{sparseexact}, and resonator model, as per
Equation~\eqref{lbmres}, are J-dimensional linear models.  The
eigenfunction model, as per Equation~\eqref{sparseexact}, is a linear combination of
orthonormal basis functions, $\phi_j$, whereas the resonator model is
a linear combination of resonators, $\psi_j$, which are
not necessarily orthogonal.

Let $g$ be some function drawn from a Gaussian process with covariance
function $K$. Then the Karhunen-Lo\'eve expansion theorem
\citep{loeve55} states that the eigenfunction basis is the orthonormal
basis that minimises the total mean squared error between the
J-dimensional model and
$g$. Further, any non-orthonormal basis with cardinality,
$\alpha$, can be converted to an orthonormal basis with
cardinality, $\upsilon$, such that $\upsilon\le \alpha$, by
Gram-Schmidt orthogonalisation \citep{arfken05} and renormalisation.
Thus, we can establish immediately that the eigenfunction basis is the
optimal mean squared basis for all J-dimensional linear
models.~\footnote{In this paper, we use a static basis chosen from the
  prior covariance function.  However, the eigenfunctions are
  dependent on the covariance function and consequently, an optimal
  J-dimensional model should adapt its basis when evidence is
  integrated with the prior. We believe it is possible to extend our
  approach to accommodate adaptable eigenfunctions and this will be
  the focus of a further paper. We note that the resonator model can
  also adapt to the evidence provided the frequency process in
  Equation~\eqref{resonator1} adapts with the data.}  

In the remainder of this section we determine the conditions under
which each version of the resonator model, as per
Equations~\eqref{resonator1} and~\eqref{resonator2}, is equivalent to
the optimal eigenfunction model.

\subsubsection{Perfectly Periodic and Stationary Covariance Functions}
A {\em perfectly periodic stationary process} $g\sim\mathcal{GP}(b,K)$
with period $D$ satisfies, $g(t+nD)=g(t)$ for all $t\in \mathbb{R}$
and $n\in \mathcal{N}$.  Such functions (for example, the
squared-exponential in Equation~\eqref{periodicsquaredexp}) are
generated from Gaussian processes with covariance functions of the
form $K(t,t^\prime)=h(t-t^\prime)$ for some function $h$.

Bochner's theorem \citep[see, for example,][]{rasmussen06} states that 
the eigenfunctions of a stationary kernel are the Fourier basis functions. 
Thus, the optimal J-dimensional linear model for a stationary Gaussian
process is a linear combination of Fourier basis functions.  Both resonator
models, in Equations~\eqref{resonator1} and~\eqref{resonator2}, can
model Fourier basis functions exactly by asserting $\omega_j(t)=0$ for all time
$t$ and all resonators, $j$, in Equation~\eqref{resonator1}, and
assigning a constant resonator frequency, $f$, in the original resonator
model, as per Equation~\eqref{resonator1}, or removing the decay term
by setting $B_j=0$ in the later model, as per
Equation~\eqref{resonator2}, and assigning $A_j=-(2 \pi f_j)^2$,
\begin{eqnarray*}
\frac{d^2 \psi_j(t)}{d t^2}=-(2 \pi f_j)^2 \psi_j(t)\ .
\end{eqnarray*}  
Thus, the optimal J-dimensional linear model for the stationary kernel
case is an instance of both resonator models.  \commentout{The resonator model
requires two entries in the KF state-vector for each basis as the
differential model for each basis is second order.  However, each
resonator represents two Fourier basis functions, a cosine and sine for
each frequency as a single sine wave with a phase shift.  Our approach
requires only one state variable per eigenfunction.  However, each
eigenfunction is a Fourier basis, a cosine or a sine. Thus two
eigenfunctions are equivalent to each resonator and Kalman filtering
with both approaches is equally computationally complex.}

\subsubsection{Perfectly Periodic and Non-stationary Covariance
  Functions}
\label{sec:perfectnonstat}  
A {\em perfectly periodic non-stationary process}
$g\sim\mathcal{GP}(b,K)$ with period $D$ satisfies, $g(t+nD)=g(t)$ for
all $t\in \mathbb{R}$ and $n\in \mathcal{N}$.  Such processes (for example,
Equation~\eqref{nonstatex})
are Gaussian processes with covariance functions of the
form $K(t,t^\prime)= h(t,t^\prime)$ where
$h(t,t^\prime)\not=h(t-t^\prime)$. Note that since the latent force
$g$ is perfectly periodic then the resonator cannot be
stochastic (that is, $\omega_j(t)=0$ for all time $t$ and resonator, $j$,
in Equation~\eqref{resonator1}).

We demonstrate that the eigenfunctions for non-stationary covariance
functions can be represented by the resonator model using the time
varying frequency model, as per Equation~\eqref{resonator2}, provided
that the eigenfunction is second order differentiable.  We note that
the eigenfunction linear basis model, as per
Equation~\eqref{sparse}, and the resonator model, as per
Equation~\eqref{lbmres}, are equivalent if,
\begin{eqnarray}
\psi_j(t)=a_j \phi_j(t)\ ,
\label{equivcond}
\end{eqnarray}
for eigenfunction, $\phi_j$,
resonator, $\psi_j$, times, $t$, and some positive coefficient, $a_j$,
as per Equation~\eqref{lbmres}. Substituting
Equation~\eqref{equivcond} into Equation~\eqref{resonator1}, asserting
$\omega_j(t)=0$ (as above) and rearranging,
\begin{eqnarray} 
(2\pi f_j(t))^2=-\frac{1}{\phi_j(t)}\frac{d^2 \phi_j(t)}{dt^2}\ .
\label{resfreq}
\end{eqnarray} 
Thus, any perfectly periodic covariance function can be encoded within
the resonator model by defining the frequency process, $f_j(t)$, in
terms of the covariance function eigenfunctions, $\phi_j(t)$.
Furthermore, we can also represent eigenfunctions via the resonator
model within Kalman filters if required. In practise, the Nystr\"om
approximation, $\tilde{\phi}$, for the eigenfunction basis is used in
place of $\phi$ in Equation~\eqref{resfreq} to calculate the frequency
process for the resonator model.~\footnote{We note by
  Equation~\eqref{resfreq} the resonator can become unstable close to
  $\tilde{\phi}=0$.  This problem is easily solved by initially adding
  some offset, $\Delta$, to $\tilde{\phi}$ for some suitably large
  $\Delta$ before calculating the frequency process $f(t)$.
  Consequently, when $f(t)$ is used in the resonator model, as per
  Equation~\eqref{resonator1} the corresponding resonator, $\psi(t)$,
  represents the eigenfunction basis plus the bias $\Delta$.  This
  bias can be removed by subtracting $\Delta$ from $\psi(t)$.}

\commentout{To fully capture the covariance function prior, $\psi$ must satisfy
Equation~\eqref{equivcond} (that is, $\psi_j=a_j \phi_j$ where
$a_j\sim\mathcal{N}(0,\mu_j)$) and thus, encompass the random eigenfunction
coefficient weights, $a_j$.  Further, if the Kalman filter is used to
infer the resonator, the prior distribution over $\psi_j$ and
its derivative, $\dot{\psi}_j$, at some time $t_0$ is required to
initialise the Kalman filter \citep{sarkka12,hartikainen12,solin13}.
Since the coefficient weights are zero mean then both the prior mean
of $\psi_j$ and $\dot{\psi}_j$ are zero.  The prior covariance,
$\text{Cov}\left([\psi_j(t_0), \dot{\psi}_j(t_0)]\right)$, of the
resonator and its derivative at time $t_0$ is,
\begin{eqnarray*}
  \text{Cov}\left([\psi_j(t_0), \dot{\psi}_j(t_0)]\right)=\mu_j^2 \begin{pmatrix} 
    \phi_j(t_0)^2 & \phi_j(t_0) \dot{\phi}_j(t_0) \\ \phi_j(t_0) \dot{\phi}_j(t_0)
    & \dot{\phi}_j(t_0)^2
\end{pmatrix}
\end{eqnarray*}
where $\psi_j(t)=a_j \phi_j(t)$.  The
eigenfunction basis, $\phi_j$, and its derivative, $\dot{\phi}_j$, at
$t_0$ can be calculated using the Nystr\"om approximation as per
Equation~\eqref{phi}.}

To illustrate the link between the eigenfunction and corresponding
resonator models for perfectly periodic covariance functions we derive
the frequency process, $f(t)$, for a variation of the non-stationary
covariance function in Equation~\eqref{nonstatex} with a low
smoothness, $\nu=3/2$,
\begin{eqnarray}
K(t,t^\prime)=\text{Mat\'ern}(\kappa(t-t^\prime),\nu,\sigma,l)\exp(-\alpha(\kappa(t)^2+\kappa(t^\prime)^2))\ ,
\label{nonstatexpp}
\end{eqnarray} 
where $\kappa(\tau)=|\sin(\pi \tau /D)|$, $D$ is the covariance
function period and $\alpha>0$ is the decay rate.  This
covariance function differs from Equation~\eqref{nonstatex} in two
crucial respects.  Firstly, it is now perfectly periodic with period
$D$ and secondly, it is second order differentiable everywhere, as
required by Equation~\eqref{resfreq}.  For $\nu=3/2$ the Mat\'ern
simplifies,
\begin{eqnarray*}
\text{Mat\'ern}(\kappa(\tau),3/2,\sigma,l)=\sigma^2 (1+\sqrt{3}
\kappa(\tau)/l) \exp(-\sqrt{3}\kappa(\tau)/l)\ .
\end{eqnarray*} 
Using the Nystr\"om approximation, as per Equation~\eqref{phi},
\begin{eqnarray}
\frac{d^2 \tilde{\phi}_i(t)}{dt^2}=\frac{\sqrt{N}}{\mu_i} \frac{d^2
  K(t,S)}{dt^2} {\bf v}_i\ .
\label{dphi2}
\end{eqnarray} 
After some algebra,
\begin{eqnarray*}
\frac{d^2 K(t,t^\prime)}{dt^2}&=&\frac{d^2
  \text{Mat\'ern}(\kappa(\tau),3/2,\sigma,l)}{d\tau^2}
\exp(-\alpha\kappa(t)^2)\exp(-\alpha\kappa(t^\prime)^2)\\
& &+\text{Mat\'ern}(\kappa(\tau),3/2,\sigma,l)
\frac{d^2 \exp(-\alpha\kappa(t)^2)}{dt^2}\exp(-\alpha\kappa(t^\prime)^2)\\
& &+2 \frac{d
  \text{Mat\'ern}(\kappa(\tau),3/2,\sigma,l)}{d\tau}
\frac{d \exp(-\alpha\kappa(t)^2)}{dt}\exp(-\alpha\kappa(t^\prime)^2)\ ,
\end{eqnarray*}
where $\tau=t-t^\prime$ and,
\begin{eqnarray*}
\frac{d
  \text{Mat\'ern}(\kappa(\tau),3/2,\sigma,l)}{d\tau}&=&
-\frac{3 \pi \sigma^2 }{2 D l^2}\sin\left(\frac{2\pi}{D}\tau\right)\exp\left(-\frac{\sqrt{3}}{l}\kappa(\tau)\right)\ ,\\
\frac{d \exp(-\alpha\kappa(t)^2)}{dt}&=&
-\frac{\pi\alpha}{D}\sin\left(\frac{2\pi}{D} t\right)\exp(-\alpha\kappa(t)^2) \ ,\\
\frac{d^2  \text{Mat\'ern}(\kappa(\tau),3/2,\sigma,l)}{d\tau^2}&=&
\frac{3 \pi^2 \sigma^2}{D^2
  l^3}\exp\left(-\frac{\sqrt{3}}{l}\kappa(\tau)\right)\left(\sqrt{3}
  \kappa(\tau) (1-\kappa(\tau)^2) -l(1-2\kappa(\tau)^2)\right)\ ,\\
\frac{d^2 \exp(-\alpha\kappa(t)^2)}{dt^2}&=&
-\frac{2\pi^2\alpha}{D^2}\left(\frac{\alpha}{2}\left([1-2\kappa(t)^2]^2-1\right)+1-2\kappa(t)^2\right)\exp(-\alpha\kappa(t)^2)\ .
\end{eqnarray*}
Subsequently, using Equations~\eqref{resfreq} and~\eqref{dphi2} we can
determine the resonator model frequency process for each resonator
model so that the resonator is equivalent to the eigenfunction,
\begin{eqnarray}
(2\pi f_j(t))^2 = -\frac{1}{\tilde\phi_j(t)}\frac{d^2
  \tilde\phi_j(t)}{dt^2}\ .
\label{frequencyprofile}
\end{eqnarray}

\begin{figure}[ht]
\begin{center}
\begin{tabular}{cc} 
\includegraphics[width=0.5\textwidth]{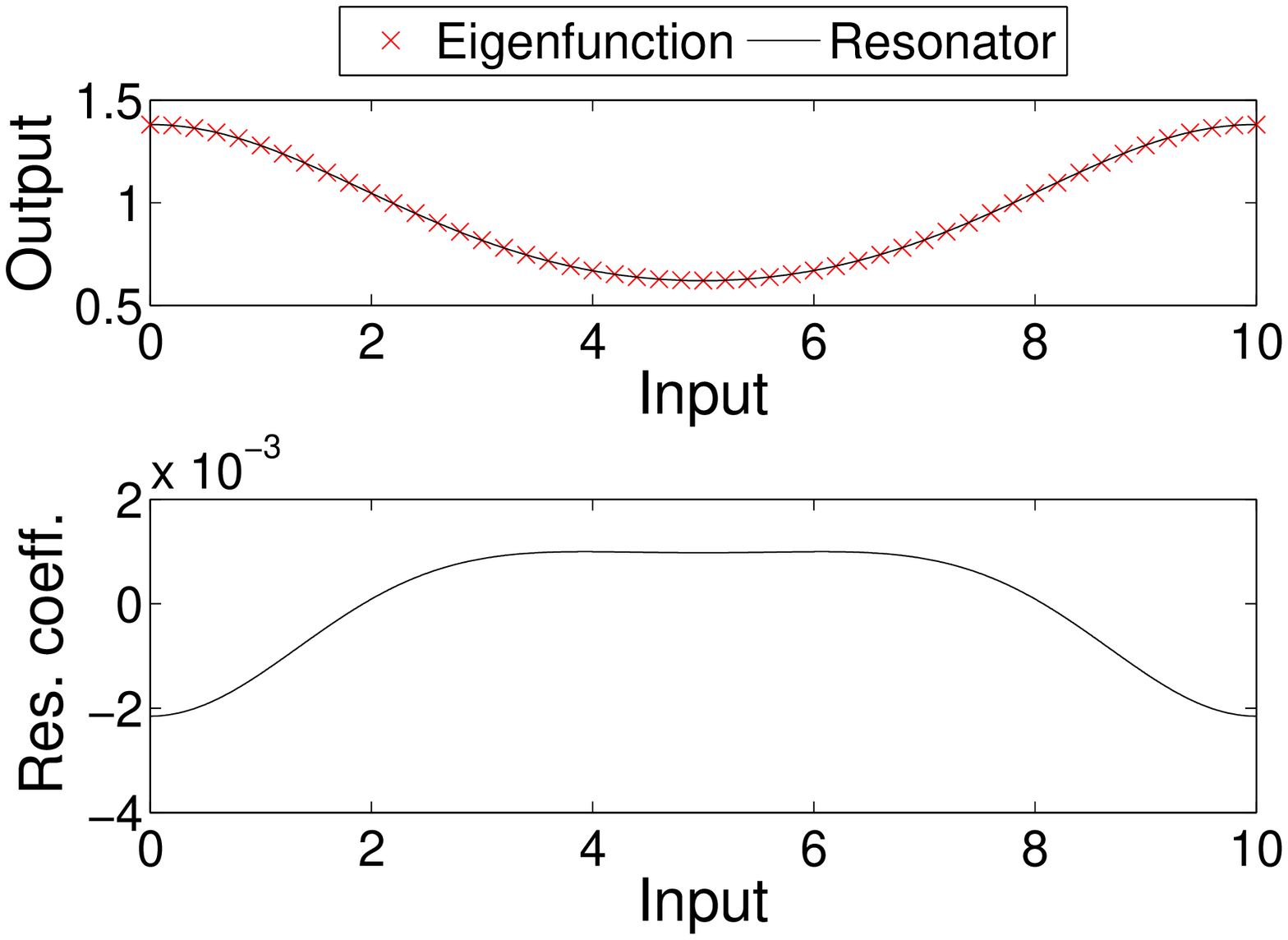}&
\includegraphics[width=0.5\textwidth]{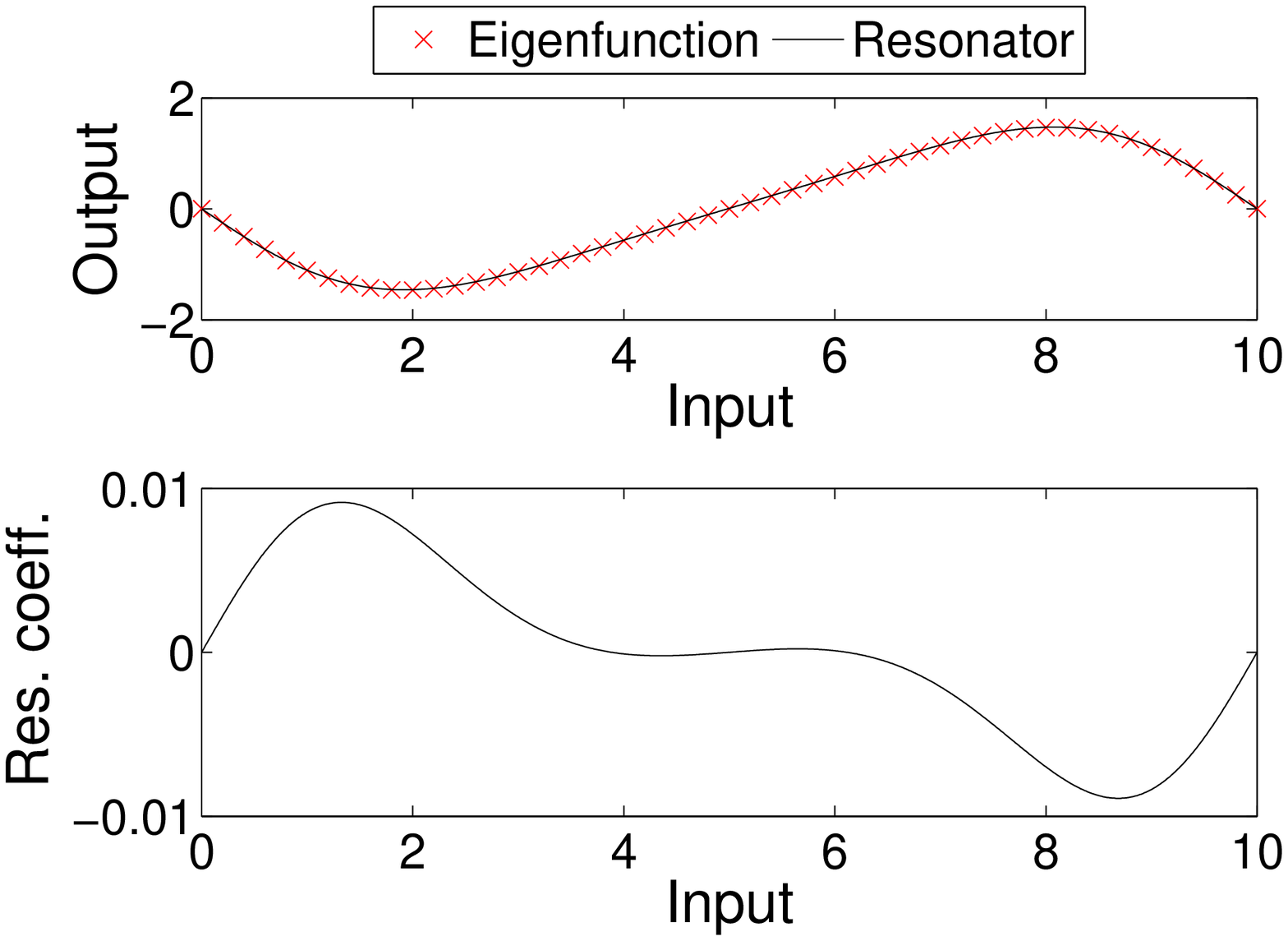}\\ 
Principal Eigenfunction & Second Eigenfunction\\ \ \\
\includegraphics[width=0.5\textwidth]{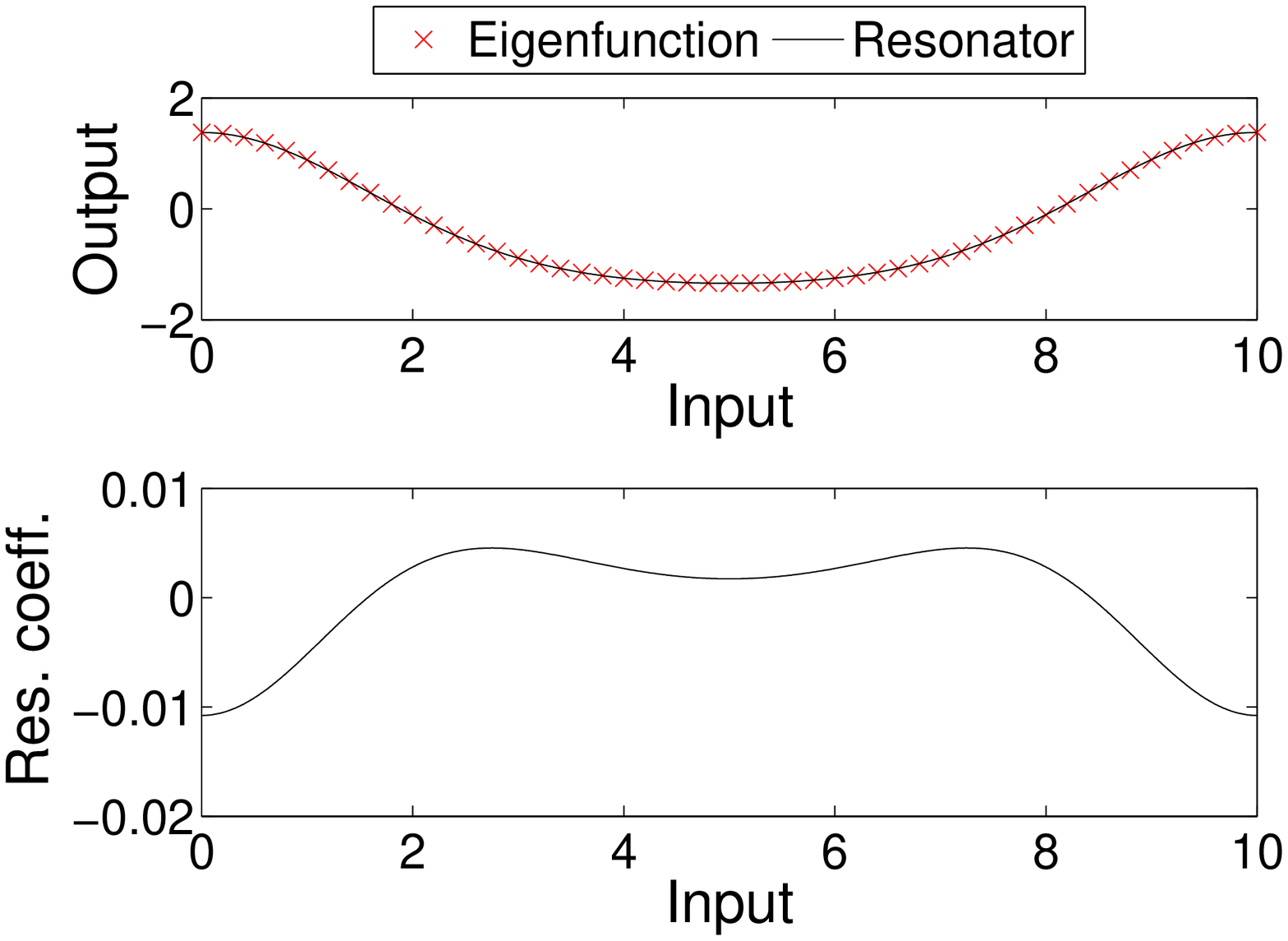}&
\includegraphics[width=0.5\textwidth]{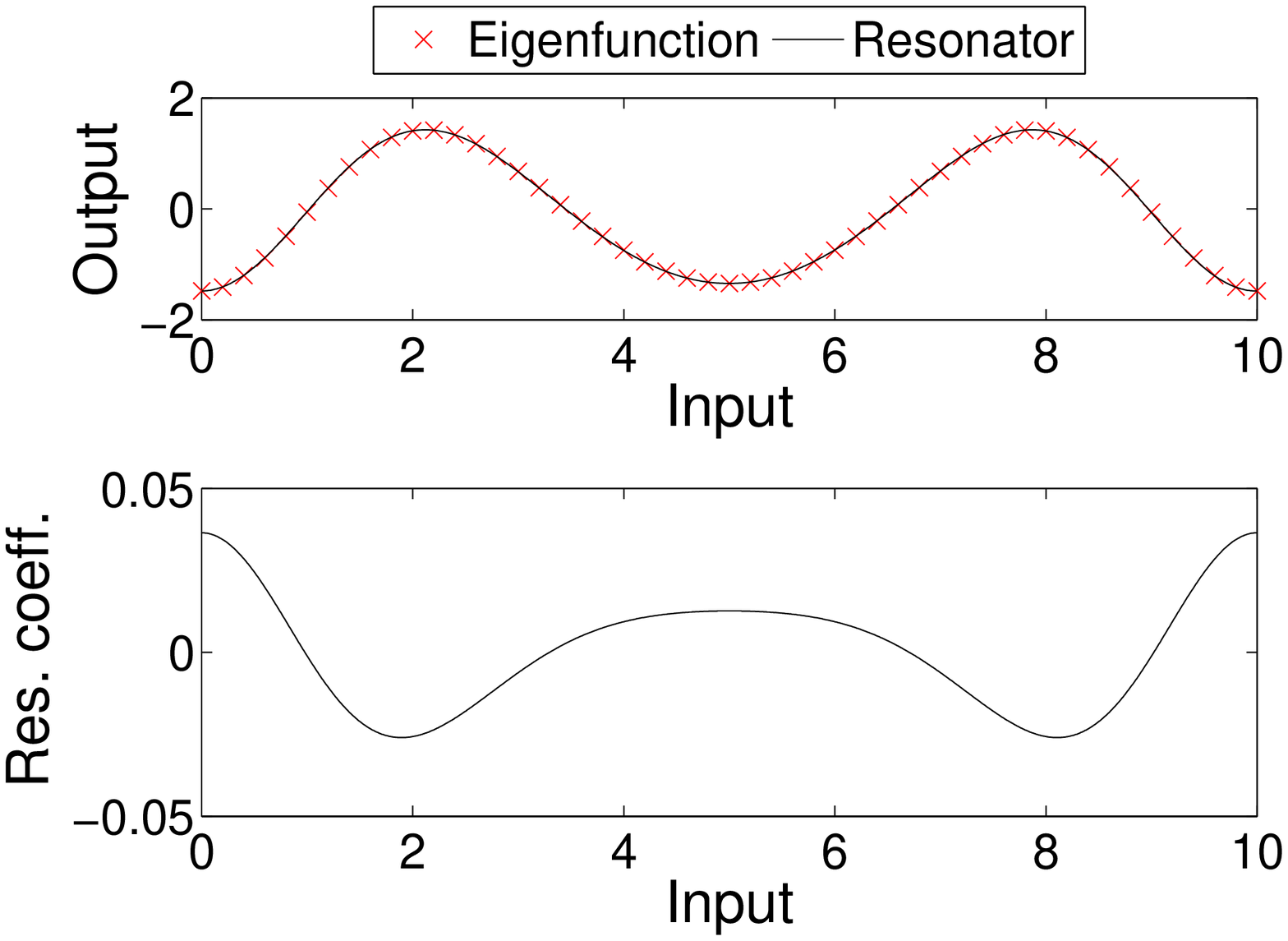}\\
Third Eigenfunction & Fourth Eigenfunction
\end{tabular}
\end{center}
\caption{\label{eigres_example} The four most significant
  Eigenfunctions and coincident resonators for the
  non-stationary periodic covariance function in
  Equation~\eqref{nonstatexpp}.  In each pane the top graph shows the
  eigenfunction and the lower graph shows the resonator coefficient
  $(2\pi f(t))^2$ profile required by the resonator model to equate
  the resonator with the eigenfunction.}
\end{figure}

Figure~\ref{eigres_example} compares the eigenfunction and
corresponding resonator whose frequency profiles are
calculated using Equation~\eqref{frequencyprofile}.  These basis
functions are the four most significant eigenfunctions for the
non-stationary periodic covariance function in
Equation~\eqref{nonstatexpp} with $D=10$, $\alpha=0.8$ and $l=20$. The
top panes show the eigenfunction and corresponding resonator and
the bottom panes show the resonator coefficient, $(2\pi f(t))^2$, as
per Equation~\eqref{frequencyprofile}, required by the resonator model
to equate the resonator with the eigenfunction.  We note the
presence of negative resonator coefficient values $(2\pi f(t))^2$.
These correspond to complex valued frequencies which model basis decay
in a manner similar to the basis decay term in the alternative
resonator model as per Equation~\eqref{resonator2}.

We next examine the properties of the alternative resonator model, as
per Equation~\eqref{resonator2}, when representing perfectly periodic,
non-stationary Gaussian processes.  The alternative resonator model
uses time invariant coefficients, $A$ and $B$,
\begin{eqnarray}
\frac{d^2 \psi_j(t)}{d t^2}+A_j\frac{d \psi_j(t)}{d
  t}+B_j\psi_j(t)=\omega_j(t)\ ,
\label{resbasis2b}
\end{eqnarray}   
where $\omega_j$ is a white noise component. Modelling the
non-stationary process via Equation~\eqref{resbasis2b} avoids the need
to compute frequency processes using the interacting multiple model
(IMM) in the original formalisation of the resonator model
\citep{sarkka12}.  For perfectly periodic covariance functions (with
period $D$) then $\psi_j(t+D)=\psi_j(t)$ for all $t$ and consequently,
as $\omega_j$ is i.i.d., then $\omega_j(t)=0$ for all $t$.  Thus, the
solution of the previous equation is,
\begin{eqnarray*}
\psi_j(t)=G_j \exp\left[(\pm i\sqrt{B_j-0.25 A_j^2}-0.5 A_j) t\right]\ .
\end{eqnarray*} 
So that $\psi_j(t+D)=\psi_j(t)$ for all $t$ then $A_j=0$ and thus,
\begin{eqnarray*}
\psi_j(t)=G_j \exp\left[ i\sqrt{B_j} t\right]\ .
\end{eqnarray*}
Consequently, expanding the exponential in terms of cosine and sine
functions we see that $\psi_j$ must be the Fourier basis functions.
The Fourier basis is a sub-optimal basis for non-stationary covariance
functions as, in general, the optimal eigenfunction basis is not
Fourier (see, for example, Figure~\ref{eigres_example}).  Thus, the
resonator model, as per Equation~\eqref{resbasis2b}, is a sub-optimal
representation for non-stationary periodic covariance functions.

\subsubsection{Quasi-Periodic Covariance Functions}
\label{sec:quasiper}
A {\em quasi-periodic process} $g\sim\mathcal{GP}(b,K)$ is generated
from a Gaussian process with covariance function of the form
$K(t,t^\prime)=K_\text{quasi}(t,t^\prime)K_\text{periodic}(t,t^\prime)$
where $K_\text{quasi}$ is non-periodic and $K_\text{periodic}$ is
perfectly periodic (either stationary or
non-stationary). Equation~\eqref{quasiperiodicexample} is an example
of a quasi-periodic process covariance function.  In general, when the
periodic kernel, $K_\text{periodic}$, has period $D$ then, with high
probability, $g(t+D)\not=g(t)$ for all $t\in \mathbb{R}$ unlike the
perfectly periodic case presented above.

Quasi-periodic eigenfunction models use a time varying weight
coefficient, $a_j(t)$, as per Equation~\eqref{stochcoeff}.  Thus,
extending Equation~\eqref{equivcond}, in this case the resonator and
eigenfunction linear basis models are equivalent if,
\begin{eqnarray*}
\psi_j(t)=a_j(t) \phi_j(t)\ .
\end{eqnarray*} 
Consequently, by substituting $\psi_j(t)$ into
Equation~\eqref{resonator1}, setting $\omega_j(t)=0$ and rearranging we get
the frequency process, $f_j(t)$, for each resonator for the
quasi-periodic process,
\begin{eqnarray*}
(2\pi f_j(t))^2=-\frac{\ddot{\phi}_j(t) a_j(t)+\phi_j(t)
  \ddot{a}_j(t)+2\dot{\phi}_j(t) \dot{a}_j(t)}{a_j(t)\phi_j(t)}\ .
\end{eqnarray*}
Since the coefficient process $a_j(t)$ is stochastic then so too is
$f_j(t)$. We note that both $\psi_j(t)$ and $a_j(t)$ must be inferred
when using the resonator model and the Kalman filter.  This places
significant computational cost on the Kalman filter prediction
equations.  Thus, we do not recommend implementing quasi-periodic GP
priors with the resonator model as per Equation~\eqref{resonator1}.

The alternative resonator model, as per Equation~\eqref{resonator2}
encodes a decay term, via the first order derivative of the basis,
appropriate for modelling quasi-periodic covariance
functions.  This model is investigated empirically in
Section~\ref{sec:appthermal} on a home heating prediction problem
which exploits quasi-periodic latent forces.

\subsection{Computational Complexity of Eigenfunction and Resonator
Models}
\label{sec:compcompl}

When the Gaussian process covariance function for each latent force is
known, so that we can generate the appropriate eigenfunction basis for
any choice of covariance function hyperparameters, then searching
over the hyperparameter values of the covariance function can be
significantly less computationally demanding than searching over the
frequency space for a potentially large number of resonators.

When constructing the eigenfunction model the greatest computational
cost arises from calculating the Nystr\"om approximation.  However,
the significant eigenfunctions can be found iteratively and
efficiently using Von Mises iteration.  At each iteration the next
largest eigenvalue and corresponding eigenfunction are found.  This
approach continues until all the significant eigenvalues are found.
If $J$ eigenfunctions with the largest eigenvalues are found
using Von Mises iteration then the complexity of our approach is
$\mathcal{O}(J N^2)$ where $N\times N$ is the size of the Gram matrix in
Equation~\eqref{gramG} obtained by sampling the periodic covariance
function. Inferring the eigenfunction model also involves searching
over a relatively small set of $p$ hyperparameters, often of the
order of about $p=3$ parameters comprising the input scale, output
scale and the period of the covariance function.  If the set of
admissible values along each hyperparameter dimension has cardinality
$\Upsilon$ then the computational complexity of searching the parameter space
is $\mathcal{O}(\Upsilon^p)$.  The overall computational complexity of inferring
the eigenfunction model is therefore $\mathcal{O}(\Upsilon^p J N^2)$.

The parameters of the J-dimensional resonator model can be found by
solving a non-convex optimisation problem over a $3J$ dimension
parameter space where the parameters are $J$ Fourier basis function
frequencies, $J$ basis function phases and $J$ magnitudes for the
basis power spectrum.  If the set of admissible values along each
dimension has cardinality $\Upsilon$ then the computational complexity
of searching the parameter space is $\mathcal{O}(\Upsilon^{3J})$.  To
identify the optimal choice of parameter values each parameter vector
constructed during the search over the parameter space requires the
comparison of $\mathcal{O}(N^2)$ entries between the sampled kernel
and the covariance matrix of the target function induced by the
resonator model.  Thus the resonator model is inferred with
computational complexity $\mathcal{O}(\Upsilon^{3J} N^2)$.  We note
that, whereas the resonator model training phase is exponentially
complex in the number of basis functions, $J$, the eigenfunction model
is linear in $J$.~\footnote{Note that, when using the Nystr\"om
  approximation of the eigenfunctions it is necessary to store the
  eigenvectors from which the eigenfunctions can be calculated.
  Although, for the stationary case, they can be calculated when
  required from the cosine and sine functions.}

We compare the run times for the eigenfunction and resonator
approaches empirically in Section~\ref{sec:appthermal}.

\subsection{Summary}
We have demonstrated the link between the resonator model and the
eigenfunction approach. Through this link we have been able to
identify that,
\begin{enumerate}
\item the eigenfunction basis is optimal in that it minimises the mean
  squared error between the J-dimensional model and the target function.
\item the variant frequency term in the resonator second order
  differential equation provides sufficient flexibility to yield basis functions
  which are equivalent to the eigenfunctions.
\item we have developed an algorithm for deriving optimal resonator
  models for all perfectly periodic covariance functions from the
  eigenfunctions of the covariance function.  Thus, we are able to
  offer an efficient mechanism for encoding the GP prior in the resonator model.
\end{enumerate}
\end{document}